\documentclass[opre,sglanonrev]{informs4}

\usepackage{microtype}

\usepackage{bm}
\usepackage{mathrsfs} 
\usepackage{amsopn}

\usepackage{bbm} %
\usepackage{enumitem}
\usepackage{natbib}
\usepackage{dsfont} %
\usepackage[normalem]{ulem} %
\usepackage{appendix}
\usepackage[unicode=true,pdfusetitle, bookmarks=true,bookmarksnumbered=true,bookmarksopen=false, breaklinks=true,pdfborder={0 0 0},pdfborderstyle={},backref=page,colorlinks=true,citecolor=blue]{hyperref}

\allowdisplaybreaks

\usepackage{graphicx}
\usepackage{wrapfig}
\usepackage{tikz} %
\usepackage{float}
\usepackage{subfigure}

\newtheorem{condition}{Condition}

\newtheorem{observation}{Observation}

\global\long\def\E{\mathbb{E}}%
\global\long\def\N{\mathbb{N}}%
\global\long\def\mS{\mathcal{S}}%
\global\long\def\mA{\mathcal{A}}%
\global\long\def\mT{\mathcal{T}}%
\global\long\def\R{\mathbb{R}}%
\global\long\def\P{\mathbb{P}}
\newcommand{\var}{\operatorname{Var}}
\global\long\def\tmT{\widetilde{\mT}}
\global\long\def\law{\mathcal{L}}

\usepackage{xcolor}
\usepackage{todonotes} %
\usepackage[showdeletions]{color-edits} %
\addauthor[Yixuan]{yz}{blue} %
\addauthor[Qiaomin]{qx}{red} %
\addauthor[Yudong]{yc}{purple} %
\addauthor[Lucy]{lh}{magenta} %

\usepackage{natbib}
 \bibpunct[, ]{(}{)}{,}{a}{}{,}%
 \def\bibfont{\small}%
 \def\newblock{\ }%

\OneAndAHalfSpacedXII

\EquationsNumberedThrough    

\TheoremsNumberedThrough     
\ECRepeatTheorems

\begin{document}

\TITLE{Prelimit Coupling and Steady-State Convergence of Constant-stepsize Nonsmooth Contractive SA}

\ARTICLEAUTHORS{%
\AUTHOR{Yixuan Zhang}
\AFF{Department of Industrial \& Systems Engineering,
University of Wisconsin-Madison, \EMAIL{yzhang2554@wisc.edu}}

\AUTHOR{Dongyan (Lucy) Huo}
\AFF{Department of Industrial Engineering \& Decision Analytics,
Hong Kong University of Science and Technology, \EMAIL{lucyhuo@ust.hk}}

\AUTHOR{Yudong Chen}
\AFF{Department of Computer Sciences,
University of Wisconsin-Madison, \EMAIL{yudongchen@cs.wisc.edu}}

\AUTHOR{Qiaomin Xie}
\AFF{Department of Industrial \& Systems Engineering,
University of Wisconsin-Madison, \EMAIL{qiaomin.xie@wisc.edu}}
} 

\AREAOFREVIEW{Stochastic Models}
\SUBJECTCLASS{\textbf{Markov processes-Probability:} Formulate SA as Markov chain; analyze limit as stepsize→0.  \textbf{Stochastic model applications-Probability:}  Our results for SA can be applied to Q-learning algorithms. \textbf{Stochastic-Programming:} We study the nonlinear and nonsmooth SA.
 }

\ABSTRACT{
    Motivated by Q-learning, we study nonsmooth contractive stochastic approximation (SA) with constant stepsize. We establish the weak convergence of the iterates to a stationary limit distribution in Wasserstein distance. Furthermore, we develop a prelimit coupling technique to establish steady-state convergence and characterize the limit of the stationary distribution as the stepsize goes to zero. As an 
intermediate step, we prove a universal result that the general nonsmooth contractive SA is asymptotically equivalent to the nonsmooth contractive SA with only additive Gaussian noise. This result, of independent interest, serves as a powerful tool to analyze SA. By steady-state convergence result, we show that the asymptotic bias of nonsmooth SA is proportional to the square root of the stepsize, which stands in sharp contrast to smooth SA. Importantly, this bias characterization enables the use of Richardson-Romberg extrapolation for bias reduction in nonsmooth SA.}
\FUNDING{Q. Xie and Y. Zhang are supported in part by National Science Foundation (NSF) grants CNS-1955997, EPCN-2339794, and EPCN-2432546. Y. Chen is supported in part by NSF grants CCF-1704828 and CCF-2233152, and by the Vilas Associate Award from the University of Wisconsin--Madison.
}


\maketitle




\section{Introduction}

Stochastic Approximation (SA) is a fundamental algorithmic paradigm for iteratively solving fixed-point problems based on noisy observations. SA methods have been widely used in many application domains, including reinforcement learning (RL), stochastic control, and optimization \citep{Bertsekas19-RL-book, Sutton18-RL-book, kushner2003-yin-sa-book, moulines2011non}. A typical SA recursion is of the form 
\begin{equation}\label{eq: iid general SA}
\theta_{t+1}^{(\alpha)} = \theta_t^{(\alpha)} + \alpha \big(\tmT(x_t, \theta_t^{(\alpha)}) - \theta_t^{(\alpha)}\big),
\quad t=0,1,\ldots
\end{equation}
where $\{x_t\}_{t \geq 0}$ is an i.i.d.\ noise sequence and $\alpha >0 $ is a constant stepsize. The SA recursion~\eqref{eq: iid general SA} aims to approximately find the solution $\theta^*$ to the fixed-point equation $\mT(\theta^*) = \theta^*,$ where $\mT(\cdot):= \E_{x} [\tmT(x, \cdot )]$ is the expectation of the operator $\tmT(x, \cdot)$ with respect to noise. Equation~\eqref{eq: iid general SA} covers many iterative algorithms, such as the stochastic gradient descent (SGD) algorithm for minimizing an objective function \citep{lan2020first}, and variants of TD-learning algorithms for policy evaluation in RL \citep{Sutton18-RL-book}. 

In this work, we focus on \emph{nonsmooth} contractive SA, where the operator $\tmT(x, \cdot)$ may be nondifferentiable (in its second argument) and $\mT(\cdot)$ is a contractive mapping with respect to a norm $\|\cdot\|_c$. One prominent example of nonsmooth contractive SA is the celebrated Q-learning algorithm for optimal control in RL~\citep{Watkins92-QLearning}, where $\tmT$ corresponds to the noisy Bellman optimality operator, which involves a max function. Other common nonsmooth mappings include the largest eigenvalue of a matrix, $\ell_1$-norm regularized functions,  and their composition with smooth functions \citep{sagastizabal2013composite,shapiro2003class}. It is of fundamental interest to gain a deeper understanding of the evolution and long-run behavior of the iterates $\{\theta_t^{(\alpha)}\}_{t \geq 0}$ generated by nonsmooth contractive SA. 

Under suitable conditions on the operator $\tmT$ and the noise sequence $\{x_t\}_{t \geq 0}$, the SA iterates $\{\theta_t^{(\alpha)}\}_{t \geq 0}$ form a time-homogeneous Markov chain and quickly converge to some limit random variable $\theta^{(\alpha)}$ \citep{dieuleveut2020bridging,yu2021analysis}. 
Recent work has obtained a suite of results for \emph{smooth} SA \citep{dieuleveut2020bridging,  huo2023bias, durmus2021riemannian,huo2024collusion,zhang2025piecewise}, including the geometric convergence of the chain, finite-time bounds on the higher moments,  as well as various properties of the limit $\theta^{(\alpha)}.$ It has been observed that  $\E[\theta^{(\alpha)}] \neq \theta^* $, due to the use of constant stepsize. The difference $\E[\theta^{(\alpha)}]-\theta^*$ is referred to as the \emph{asymptotic bias}. 
In particular, for SA with \emph{differentiable} dynamic, the work in \cite{dieuleveut2020bridging, huo2023bias} makes use of the Taylor expansion of $\tmT$ to establish that the asymptotic bias is proportional to the stepsize $\alpha$ (up to a higher order term), i.e., 
\begin{equation}\label{eq: bridge}
\E[\theta^{(\alpha)}] - \theta^* = \alpha c + o(\alpha),
\end{equation}
where $c$ is some vector independent of $\alpha$ and $o(\alpha)$ denotes a term that decays faster than $\alpha$. Such a fine-grained characterization of SA iterates allows for the use of variance and bias reduction techniques that improve the estimation of the target solution $\theta^*$, as well as the development of efficient statistical inference procedures~\citep{dieuleveut2020bridging,huo2023bias,huo2023effectiveness}. \medskip

For nonsmooth SA, relatively little is known. The existing analysis based on the linearization or Taylor expansion of $\tmT$ is no longer applicable due to the nondifferentiability of $\tmT$. Consequely, key results such as distributional convergence and bias characterization as in~\eqref{eq: bridge} have not been established for nonsmooth SA procedures, including Q-learning. In fact, it is not even clear whether equation~\eqref{eq: bridge} remains valid for nonsmooth SA, or if not, what the correct characterization should be. 


\textbf{Our Contributions:} The first main result of this paper shows that the Markov chain
$\{\theta_t^{(\alpha)}\}_{t \ge 0}$, generated by a general contractive SA scheme converges weakly in
$W_2$ to a \emph{unique} stationary distribution, where $W_2$ denotes the Wasserstein distance of order 2 induced by the contraction norm
$\|\cdot\|_c$.
Moreover, we establish an explicit geometric convergence rate.
As a by-product, our analysis delivers finite-time upper bounds on the
$2n$-th moments of the estimation error, $\E \|\theta_t^{(\alpha)} - \theta^*\|_c^{2n}$, thereby extending the mean-square error bounds of
\cite{chen2020finite,chen2023lyapunov} to higher moments and
generalizing the smooth contractive SA results of
\cite{dieuleveut2020bridging,srikant2019finite} to the general contractive
setting.

We next provide a precise characterization of the stationary distribution
of $\{\theta_t^{(\alpha)}\}_{t \ge 0}$.
Here we focus on the class of potentially nonsmooth SA procedures whose
operator $\mT$ is only assumed to be \emph{one-sided directionally differentiable} at
$\theta^*$ (see
Assumption~\ref{as: contraction}).
 As existing techniques based on linearizing $\tmT(x, \theta)$ as $\theta\to\theta^*$ are no longer applicable for nonsmooth SA, we take an alternative approach. In particular, we study the limiting behavior of properly rescaled iterates as the constant stepsize $\alpha$ approaches 0. 
Since the MSE of $\theta_t^{(\alpha)}$ is of order $\mathcal{O}(\alpha)$ \citep{chen2020finite}, the appropriate scaling is by $\sqrt{\alpha}$, often known as the diffusion scaling. Specifically, we consider the centered and $\sqrt{\alpha}$-scaled iterates $Y_t^{(\alpha)}:= \frac{\theta_t^{(\alpha)} - \theta^*}{\sqrt{\alpha}}.$
The weak convergence of $\theta_t^{(\alpha)} $ to a limit $\theta^{(\alpha)} $ implies that $Y_t^{(\alpha)}$ converges to the limit $ Y^{(\alpha)} :=\frac{\theta^{(\alpha)} - \theta^*}{\sqrt{\alpha}}$ as $t\to \infty$. Therefore, to understand the stationary distribution $\theta^{(\alpha)}$ and its scaled version $Y^{(\alpha)}$, we focus on characterizing  \emph{steady-state convergence}, i.e., the convergence of $Y^{(\alpha)}$ as $\alpha \to 0$ and the corresponding limit $Y$ (if  exists). Our approach is illustrated by the red solid path in Figure~\ref{fig:limit-plot}.

\begin{figure}[htbp]
    \centering
     \begin{tikzpicture}
     \draw[dashed,->] (0,2.5)--(0,0.5);
      \draw[dashed, ->] (0.5,0)--(4.5,0);
       \draw[red, thick, ->] (0.5,3)--(4.5,3);
        \draw[red, thick, ->] (5,2.5)--(5,0.5);
    \draw (0,3) node {$Y_t^{(\alpha)}$};
    \draw (5,3) node {$Y^{(\alpha)}$};
    \draw (2.5,3.2) node {$t\to\infty $};
     \draw (6,1.5) node {$\alpha\downarrow0 $};
     \draw (-0.2,1.5) node {$\alpha t = \bar{t}\quad$ $\alpha \downarrow 0$};
    \draw (0,0) node {$\overline{Y}_{\bar{t}}$};
    \draw (2.5 ,-0.3) node {$\bar{t} \to \infty$};
    \draw (5,0) node {$Y$};
    \end{tikzpicture}
    \caption{Steady-state convergence.
    }
    \label{fig:limit-plot}
\end{figure}

As to be discussed in Section~\ref{sec: challenge}, existing approaches for analyzing steady-state convergence encounter significant challenges in the context of nonsmooth SA. In this work, we develop a new \emph{prelimit coupling technique}, which allows us to establish the weak convergence of $Y^{(\alpha)}$ in $W_2$ to a unique limiting random variable $Y$ as $\alpha \to 0$. Importantly, our technique handles both additive noise and multiplicative noise, provides an explicit convergence rate, and is outlined in Section~\ref{sec: technique}. We remark that our technique may be of independent interest and potentially applied to the study of steady-state convergence in other stochastic dynamical systems.

Since convergence in $W_2$ implies convergence of the first two moments of $Y^{(\alpha)}$, we obtain the following characterization of the asymptotic bias of the SA iterates:
\begin{equation}\label{eq: bias}
\E[\theta^{(\alpha)}] - \theta^* = \E[Y]\cdot\sqrt{\alpha} + o(\sqrt{\alpha}).
\end{equation}
We further provide a characterization of the coefficient $\E[Y]$ in the leading term in \eqref{eq: bias} and relate this coefficient to the structure of the SA update~\eqref{eq: iid general SA}. Our results show that $\E[Y]\neq0$ precisely when the operator $\mT$ is truly nonsmooth, in which case the asymptotic bias is of order $\sqrt{\alpha}$. This result stands in sharp contrast to the $\alpha$-order bias of smooth SA in equation \eqref{eq: bridge}.

We next investigate the application of our result to Q-learning. Exploiting the specific structure of the Q-learning algorithms, we provide an explicit convergence rate of steady-state convergence, which leads to a more precise characterization of the bias. 

Finally, we explore the implications of the above results for iterate averaging and extrapolation. In particular, we consider applying Polyak-Ruppert (PR) tail averaging \citep{ruppert1988efficient, polyak1992acceleration, jain2018parallelizing} and Richardson-Romberg (RR) extrapolation \citep{hildebrand1987introduction} to the SA iterates. We study the resulting estimation errors and biases in the presence of nonsmoothness. In particular, thanks to the bias characterization in \eqref{eq: bias}, we show that one can employ the RR extrapolation technique to eliminate the leading term $\E[Y]\cdot\sqrt{\alpha}$ and reduce the asymptotic bias to a higher order of $\sqrt{\alpha}$.


\subsection{Challenges of Applying Existing Techniques to Nonsmooth SA}\label{sec: challenge}

Steady-state convergence, i.e., showing $Y^{(\alpha)} \to Y$ in Figure~\ref{fig:limit-plot}, is a problem of fundamental interest in stochastic dynamical systems, such as queueing networks~\citep{GamaZeev2006}. One well-known approach for proving steady-state convergence in queueing networks is to justify the \emph{interchange of limits} \citep{GamaZeev2006, Gurv2014a, YeYao2016, YeYao2018}, i.e., equivalence of the solid and dashed paths in Figure~\ref{fig:limit-plot}, followed by analyzing the dashed path. This approach has been well recognized as technically challenging, often requiring sophisticated ``hydro-dynamic limits" methodology~\citep{Bram1998a} and a well-defined stochastic differential equation (SDE) for $\bar{Y}_{\bar{t}}$ with a stationary distribution. In our setting, it is unclear whether nonsmooth SA is associated with such an SDE, let alone the validity of interchanging the limits.

An alternative approach to steady-state convergence is based on the \emph{Basic Adjoint Relationship} (BAR) for the generator of the stochastic process. By using BAR with an exponential test function, one may be able to prove convergence of the moment generating function and consequently the weak convergence of the corresponding random variables \citep{BravDaiMiya2017, BravDaiMiya2023,chen2022stationary}. In our setting, however, applying BAR with moment generating functions does not always lead to a straightforward solution. In fact, even for \emph{smooth} SA dynamics with only additive noise, i.e., $\tmT(x, \theta) = \mT(\theta) + x$,  steady-state convergence is proved in the work \cite{chen2022stationary} under the assumption that the following equation from BAR has a unique solution in $Y$:
\begin{equation}\label{eq: zaiwei assumption}
\mathbb{E}\big[\big(\varphi^{\top} \var(x) \varphi-2 i \varphi^{\top} \nabla\mT(\theta^*) Y\big) e^{i \varphi^{\top} Y}\big]=0, \quad\forall \varphi \in \mathbb{R}^d.
\end{equation}
Verifying this assumption is challenging in general; in \cite{chen2022stationary} this is done only when $d = 1$ or under restricted conditions when $d \geq 2$. This difficulty is only exacerbated in the nonsmooth SA setting, which lacks many nice structures of smooth SA as considered in \cite{chen2022stationary} . 

The $\sqrt{\alpha}$-scaling suggests a potential connection to the Langevin diffusion SDE and recent work on the Unadjusted Langevin Algorithm (ULA)~\citep{durmus2017nonasymptotic, durmus2019high}.
 ULA corresponds to the Euler-Maruyama discretization of the Langevin diffusion and is given by
\[Z_{t+1}^{(\alpha)} = Z_t^{(\alpha)} - \alpha \nabla U(Z_t^{(\alpha)}) + \sqrt{\alpha}x_t,\]
where $U:\R^d\to\R$ is a potential function and $\{x_t\}_{t \geq 0}$ are i.i.d.\ Gaussian noise.
However, by comparing ULA with the SA update~\eqref{eq: iid general SA} scaled by $\sqrt{\alpha}$, we note that the latter reduces to ULA only when
the noise is additive and Gaussian and $\mT$ is a gradient field that is positive homogeneous at $\theta^*$.

We complement the above discussion with a simple example of nonsmooth contractive SA:
\begin{equation}\label{eq: simple example}
\theta_{t+1}^{(\alpha)} = \theta_t^{(\alpha)} + \alpha \Big(-\frac{1}{2}{|\theta_t^{(\alpha)}|} -b - \theta_t^{(\alpha)} + x_t \Big),
\end{equation}
where $x_t \overset{\operatorname{i.i.d.}}{\sim} \mathcal{N}(0,1)$ and $b \in \R$. The above dynamic is a special case of~\eqref{eq: iid general SA} with $\tmT(x, \theta)=-|\theta|/2-b+x$, which is non-differentiable at $\theta = 0$. Despite its apparent simplicity, this example already demonstrates some of the complexity in understanding the steady-state behavior of nonsmooth SA. For instance, it is unclear how to apply the BAR approach to obtain a functional equation like \eqref{eq: zaiwei assumption} for the limit $Y$. The derivation of Equation~\eqref{eq: zaiwei assumption} in \cite{chen2022stationary} relies on the continuous differentiability of the contraction operator $\tilde{T}(x, \theta)$. In particular, deriving the limit of the $T_2$ term in \citet[Page~15]{chen2022stationary} requires this property, which does not hold here.
 On the other hand, when $b = 0$, the dynamic~\eqref{eq: simple example} reduces to an ULA update with $U(z)= \begin{cases}
3z^2/4, & \text{if } z \geq 0, \\
z^2/4, & \text{otherwise}.
\end{cases}$ The results in \cite{durmus2017nonasymptotic, durmus2019high} on ULA suggest that the limit $Y$ is not Gaussian, as its density function $e^{-U(y)}/\int e^{-U(z)}dz$ involves a non-quadratic exponent $U$. This is in contrast to smooth SA, for which the BAR approach shows that $Y$ is Gaussian~\citep{chen2022stationary}. As we see in the sequel, the techniques developed in this paper bypass the need of working directly with functional equations like~\eqref{eq: zaiwei assumption} or imposing restrictive assumptions on such equations.

\subsection{Prelimit Coupling Technique}
\label{sec: technique}

To overcome the challenges discussed in the previous subsection, we develop a prelimit coupling technique for establishing the desired steady-state convergence without restrictive assumptions.
In particular, we seek to prove convergence in Wasserstein distance $W_2$, i.e., 
\begin{equation*}
    \lim_{\alpha \to 0}W_2\big(\mathcal{L}({Y}^{(\alpha)}), \mathcal{L}(Y)\big)=0
\end{equation*} 
for some random variable $Y$ with distribution $\mathcal{L}(Y)$. Our technique applies coupling arguments to the prelimit random variables $Y^{(\alpha)}_t$ with finite $\alpha>0,t<\infty,$ and consists of three steps.
    
\paragraph{Step 1: Reduction to Additive Gaussian Noise} 

Rewrite the SA update \eqref{eq: iid general SA} as
\begin{align}
    \theta_{t+1}^{(\alpha)} &= \theta_{t}^{(\alpha)} + \alpha \big( \mT(\theta_{t}^{(\alpha)})-\theta_{t}^{(\alpha)}+w_{t}(\theta_{t}^{(\alpha)}) \big),\label{eq: iid general SA2}
\end{align}
where $w_{t}(\theta_{t}^{(\alpha)}) := \tmT(x_t,\theta_{t}^{(\alpha)})-\mT(\theta_{t}^{(\alpha)})$ is the \emph{multiplicative} noise that can depend on the iterate $\theta_{t}^{(\alpha)}$. Now let us consider a simpler update with the same expected operator $\mT$:
\begin{align}
    \beta_{t+1}^{(\alpha)} &= \beta_{t}^{(\alpha)}+\alpha\big(\mT(\beta_{t}^{(\alpha)})-\beta_{t}^{(\alpha)}+w_{t}'\big),\label{eq: iid approximate}
\end{align}
where $w_{t}'\sim N(0,\Sigma)$ is i.i.d.\ noise with covariance matrix $\Sigma=\operatorname{Var}(w_{0}(\theta^{*})),$ matching that of $w_t(\theta^*)$. Importantly, the update \eqref{eq: iid approximate} only involves \emph{additive Gaussian} noise $w_{t}'$ that is \emph{independent} of $\theta_{t}^{(\alpha)}$. 
Recall that $Y^{(\alpha)} = \frac{\theta^{(\alpha)}-\theta^*}{\sqrt{\alpha}}$ represents the scaled version of the steady-state random variable $\theta^{(\alpha)}$, to which the process $\{\theta^{(\alpha)}_t\}_{t\ge 0}$ converges in $W_2$. Similarly, we consider the steady-state $\beta^{(\alpha)}$ for the process $\{\beta_t^{(\alpha)}\}_{t\ge 0}$ and define $Z^{(\alpha)}= \frac{\beta^{(\alpha)}-\theta^*}{\sqrt{\alpha}}$. 

In Step 1 of the analysis, we establish the following key observation.


\begin{observation}[Informal statement of Proposition \ref{prop:gaussian}]\label{ob:bridging}
As $\alpha \to 0$, if $Z^{(\alpha)}$ converges in $W_2$ to some random variable $Y$, then $Y^{(\alpha)}$ converges to the same limit $Y$.
\end{observation}

Therefore, for the purpose of establishing steady-state convergence of SA in \eqref{eq: iid general SA2}, it suffices to analyze the update~\eqref{eq: iid approximate} with noise that is additive and Gaussian. These two structures of the noise are heavily exploited in the next two steps of our analysis, where we establish convergence of $Z^{(\alpha)}$.

Notably, Observation~\ref{ob:bridging} establishes a \emph{universality} property: the limit $Y$  only depends on the expected operator $\mT$ and the covariance of the noise random field $w_t(\cdot)$ at $\theta^*$, and it is  independent of the specific distribution of the noise $w_t$. Exploiting this universality property, which is reminiscent of the Lindeberg replacement method~\citep{lindeberg1922neue}, greatly simplifies the task of characterizing the steady-state limit $Y$, as it suffices to consider additive Gaussian noise and focus on the influence of the expected operator $\mT$ on $Y$.

To prove Observation \ref{ob:bridging}, we bound the $W_2$ difference between the updates~\eqref{eq: iid general SA2} and~\eqref{eq: iid approximate}  by coupling ${Y_t}^{(\alpha)}$ and $Z^{(\alpha)}_t$, respectively the scaled versions of the prelimit iterates $\theta_t^{(\alpha)}$ and $\beta_{t}^{(\alpha)}$. In particular, setting $\kappa= \lfloor \alpha^{-1/2} \rfloor$, we use a multivariate Berry-Esseen bound in Wasserstein distance \citep{bonis2020stein} to show that there exists a coupling between the noises $\{w_{m\kappa+t}(\theta^*)\}_{t=1}^\kappa$ and $\{w'_{m\kappa+t}\}_{t=1}^\kappa$, such that
\begin{align*}
\E \bigg\|\frac{1}{\sqrt{\kappa}}\sum_{t=1}^\kappa w_{m\kappa+t}(\theta^*)- \frac{1}{\sqrt{\kappa}}\sum_{t=1}^\kappa w^\prime_{m\kappa+t} \bigg\|_2^2 &= W_2^2\bigg(\mathcal{L}\Big(\frac{1}{\sqrt{\kappa}}\sum_{t=1}^\kappa w_{m\kappa+t}(\theta^*)\Big), \mathcal{L}\Big(\frac{1}{\sqrt{\kappa}}\sum_{t=1}^\kappa w^\prime_{m\kappa+t}\Big)\bigg) \\
&\in \mathcal{O}\Big(\frac{1}{\kappa}\Big),\;\; \forall m \in \mathbb{N}.
\end{align*}
Under this noise coupling, we show that $ \E \| {Y _{\kappa t}}^{(\alpha)} - Z_{\kappa t}^{(\alpha)} \|_c^2\in \mathcal{O}(\alpha^{\frac{1}{2}})$ and consequently obtain 
$W_2\big(\mathcal{L}(Y^{(\alpha
)}), \mathcal{L}(Z^{(\alpha
)})\big) \in \mathcal{O}(\alpha^{\frac{1}{4}}),$
which implies Observation~\ref{ob:bridging} and gives an explicit convergence rate. The next two steps are devoted to establishing the convergence of $Z^{(\alpha)}$ as $\alpha \to 0$.

\paragraph{Step 2: Convergence under Additive Gaussian Noise and Rational Stepsize}\label{sec: technique step 1} 

Consider the dynamic \eqref{eq: iid approximate} with  two different stepsizes $\alpha$ and $\alpha'=\alpha/k$, where $k \in \mathbb{N}^+$. Let $\{Z_t^{(\alpha)}\}_{t\geq 0}$ and $\{Z_t^{(\alpha')}\}_{t\geq 0}$ be the corresponding scaled iterates  generated by  \eqref{eq: iid approximate}. The key idea is to couple these two sequences in a way such that one step of $Z_t^{(\alpha)}$ approximately corresponds to  $k$ steps of $Z_t^{(\alpha')}$:
\begin{equation}\label{eq:step2couple}
\begin{aligned}
    Z_{t+1}^{(\alpha)} &= (1-\alpha)Z_t^{(\alpha)} + \sqrt{\alpha}\Big[ \mT(\sqrt{\alpha}Z_t^{(\alpha)} + \theta^*)  - \mT(\theta^*) +  \frac{w_{kt}' + \dots+ w_{kt+k-1}'}{\sqrt{k}}\Big],\\
    Z_{t+1}^{(\alpha')} &= (1-\alpha')Z_t^{(\alpha')} + \sqrt{\alpha'}\big[ \mT(\sqrt{\alpha'}Z_t^{(\alpha')} + \theta^*)  - \mT(\theta^*) +  w_t'\big]. 
\end{aligned}
\end{equation}
Thanks to Gaussianity, the noises ${(w_{kt}' + \dots+ w'_{kt+k-1})}/{\sqrt{k}}$ and $w_t'$ are  identically distributed. Under this coupling, we establish convergence of the squared distance $ \E \| Z_t^{(\alpha)} - Z_{kt}^{(\alpha')} \|_c^2$ under some appropriate norm $\|\cdot\|_c$. Sending $t$ to infinity gives $W_2\big(\mathcal{L}(Z^{(\alpha
)}), \mathcal{L}(Z^{(\alpha')})\big) \in \mathcal{O}(\alpha^{r}),$ for some $r >0.$ 

Next we consider stepsizes $\alpha>0$ and $\alpha/k$ with $k = p/q \in \mathbb{Q}^+,$ where $ p,q \in \mathbb{N}^+$ and $ p>q$. By triangle equality for the $W_2$ metric, we have
\begin{align}
W_2\big(\mathcal{L}(Z^{(\alpha
)}), \mathcal{L}(Z^{(\alpha
/k)})\big) &\leq W_2\big(\mathcal{L}(Z^{(\alpha
)}), \mathcal{L}(Z^{(\alpha
/p)})\big) + W_2\big(\mathcal{L}(Z^{(\alpha/k
)}), \mathcal{L}(Z^{(\alpha/p)})\big) \notag \\
&{\leq} \mathcal{O}(\alpha^{r}) + \mathcal{O}(\alpha^{r}) \in \mathcal{O}(\alpha^{r}).
\label{eq:steady-state-rate-outline}
\end{align}
Therefore, for any rational sequence $\{\alpha_j\}_{j=0}^\infty$ with $\alpha_j \to 0$, the sequence $\{\mathcal{L}(Z^{(\alpha_j)})\}$ is Cauchy in $W_2$ and thus has a limit. Suppose, for contradiction, that there exist two rational sequences $\{\alpha_j\}$ and $\{\beta_j\}$ tending to zero such that their limits $\mathcal{L}(\bar{Y})$ and $\mathcal{L}(\hat{Y})$ differ, i.e., $W_2\big(\mathcal{L}(\bar{Y}), \mathcal{L}(\hat{Y})\big) > \epsilon$ for some $\epsilon > 0$. Construct a new rational sequence $\gamma_j$ by interleaving: $\gamma_{2j} = \alpha_j$, $\gamma_{2j+1} = \beta_j$. Then $\gamma_j \to 0$, and $\{\mathcal{L}(Z^{(\gamma_j)})\}$ should be Cauchy. However,
\begin{align*}
\lim_{j \to \infty} W_2\big(\mathcal{L}(Z^{(\gamma_{2j})}), \mathcal{L}(Z^{(\gamma_{2j+1})})\big) &= \lim_{j \to \infty} W_2\big(\mathcal{L}(Z^{(\alpha_j)}), \mathcal{L}(Z^{(\beta_j)})\big) \geq W_2\big(\mathcal{L}(\bar{Y}), \mathcal{L}(\hat{Y})\big) > \epsilon,
\end{align*}
which contradicts the Cauchy property of $\{\mathcal{L}(Z^{(\gamma_j)})\}$. Therefore, the limit $\mathcal{L}(Y)$ is unique, and we have $\lim_{\alpha \to 0,\, \alpha \in \mathbb{Q}} W_2\big(\mathcal{L}(Z^{(\alpha)}), \mathcal{L}(Y)\big) = 0.$

\paragraph{Step 3: Convergence under Additive Gaussian Noise and General Stepsize}\label{sec: technique step 2}

Continuing with the dynamic \eqref{eq: iid approximate}, we show that $\mathcal{L}(Z^{(\alpha)})$ is continuous in $\alpha$ with respect to  $W_2$. To this end, we consider two real-valued stepsizes $\alpha$ and $\alpha'$, and couple the sequences $\{Z_t^{(\alpha)}\}_{t\geq 0}$ and $\{Z_t^{(\alpha^\prime)}\}_{t\geq 0}$ by letting them share the same noise:
\begin{equation}\label{eq:step3couple}
\begin{aligned}
Z_{t+1}^{(\alpha)} &= (1-\alpha)Z_t^{(\alpha)} + \sqrt{\alpha}\big[ \mT(\sqrt{\alpha}Z_t^{(\alpha)} + \theta^*)  - \mT(\theta^*) +  w_t'\big],\\
    Z_{t+1}^{(\alpha^\prime)} &= (1-\alpha^\prime)Z_t^{(\alpha^\prime)} + \sqrt{\alpha^\prime}\big[ \mT(\sqrt{\alpha'}Z_t^{(\alpha')} + \theta^*)  - \mT(\theta^*) +  w_t'\big].
\end{aligned}
\end{equation}
We again analyze the square distance $ \E \| Z_t^{(\alpha)} - Z_{t}^{(\alpha')} \|_c^2$. Setting $t\to\infty$ followed by $\alpha'\to \alpha$, we establish the continuity property $\lim_{\alpha^\prime \to \alpha}W_2\big(\mathcal{L}(Z^{(\alpha
)}), \mathcal{L}(Z^{(\alpha^\prime)})\big) =0.$ Since $\mathbb{Q}^+$ is dense in $\mathbb{R}^+$, together with the result of Step 1, we obtain
$
\lim_{\alpha \to 0} W_2\big(\mathcal{L}(Z^{(\alpha
)}), \mathcal{L}(Y)\big) = 0,
$
i.e. the steady-state convergence of the dynamic \eqref{eq: iid approximate}.

To obtain an explicit convergence rate of $\mathcal{L}(Z^{(\alpha
)})$ to the limit $\mathcal{L}(Y)$, we observe that
\begin{align*}
W_2\big(\mathcal{L}(Z^{(\alpha
)}), \mathcal{L}(Y)\big) &\leq W_2\big(\mathcal{L}(Z^{(\alpha
)}), \mathcal{L}(Z^{(\alpha/k
)})\big) + W_2\big(\mathcal{L}(Z^{(\alpha/k
)}), \mathcal{L}(Y)\big),\qquad \forall k\in \N^+.
\end{align*}
Letting $k\to\infty$ on both sides and applying the bound~\eqref{eq:steady-state-rate-outline}, we obtain the desired rate:
\begin{align*}
W_2\big(\mathcal{L}(Z^{(\alpha
)}), \mathcal{L}(Y)\big) 
\in \mathcal{O}(\alpha^{r}).
\end{align*}

Therefore, by the above three steps, we have 
\begin{align*}
W_2\big(\mathcal{L}({Y}^{(\alpha)}), \mathcal{L}(Y)\big) \leq W_2\big(\mathcal{L}({Y}^{(\alpha)}), \mathcal{L}(Z^{(\alpha)})\big) + W_2\big(\mathcal{L}(Z^{(\alpha)}), \mathcal{L}(Y)\big) \in \mathcal{O}(\alpha^{\min(r, \frac{1}{4})}),
\end{align*}
which shows that  ${Y}^{(\alpha)}$ for general contractive SA update \eqref{eq: iid general SA2} converges in $W_2$ at a rate $\mathcal{O}(\alpha^{\min(r, \frac{1}{4})})$.

Following the three-step procedure outlined above, much of the technical effort focuses on obtaining tight estimates for square distances of the form $\E \| Y_t^{(\alpha)} - Y_{t'}^{(\alpha')} \|_c^{2}$, with potentially mismatched stepsizes $(\alpha,\alpha')$ and/or time indices $(t,t')$. Achieving this under the nonsmooth SA dynamics requires a careful analysis of the multi-step dynamics and exploitation of the contractive property via a  Moreau envelope argument.

\subsection{Notations}


We write $\mathbb{S}^{d-1} := \{\theta \in \mathbb{R}^d : \|\theta\|_2 = 1\}$ for the unit sphere in $\mathbb{R}^d$, where $\|\cdot\|_2$ denotes the Euclidean norm.
 We use $B^d(\theta, \epsilon)$ to denote an open ball in $\R^d$ centering at $\theta$ with radius $\epsilon>0$ with respect to $\|\cdot\|_2$. A function is $C^k$ if it is $k$ times continuously differentiable. An operator $\mT : \R^d \to \R^d$ is said to be $\gamma$-contractive  w.r.t.\ the norm $\|\cdot\|_c$ if the following holds for some $\gamma\in(0,1)$,
\begin{equation}\label{eq: contraction}
\|\mT(\theta) - \mT(\theta^\prime)\|_c \leq \gamma\|\theta - \theta^\prime\|_c, \quad \forall \theta, \theta^\prime \in \R^d.
\end{equation}
A function $h$ is called $L$-smooth w.r.t.\ some norm $\|\cdot\|$ if $\|\nabla h(x) - \nabla h(y)\|_* \le L\|x-y\|,\forall x,y$, where $\|\cdot\|_*$ is the dual norm of $\|\cdot\|.$ A function $g: \R^m \to \R^d$ is positively homogeneous if $g(cx) = cg(x)$ for all $ c \geq 0$ and $x \in \R^m$.

Let $\mathcal{P}_2(\mathbb{R}^d)$ denote the space of square-integrable distributions on $\mathbb{R}^d$. For a random vector $\theta$, let $\mathcal{L}(\theta)$ denote the distribution of $\theta$ and $\var(\theta)$ its covariance matrix.
The Wasserstein 2-distance between two distributions $\mu$ and $\nu$ in $\mathcal{P}_2(\mathbb{R}^d)$ is defined as 
\begin{align*}
W_2(\mu, \nu) & =\inf _{\xi \in \Pi(\mu, \nu)}\Big(\int_{\mathbb{R}^d}\|u-v\|_c^2 \mathrm{~d} \xi(u, v)\Big)^{\frac{1}{2}} 
=\inf \Big\{\mathbb{E}\big[\|\theta-\theta^{\prime}\|_c^2\big]^{\frac{1}{2}}: \mathcal{L}(\theta)=\mu, \mathcal{L}\left(\theta^{\prime}\right)=\nu\Big\},
\end{align*}
where  $\Pi(\mu, \nu)$ is the set of joint distributions in $\mathcal{P}_2(\mathbb{R}^d \times \mathbb{R}^d)$ with marginal distributions $\mu$ and $\nu$. 

For a finite set $\mS$, we use $\Delta(\mS)$ to denote the probability simplex over $\mS$. Given $\pi \in \Delta(\mS)$,  we denote by $\operatorname{Multinomial}\left(\pi, n\right)$ the multinomial distribution with event probabilities $\pi$ and number of trials $n$. For two random variables/vectors $X$ and $Y$, we use the notation $X\overset{\textup{d}}{=} Y$ to indicate that $X$ and $Y$ have the same distribution.

For two real valued functions $f(x), g(x): \R^+ \to \R$,  we write $f(x) \in o(g(x))$ if $\lim_{x \to 0}\frac{f(x)}{g(x)} = 0$, and we write $f(x) \in \mathcal{O}(g(x))$ if there exist $x_0, M >0$ such that $|f(x)| \leq Mg(x), \forall x \leq x_0$. We say that $f(x)$ is super-polynomial if $f(x) \in o(x^n), \forall n \in \mathbb{N}^+.$

\begin{remark}\label{rem:unifrom_in_t}
We adopt the following convention (unless specified otherwise) to avoid cluttered notations. When we write $f(\alpha,t)\in\mathcal O(\alpha^{n})$, we mean that there exist constants
$\alpha_0>0$ and $M>0$ such that, for every $\alpha\le \alpha_0$, there exists a  threshold $t_{\alpha,n}$ with
\[
f(\alpha,t)\le M\,\alpha^{n},\qquad \forall\, t\ge t_{\alpha,n}.
\]
Similarly, when we write $f(\alpha,t)\in o(\alpha^{n})$, we mean that there exist $\alpha_0>0$
and a sequence $M_\alpha>0$ satisfying $M_\alpha\to 0$ as $\alpha\to 0$, such that, for every
$\alpha\le \alpha_0$, there exists a threshold $t_{\alpha,n}$ with
\[
f(\alpha,t)\le M_\alpha\,\alpha^{n},\qquad \forall\, t\ge t_{\alpha,n}.
\]
See also Remark~\ref{rem:uniform_in_t_appendix}.
\end{remark}

\textbf{Paper organization.}

In Section~\ref{sec:add-noise}, we present the model and the main results.  Section~\ref{sec:syn-q} applies these results to Q-learning algorithms. In Section~\ref{sec: averaging and extrapolation}, we explore the implications of our results for Polyak-Ruppert averaging and Richardson-Romberg extrapolation. Section~\ref{sec:outline} outlines the proofs of our main results. We provide numerical experiments that corroborate our theoretical results in Section~\ref{sec:experiments}, followed by a discussion of additional related work in Section~\ref{sec: related work}.

\section{Nonsmooth Stochastic Approximation with I.I.D\ Noise}
\label{sec:add-noise}

In this section, we consider contractive nonsmooth stochastic approximation with i.i.d\ noise. 

\subsection{Model Setup}\label{sec: model}

We consider the following stochastic approximation iteration with i.i.d\ noise:
\begin{equation}\label{eq: additive raw dynamic}
\theta_{t+1}^{(\alpha)} = \theta_t^{(\alpha)} + \alpha\big(\tmT(x_t, \theta_t^{(\alpha)}) - \theta_t^{(\alpha)}\big),
\end{equation}
where  $\tmT : \R^{d_x}\times \R^d \to \R^d$ is an operator, $\alpha>0$ is a constant stepsize and $\{x_t\}_{t \geq 0}$ is a sequence of i.i.d  noise in  $\R^{d_x}$. We denote by $\mT(\cdot):= \E_{x} [\tmT(x, \cdot)]$ the expectation of the operator $\tmT(x, \cdot )$ with respect to the noise $x$.

Stochastic approximation covers many important iterative algorithms as special cases. For instance, if $\mT(\theta)=-\nabla U(\theta)+\theta$ for some  function $U: \R^d \to \R$ that is twice continuously differentiable, $L$-smooth
and strongly convex, then the update \eqref{eq: additive raw dynamic} corresponds to Stochastic Gradient Descent (SGD) for minimizing $U$ \citep{lan2020first}. If $\mT(\theta)=A\theta +b,$ where $A\in \R^{d\times d}$ is a Hurwitz matrix, then \eqref{eq: additive raw dynamic} becomes Linear SA, which in turn covers the TD-learning algorithm in reinforcement learning \citep{huo2023bias,srikant2019finite}. In both examples, the operator $\mT$ is at least $C^1$-smooth and contractive in $\|\cdot \|_2$ (or its weighted version).

In this work, we consider a more general class of SA algorithms with a potentially nonsmooth operator $\mT$. We only assume that $\mT$ is contractive with respect to some norm.

\begin{assumption}[Contractive $\mT$]\label{as: contraction0}
The operator $\mT : \R^d \to \R^d$ is $\gamma$-contractive for some $\gamma\in(0,1)$ with respect to  some norm $\|\cdot\|_c$. 
\end{assumption}
By Banach fixed point theorem, under Assumption \ref{as: contraction0}, the fixed point equation $\mT(\theta) = \theta$ has a unique solution $\theta^* \in \R^d$. 

We impose on $\tmT$ the following \emph{expected Lipschitz-continuity} assumption (indexed by $n \ge 1$), which is also used in \cite{listochastic}. We remark that this condition is weaker than the almost-sure $L$-co-coercivity assumed in \cite{dieuleveut2020bridging}, as the latter implies almost-sure Lipschitz continuity (up to a multiplicative constant in $L_n$) while our assumption is in expectation.

\begin{assumption}[$\bm{n}$]\label{as:lip}
There exists a constant $L_n >0$ such that 
\[
\E_{x_0}\big[\|\tmT(x_0,\theta)-\tmT(x_0,\theta')\|_c^{2n}\big] \leq L_n\|\theta-\theta'\|_c^{2n}, \quad \forall \theta, \theta' \in \R^d.
\]
\end{assumption}

We consider the following moment assumption for the noise $\{x_t\}_{t \geq 0}$, indexed by $n\geq 1$.

\begin{assumption}[$\bm{n}$]\label{as: additive noise}
The random variables $\tmT(x_0, \theta^*)$ have finite $(2n)$-th moments.
\end{assumption}

Such moment assumptions, for example with $n = 1$ or $2$, are standard in prior work on the analysis of SGD and SA~\citep{dieuleveut2020bridging,kushner2003-yin-sa-book,srikant2019finite}. In general, under Assumption~\ref{as: additive noise}($\bm{n}$), we can control the $2n$-th moment of the contractive SA iterates $\{\theta_t^{(\alpha)}\}_{t \geq 0}$. A similar observation was made for SGD in~\cite{yu2021analysis,dieuleveut2020bridging}.


\subsection{Moments Bounds and Convergence to Stationary Distribution}

In this subsection, we first derive finite-time upper bounds on $\E \|\theta_t^{(\alpha)} - \theta^*\|_c^{2n}$, the $2n$-th moments  of the estimation error, under the matching moment assumptions. Our bounds generalize the mean-square error (MSE) bound in prior work in~\cite{chen2020finite, chen2023lyapunov} to higher moments, and extend the results on smooth SGD by~\cite{dieuleveut2020bridging} to nonsmooth SA. 

\begin{proposition}[Moment Bounds]\label{thm: additive 2n moment}
For each integer $n \geq 1$, under Assumptions~\ref{as: contraction0}, \ref{as:lip}(\textbf{n}) and \ref{as: additive noise}(\textbf{n}), there exists $\bar{\alpha}_n>0$ for which the following holds: for any $\alpha \leq \bar{\alpha}_n$, there exists $t_{\alpha,n} > 0$ such that 
\begin{equation}\label{eq: additive 2n moment}
\E[\|\theta_t^{(\alpha)} - \theta^*\|_c^{2n}] \leq c_n\E[\|\theta^{(\alpha)}_{t_{\alpha,n}} - \theta^*\|_c^{2n}](1- \alpha(1 - \sqrt{\gamma}))^{t-t_{\alpha,n}} + c_n^\prime \alpha^n, \quad \forall t \geq t_{\alpha,n},
\end{equation}
where $c_n$ and $c_n^\prime$ are constants that are independent of $\alpha$ and $t$. Moreover, $t_{\alpha,1} = 0.$
\end{proposition}

Proposition \ref{thm: additive 2n moment} indicates that the bound on the 2$n$-th moments of the estimation error decays geometrically fast, up to a term of the order $\mathcal{O}(\alpha^n).$ For fixed $\alpha>0$, when $t$ is sufficiently large, Proposition~\ref{thm: additive 2n moment} ensures $\E[\|\theta_t^{(\alpha)}-\theta^*\|_c^{2n}] \in \mathcal{O}(\alpha^n)$. The forthcoming main theorems analyze two regimes: 1) $t\to\infty$ with $\alpha$ held fixed (Theorem \ref{thm: additive convergence}), and 2) $t\to\infty$ with $\alpha$ fixed, followed by $\alpha\to 0$ (Theorems \ref{thm: additive limit} and \ref{thm: additive smooth}). Although the threshold $t_{\alpha,n}$ depends on $\alpha$, Proposition~\ref{thm: additive 2n moment} suffices in both regimes as the bounds are used under a fixed $\alpha$. 

In subsequent analysis, we mostly use Assumptions~\ref{as:lip}($\bm{n}$), \ref{as: additive noise}($\bm{n}$) and Proposition~\ref{thm: additive 2n moment} with $n\in\{1,2\}$. 
In particular, Proposition~\ref{thm: additive 2n moment} with $n=1$ provides a finite-time MSE bound. Building on this bound, we can establish our first main theorem, which proves the weak convergence of the stochastic process $\{\theta_t^{(\alpha)}\}_{t \geq 0}$  to a unique stationary distribution in $W_2$; moreover, we characterize its geometric convergence rate.

\begin{theorem}[Distributional Convergence]\label{thm: additive convergence}
Under Assumptions~\ref{as: contraction0}, \ref{as:lip}($\bm{1}$) and \ref{as: additive noise}($\bm{1}$), there exists $\bar{\alpha}^\prime>0$ such that for any stepsize $ \alpha \leq \bar{\alpha}^\prime$ and any initial distribution of $\theta_0^{(\alpha)}$, the sequence $\{ \theta_t^{(\alpha)}\}_{t \geq 0}$ converges geometrically in $ W_2$ to a random variable $\theta^{(\alpha)}$ with
\[W^2_2\big(\mathcal{L}(\theta_t^{(\alpha)}), \mathcal{L}(\theta^{(\alpha)}) \big) \leq c' \cdot (1-\alpha(1-\sqrt{\gamma}) )^t, \quad \forall t \geq 0,\]
where $c'$ is a constant that is independent of $\alpha$ and $t$. Moreover, $\E[\|\theta^{(\alpha)} - \theta^*\|_2^2] \in \mathcal{O}(\alpha).$
\end{theorem}

A key step in proving Proposition~\ref{thm: additive 2n moment} and Theorem \ref{thm: additive convergence} is to construct a proper \emph{smooth} Lyapunov function for the nonsmooth dynamics. Previous work on moment bounds and $W_2$ convergence focuses on linear SA and smooth SGD \citep{dieuleveut2020bridging, huo2023bias}. These dynamics are smooth and contractive in the $\ell_2$ norm $\|\cdot\|_c = \|\cdot\|_2$, the square of which can be used as a smooth Lyapunov function. However, for general contractive SA, the norm $\|\cdot\|_c$ may be non-differentiable, e.g.,  $\|\cdot\|_{\infty}$. We instead make use of the  Moreau envelope \citep{parikh2014proximal} of $\|\cdot\|_c$, a technique that has been used to study the MSE (i.e., $n=1$) of contractive SA \citep{chen2020finite, chen2023lyapunov}. To further establish the weak convergence result in Theorem~\ref{thm: additive convergence}, we develop a novel coupling argument using the Moreau envelope, going beyond the $\ell_2$ norm-based analysis in prior work~\citep{huo2023bias,dieuleveut2020bridging}.
We outline the proofs of Proposition~\ref{thm: additive 2n moment} and Theorem~\ref{thm: additive convergence} in Section~\ref{sec:outline}, and provide the full proofs in Sections~\ref{sec: addtive 2n moment} and~\ref{sec: additive convergence}, respectively.

\subsection{Steady-State Convergence and Bias Characterization}\label{sec:results_additive_steady_bias}

Recall the SA iterates $\{ \beta_t^{(\alpha)}\}_{t \geq 0}$  with additive additive Gaussian noise~\eqref{eq: iid approximate}, which is a special case of the general SA update~\eqref{eq: iid general SA2}. Consequently, Theorem \ref{thm: additive convergence} implies that $\{ \beta_t^{(\alpha)}\}_{t \geq 0}$ also converges to a random variable $\beta^{(\alpha)}.$ 
Recall that $Y^{(\alpha)} = \frac{\theta^{(\alpha)}-\theta^*}{\sqrt{\alpha}}$ and $Z^{(\alpha)}=\frac{\beta^{(\alpha)}-\theta^*}{\sqrt{\alpha}}$ are the scaled versions of the steady-state random variables $\theta^{(\alpha)}$ and $\beta^{(\alpha)}$ for the general SA~\eqref{eq: iid general SA2} and Gaussian version \eqref{eq: iid approximate}, respectively; see Section \ref{sec: technique}.
In the following proposition, we formalize Observation \ref{ob:bridging} and show that $Y^{(\alpha)}$ and $Z^{(\alpha)}$ have the same limit as $\alpha\rightarrow 0.$

\begin{proposition}[Universality of Steady-State Convergence]\label{prop:gaussian} Under Assumptions \ref{as: contraction0}, \ref{as:lip}($\bm{2}$) and \ref{as: additive noise}($\bm{2}$), we have
$
W_{2}\big(\law(Y^{(\alpha)}),\law(Z^{(\alpha)})\big)\in \mathcal{O}(\alpha^{1/4}),
$
Consequently, if $\lim_{\alpha \to 0}W_{2}\big(\law(Z^{(\alpha)}),\law(Y)\big) = 0$ for some random variable $Y$, then  $\lim_{\alpha \to 0}W_{2}\big(\law(Y^{(\alpha)}),\law(Y)\big) = 0$.

\end{proposition}

Thanks to Proposition \ref{prop:gaussian}, in order to study the  convergence of the steady-state $Y^{(\alpha)}$, it suffices to analyze the SA with additive Gaussian noise \eqref{eq: iid approximate}. 

Below we state our steady-state convergence result and characterize the bias at steady-state. To this end, we focus on a broad class of potentially nonsmooth SA dynamics~\eqref{eq: additive raw dynamic} where 
the expected operator $\mT$ is only assumed to be one-sided directionally differentiable at $\theta^*$ \citep{rockafellar1998variational}. 




\begin{assumption}[Nonsmooth Class]\label{as: contraction}
The operator $\mT$ is one-sided direactionally differentiable at its fixed point $\theta^*$ in the following sense: there exists a function $G$: $\mathbb{S}^{d-1}\to \R^d$ such that
\begin{align*}
    \lim_{w \to 0^+} \Big\|\frac{\mT(\theta^* + w\theta)-\mT( \theta^*)}{w} - G(\theta)\Big\|_2 = 0, \quad \forall \theta \in \mathbb{S}^{d-1}.
\end{align*}
The function \(G\) is called the one‑sided directional derivative of \(\mT\) at \(\theta^*\).
\end{assumption}

Note that $\mT$ is differentiable at $\theta^*$ if and only if there exists a (Jacobian) matrix $J$ such that $G(\theta) = J\theta$ for every  $\theta \in \mathbb{S}^{d-1}$; see Lemma \ref{usefullemma2}. In particular, $\mT$ is non-differentiable at $\theta^*$ when $G(\theta) \neq -G(-\theta)$ for some $\theta \in \mathbb{S}^{d-1}$, which highlights the nonsmoothness of the example in \eqref{eq: simple example} discussed in Section \ref{sec: challenge}.

The class of {one-sided directionally differentiable} mappings is rich.  
It strictly contains several widely studied nonsmooth subclasses—including 
\(g\!\circ\!F\) composite functions~\citep{shapiro2003class,sagastizabal2013composite}, 
prox–regular functions~\citep{poliquin1996prox}, and Clarke–regular functions~\citep{clarke1990optimization,hiriart2004fundamentals}.  
Moreover, it covers classical max-type functions, spectral operators such as the largest-eigenvalue map, 
the sparsity-promoting \(\ell_1\)-norm, and any of their compositions with smooth transformations. 


In the context of SA, Assumption~\ref{as: contraction} allows for any $\mT$ that is locally $C^1$ and contractive, hence covering strongly convex SGD and Linear SA discussed in Section \ref{sec: model}.
In addition, our model covers operators $\mT$ that are not differentiable at $\theta^*$, e.g., the example in~\eqref{eq: simple example} with $b = 0$ (corresponding to $G(\theta) = -{\operatorname{sign}(\theta)}/{2}$), as well as the Bellman optimality operator underlying various Q-learning algorithms (see Section~\ref{sec:syn-q}). We note that an arbitrary contractive operator might not satisfy Assumption~\ref{as: contraction}; see the example in Lemma~\ref{lem:counterexample}.  Extending the steady-state convergence analysis to an even broader nonsmooth class beyond Assumption~\ref{as: contraction} is an interesting direction for future research.

Theorem \ref{thm: additive convergence} implies that the centered and rescaled iterate $Y_t^{(\alpha)} = \frac{\theta_t^{(\alpha)} - \theta^*}{\sqrt{\alpha}}$ converges weakly to a steady-state random variable $Y^{(\alpha)}:= \frac{\theta^{(\alpha)} - \theta^*}{\sqrt{\alpha}}$ as $t\to\infty$. Focusing on SA satisfying the $g\circ F$ decomposability Assumption~\ref{as: contraction}, our next theorem establishes steady-state convergence, i.e., the convergence of $\{Y^{(\alpha)}\}_{\alpha \in (0,\bar{\alpha}^\prime)}$ as $\alpha \to 0$.

\begin{theorem}[Steady-State Convergence]\label{thm: additive limit}
Suppose Assumptions \ref{as: contraction0}, \ref{as:lip}($\bm{2}$), \ref{as: additive noise}($\bm{2}$), and \ref{as: contraction} hold. Then there exists a unique random variable $Y$, depending only on $\mT$ and $\var(\tmT(x_0, \theta^*))$, such that 
\[\lim_{\alpha \to 0}W_2\big(\mathcal{L}(Y^{(\alpha
)}), \mathcal{L}(Y)\big) = 0.\]
Consequently, we have 
\begin{equation}
\label{eq: additive bias}
\E[\theta^{(\alpha)}] = \theta^* + \sqrt{\alpha}\E[Y] + o(\sqrt{\alpha}).
\end{equation}
\end{theorem}

Theorem~\ref{thm: additive limit} implies that the steady-state bias, $\E[\theta^{(\alpha)}] - \theta^*$, is generally on the order of $\Theta(\sqrt{\alpha})$ for small stepsizes $\alpha$. This stands in sharp contrast to  smooth SA, which has an order-wise smaller, $\Theta(\alpha)$ bias. This $\sqrt{\alpha}$-bias property, which arises precisely due to the nonsmoothness of the SA dynamic, is further characterized in our next theorem.
We remark that Theorem \ref{thm: additive limit} provides a universality result: the limit $Y$ depends on the noise $\{x_t\}_{t\ge0}$ only through $\operatorname{Var}(\tmT(x_0, \theta^*))$, the variance of the induced stochastic operator at $\theta^*,$ and is independent of the noise distribution.

Note that Theorem~\ref{thm: additive limit} applies to any contractive SA satisfying the one-sided directionally differentiable condition. In this general setting, the convergence result for $\{Y^{(\alpha)}\}_{\alpha \in (0,\bar{\alpha}^\prime)}$ in Theorem~\ref{thm: additive limit} is asymptotic. The actual convergence rate—in particular, the order of the \(o(\sqrt{\alpha})\) remainder—depends on how quickly the term $\sup_{\theta \in \mathbb{S}^{d-1}}\|\frac{\mT(w\theta+\theta^*)-\mT( \theta^*)}{w} - G(\theta)\|$ 
in Assumption~\ref{as: contraction} vanishes when $w \to 0^+$; see equation~\eqref{eq:T13exp_nin_ball} for the precise relationship.  Our analytical technique enables one to obtain explicit, non-asymptotic bounds on the convergence rate for specific SA dynamics and $\mT$ operators. For instance, in the next section, we establish a  convergence rate of $\mathcal{O}(\alpha^\frac{1}{4})$ for Q-learning algorithms.

The work in \cite{chen2022stationary} also provides a steady-state convergence result, but under a restrictive uniqueness assumption that is difficult to verify in most cases. Besides that, the paper only considers the additive noise setting and it is unclear how to generalize their analysis to the multiplicative noise case. Our results are established using a different technique, by directly proving the weak convergence of $Y^{(\alpha)}$ in $W_2$ using prelimit coupling. 
Section \ref{sec: technique} outlines the key steps of the proof of Theorem \ref{thm: additive limit}. We defer the complete proof to Section \ref{sec: additive limit}.

The next theorem offers a finer characterization of
\(\mathbb{E}[Y]\), the coefficient of the leading
\(\sqrt{\alpha}\) term in the steady-state bias of
\eqref{eq: additive bias}.
It shows that this bias is governed by the
following \emph{positively homogeneous extension} of $G$:
\begin{equation}\label{eq:h}
H(\theta) := \begin{cases}
\|\theta\|G(\theta/\|\theta\|), & \text{if } \theta \neq 0, \\
0, & \text{otherwise},
\end{cases}
\end{equation}
which lifts \(G\) from the unit sphere to the whole space $\R^d$ while preserving
positive homogeneity.

\begin{theorem}[Bias Characterization]\label{thm: additive smooth}
Under the same setting as in Theorem \ref{thm: additive limit}, we have
\begin{enumerate}
    \item If $\mT$ is differentiable at $\theta^*$, which is equivalent to the existence of a matrix $J\in\mathbb R^{d\times d}$ such that $H(\theta) = J\theta$, then $\E[Y]=0$.
    \item If
\(\operatorname{Var} \bigl(\tilde{\mathcal T}(x_{0},\theta^{*})\bigr)\)
is positive definite and there exists an index
\(i\in\{1,\ldots,d\}\) such that the $i$-th coordinate
\(H_{i}\) of  $H$ admits a
\emph{nonsingleton} sub\-differential (or superdifferential)
at $\theta = 0$, then $\E[Y]\neq 0$.
\end{enumerate}
\end{theorem}

Roughly speaking, the condition in Part (2) of Theorem \ref{thm: additive smooth} says that $\mT$ is not differentiable at $\theta^*$ and $H$ is not a linear mapping in $\R^d$. In this case, provided that the noise at $\theta^*$: $w_0(\theta^*)= \tmT(x_0, \theta^*)$ is non-degenerate, we have $\E[Y]\neq 0$. Consequently, equation~\eqref{eq: additive bias} implies that the steady-state bias is on the order of $\Theta(\sqrt{\alpha})$. We conjecture that this result holds under more general settings of nonsmooth $\mT$ where the sub/supdifferential of $H(\theta)$ at $\theta=0$ may not exist, which we leave as future work. This $\sqrt{\alpha}$-order of the bias has important implications for bias reduction via the Richardson-Romberg extrapolation, which we will discuss in details in Section~\ref{sec: averaging and extrapolation}.

Part (1) of Theorem \ref{thm: additive smooth}, on the other hand, implies that for any smooth SA where $\mT$ is continuously differentiable at $\theta^*$,  the asymptotic bias is order-wise smaller than $\sqrt{\alpha}$. This result is consistent with prior work in \cite{dieuleveut2020bridging, huo2023bias}, which show that the asymptotic biases of smooth SGD and Linear SA with i.i.d.\ noise are of order $\Theta(\alpha)$ and $0$, respectively. 

Recall that the simple example \eqref{eq: simple example} in Section \ref{sec: challenge} is a SA with the expected operator $\mT(\theta) = -\frac{1}{2}|\theta| -b$. It is easy to verify that $\theta^* = \max(2b, 2b/3)$ in this example. Therefore, when $b \neq 0$, $\mT(\theta)$ is locally linear around $\theta^*$, thereby implying the asymptotic bias is order wise smaller than $\sqrt{\alpha}$. When $b = 0$, $ H(\theta)=\frac{1}{2}|\theta|$, whose subdifferential is not singleton at $\theta^* = 0$. Hence, for the case $b = 0$, the asymptotic bias is on the order of $\Theta(\sqrt{\alpha})$.

We outline the proof of Theorem \ref{thm: additive smooth} in Section \ref{sec:prf_outline_thm_additive_smooth}. The full proof is provided in Section \ref{sec: prf_additive smooth}.

\section{Application in Q-learning}
\label{sec:syn-q}
In this section, we specialize the general findings of Section~\ref{sec:add-noise} to Q-learning, a prototypical example of a nonsmooth SA driven by i.i.d.\ multiplicative noise.
 Going beyond the results in Section \ref{sec:add-noise}, we also provide an explicit convergence rate of steady-state convergence and establish a fine-grained bias characterization for Q-learning---as a demonstration of our  technique for more precise analysis on specific SA dynamics. 

\subsection{Model Setup}
Consider a discounted Markov decision
process (MDP) defined by the tuple $(\mS,\mA, P, r,\gamma),$ where $\mS$ and $\mA$ are respectively the finite
state and action spaces,   $P:\mS\times \mA \rightarrow \Delta(\mS)$ is the transition kernel,
$r:\mS\times \mA \rightarrow \R$ is the stochastic reward function, and $\gamma\in(0,1)$ is the discount factor. Given a policy $\pi:\mS\rightarrow\Delta(\mA),$ the Q-function $q^{\pi}:\mS\times \mA \rightarrow \R$  is defined as $q^{\pi}(s,a)=\E_{\pi}\big[ \sum_{k=0}^{\infty} \gamma^k r_k (s_k,a_k) \mid s_0 = s, a_0 = a\big]$, where $a_k \sim \pi(\cdot|s_k), s_{k+1}\sim P(\cdot|s_k,a_k) $ and $r_k$  is an independent copy of $r$. The goal is to learn an optimal policy $\pi^*$ that maximizes $q^{\pi}$ without the knowledge of $P$ and $r$. Below we often view $P$ as an $|\mS||\mA|$-by-$|\mA|$ matrix,  $r$ a random vector in $\R^{|\mS||\mA|}$, and $q^\pi$ a vector in $\R^{|\mS||\mA|}$.

\textbf{Q-learning}~\citep{Watkins92-QLearning} is a  class of reinforcement learning methods that use data to iteratively approximate the optimal Q-function $q^*\equiv q^{\pi^*},$ from which one can recover the optimal policy as $\pi^*(s)\in \argmax_{a\in \mA} q^*(s,a)$. We consider a general form of Q-learning that iteratively generates a sequence of Q-function estimates, $\{q_t:\mS\times\mA \rightarrow \R\}_{t \geq 0},$  according to the following recursion:
\begin{equation}\label{eq: theta}
q_{t+1}^{(\alpha)} = q_t^{(\alpha)} + \alpha D_t \big(\gamma P_t f(q_t^{(\alpha)}) - q_t^{(\alpha)}+ r_t\big),
\end{equation}
where the function $f: \R^{|\mS||\mA|} \to \R^{|\mS|}$ is given by 
\[f_s(q)  :=  \max_{a \in \mA}q(s,a), \quad \forall s \in \mS,\]
and $\{(D_t,P_t,r_t)\}_{t \geq 0}$ are i.i.d.\ random matrices/vectors satisfying the following: (i) $D=\E[D_0]$ is a $|\mS||\mA|$-by-$|\mS||\mA|$ diagonal matrix with $D_{ii} \in (0,1], \forall i \in \mS \times \mA$; (ii) $\E[P_0] = P$; (iii) $\{r_t\}_{t \geq 0}$ are independent copies of $r$; (iv) $D_0, P_0$ and $r_0$ are independent with each other. Here $D_t, P_t$ and $r_t$ correspond to the empirical state-action distribution, empirical transition and empirical reward function, respectively, observed at the $t$-th iteration. 

We discuss two important special cases of the above model.
\begin{itemize}[leftmargin=*]
    \item \textbf{Synchronous Q-learning \citep{wainwright2019q}:} At each time step $t$, for each state-action pair $(s,a)$, we observe a reward $r_t(s,a)\overset{\textup{d}}{=}r(s,a)$  and a next state $x_t(s,a)$ drawn  from the transition kernel $P(\cdot|s,a)$. The Q-function estimates are updated as 
    \begin{align*}
        q_{t+1}^{(\alpha)} (s,a) = q_t^{(\alpha)} (s,a) + \alpha \big(\gamma \max_{a'\in\mA} q_t^{(\alpha)}(x_t(s,a),a') -q_t^{(\alpha)} (s,a) + r_t(s,a) \big), \;\; \forall (s,a) \in \mS \times \mA.
    \end{align*}
    Synchronous Q-learning corresponds to the update rule \eqref{eq: theta} where $D_t \equiv I$ and $P_t$ is a binary random matrix where $(s,a)$-th row is independently distributed as $\operatorname{Multinomial}\left(P\left(\cdot|s,a\right), 1\right)$. 
    
    \item \textbf{Asynchronous Q-learning~\citep{chen2023concentration}:} At each time step $t$, we observe a state-action pair $(s_t,a_t) \sim \kappa_b,$ where the distribution $\kappa_b \in \Delta(\mS\times\mA)$ can be the stationary state-action distribution of some behavior policy. Conditioned on $(s_t,a_t)$, we  observe the reward $r_t(s_t,a_t)\overset{\textup{d}}{=} r(s_t,a_t)$ and the next state $s'_{t+1}$ drawn according to $P(\cdot|s_t,a_t)$.  The Q-function estimates are updated as 
    \begin{align*}
        q_{t+1}^{(\alpha)} (s_t,a_t) &= q_t^{(\alpha)} (s_t,a_t) + \alpha \big(\gamma \max_{a'\in\mA} q_t^{(\alpha)}(s'_{t+1},a') -q_t^{(\alpha)} (s_t,a_t) + r_t(s_t,a_t) \big), \\
        q_{t+1}^{(\alpha)}(s,a) &= q_{t}^{(\alpha)}(s,a), \;\; \forall (s,a)\neq (s_t,a_t).
    \end{align*}
    Asynchronous Q-learning corresponds to the update rule \eqref{eq: theta} with $\operatorname{diag}(D_t) \sim \operatorname{Multinomial}\left(\kappa_b, 1\right)$ and the same $P_t$ before. Note that only the $(s_t,a_t)$ entry of $q_t^{(\alpha)}$ is updated at iteration $t$, with $D_t$ acting as the corresponding mask matrix.
\end{itemize}

With other choices of $(D_t, P_t,r_t)$, the update rule~\eqref{eq: theta} can capture other forms of Q-learning with different sampling models.
The Q-learning update~\eqref{eq: theta} can be cast as contractive SA. To this end, we define a random operator $\tmT$
as
\begin{equation}\label{eq:tmT-q}
\tmT(\{D_0,P_0,r_0\},q) = \gamma D_0P_0f(q) +(I-D_0)q +D_0 r_0, \qquad \forall q\in \R^{|\mS||\mA|}.
\end{equation}
Denote by $\mT: \R^{|\mS||\mA|} \to \R^{|\mS||\mA|}$ the expected operator, where 
$$\mT(q):= \E_{\{D_0,P_0,r_0\}}\big[\tmT(\{D_0,P_0,r_0\},q)\big] 
= \gamma DPf(q)+(I-D)q + D \bar{r},$$
with $\bar{r} := \mathbb{E}[r_0]$.
It can be verified that $\mT$ is a $\gamma_0$-contractive operator with respect to the infinity norm $\|\cdot\|_\infty$, where $\gamma_0 = 1-(1-\gamma)\min_{i \in \mS \times \mA}D_{ii}\in (0,1)$ \cite[Proposition 3.3]{chen2023concentration}.  Moreover, the optimal Q-function $q^*$ is the unique solution to the fixed point equation ${\mT}(q^*)=q^*$, which can be viewed as equivalent to the optimal Bellman equation. To be consistent with the general setting, below we use $\|\cdot\|_c$ to denote $\|\cdot\|_\infty$. 

With the above notations,  the Q-learning update \eqref{eq: theta} can rewritten as a contractive SA iteration:
\begin{align}\label{eq:q_learning_SA}
q_{t+1}^{(\alpha)} &= q_t^{(\alpha)} + \alpha\big( \tmT  (\{D_t,P_t,r_t\} , q_t^{(\alpha)} )-q_t^{(\alpha)}\big).
\end{align}
Note that the iteration \eqref{eq:q_learning_SA} is nonsmooth due to the max operation in the function $f$ in~\eqref{eq: theta}; moreover, it involves \emph{multiplicative} noise due to multiplication with the random matrices $D_t$ and $P_t$, which are the stochastic versions of $D$ and $P$.

For the noise we consider the following moment assumption, indexed by an integer $n \geq 1$:
\begin{assumption}[$\bm{n}$]\label{as: q noise}
The random variables $\{(D_t, P_t, r_t)\}_{t \geq 0}$ have finite $(2n)$-th moments.
\end{assumption}


\subsection{Moments Bounds and Convergence to Stationary Distribution}

Under Assumption \ref{as: q noise}(\textbf{n}),  we can verify that the Q-learning algorithm \eqref{eq:q_learning_SA} is a nonsmooth SA satisfying Assumptions~\ref{as:lip}(\textbf{n}), \ref{as: additive noise}(\textbf{n}) and \ref{as: contraction}.

\begin{proposition}\label{prop:q}
Under Assumption \ref{as: q noise}(\textbf{n}), $\tmT$ defined in equation~\eqref{eq:tmT-q} satisfies Assumptions~\ref{as:lip}(\textbf{n}), \ref{as: additive noise}(\textbf{n}) and \ref{as: contraction}. 
\end{proposition}
\begin{proof}{Proof of Proposition \ref{prop:q}}
Because \(D_0\), \(P_0\), and \(r_0\) are mutually independent, it follows that
$
\E_{\{D_0,P_0,r_0\}}\big[\|\tmT(\{D_0,P_0,r_0\},q^*)\|_c^{2n}\big] < \infty,
$
which verifies that Assumption \ref{as: additive noise}(\textbf{n}) holds.
We have
\begin{align*}
&\E_{\{D_0,P_0,r_0\}}\big[\|\tmT(\{D_0,P_0,r_0\},q)-\tmT(\{D_0,P_0,r_0\},q')\|_c^{2n}\big]\\
\leq& 2^{2n-1}\Big(\E_{\{D_0,P_0,r_0\}}\big[\|\gamma D_0P_0(f(q)-f(q'))\|_c^{2n}\big]+\E_{\{D_0,P_0,r_0\}}\big[\| (I-D_0)(q-q')\|_c^{2n}\big]\Big)\\
\leq& 2^{2n-1}\gamma^{2n} \E_{\{D_0,P_0,r_0\}}\big[\| D_0P_0 \|_c^{2n}+\| (I-D_0) \|_c^{2n}\big] \|q-q'\|_c^{2n},
\end{align*}
where the last inequality holds because \(\|f(q) - f(q')\|_c \leq \|q - q'\|_c\). This verifies that Assumption \ref{as:lip}(\textbf{n}) holds with $L_n = 2^{2n-1}\gamma^{2n} \E_{\{D_0,P_0,r_0\}}\big[\| D_0P_0 \|_c^{2n}+\| (I-D_0) \|_c^{2n}\big]$.

To verify that Assumption \ref{as: contraction} holds, recall that $\mA^*(s): = \argmax_{a \in \mA}q^*(s,a)$ is the set of optimal actions for state $s$, and define $g_s(q) := \max_{a \in \mA^*(s)} q(s,a)$ and $g(q):\R^{|\mS||\mA|}\to \R^{|\mS|}$ such that $g(q)[s] = g_s(q), \forall s \in \mS.$
Since $\mT(q^*)  = q^*$, we obtain $\mT(q) = \gamma DPf(q)+(I-D)q + D \bar{r}= q^* +\gamma DP(f(q)-f(q^*))+(I-D)(q-q^*).$ Then, we define the function $\Delta:\mS \to \mathbb{R}$ as $\Delta(s) := \infty$ if $\mA = \mA^*(s)$ and $\Delta(s):=
\max_{a \in \mA} q^*(s,a) -\max_{a \in \mA \setminus \mA^*(s)}q^*(s,a)$ if $\mA \neq \mA^*(s)$,
and let $\Delta_{\min} = \min_{s \in \mS}\Delta(s)$. Because $\Delta(s) >0, \forall s \in \mS$, we have $\Delta_{\min}>0.$ Therefore, when $\|q-q^*\|_c \leq \frac{\Delta_{\min}}{2},$
\begin{align}
\mT(q) = q^* +\gamma DPg(q-q^*)+(I-D)(q-q^*) \label{eq:q-ph1},
\end{align}
which verifies that Assumption \ref{as: contraction} holds with $G(\theta) = \gamma DPg(\theta) + (I-D)\theta$ for every $\theta \in \mathbb{S}^{d-1}$ because \(g\) is positively homogeneous of degree 1.
\end{proof}

Consequently, we derive the following moments bounds and distributional convergence results as immediate corollaries of Proposition~\ref{thm: additive 2n moment} and Theorem~\ref{thm: additive convergence}.

\begin{corollary}[Moment Bounds]\label{thm: q 2n moment}
For each integer $n \geq 1$, under Assumption \ref{as: q noise}(\textbf{n}), there exists $\alpha_n>0$ such that for any $\alpha \leq \alpha_n$, there exists $t_{\alpha,n} \ge 0$ such that
\begin{equation}\label{eq: q 2n moment}
\E[\|q_t^{(\alpha)} - q^*\|_c^{2n}] \leq c_n\E[\|q^{(\alpha)}_{t_{\alpha,n}} - q^*\|_c^{2n}](1- \alpha(1 - \sqrt{\gamma_0}))^{t-t_{\alpha,n}} + c_n^\prime \alpha^n, \quad t \geq t_{\alpha,n},
\end{equation}
where $c_n$ and $c_n^\prime$ are  constants that are independent of $\alpha$ and $t$. Moreover, $t_{\alpha,1} = 0.$
\end{corollary}


\begin{corollary}[Distributional Convergence]\label{thm: q convergence} Under Assumption \ref{as: q noise}($\bm{1}$), there exists $\bar{\alpha}^\prime_0 > 0$ such that for $\forall \alpha \leq \bar{\alpha}^\prime_0$ and all initial distribution of $q_0^{(\alpha)}$, the sequence
    $\{ q_t^{(\alpha)}\}_{t \geq 0}$ converges geometrically fast in $ W_2$ to  
    a random variable $q^{(\alpha)}$ with
\[W^2_2\big(\mathcal{L}(q_t^{(\alpha)}), \mathcal{L}(q^{(\alpha)})\big) \leq c' \cdot (1-\alpha(1-\sqrt{\gamma_0}) )^t, \quad \forall t \geq 0,\]
where $c'$ is a constant independent of $\alpha$ and $t$. Moreover, $\E[\|q^{(\alpha)} - q^*\|_2^2] \in \mathcal{O}(\alpha).$
\end{corollary}


\subsection{Steady-State Convergence and Bias Characterization}

Consider the centered and rescaled iterate $Y_t^{(\alpha)} ={(q_t^{(\alpha)} - q^*)}/{\sqrt{\alpha}}$. Corollary \ref{thm: q convergence} implies that the sequence $\{Y_t^{(\alpha)}\}_{t\geq 0}$ converges weakly to a steady-state random variable $Y^{(\alpha)} = {(q^{(\alpha)} - q^*)}/{\sqrt{\alpha}}$. In the following theorem, we establish the steady-state convergence for $\{Y^{(\alpha)}\}$ as $\alpha\rightarrow 0$.
\begin{theorem}
[Steady-State Convergence]\label{thm: q limit}
Suppose Assumption \ref{as: q noise}($\bm{2}$) holds. There exists a unique random variable $Y$, depending only  on $\mT$ and $\operatorname{Var}(\tmT(\{D_0, P_0, r_0\}, q^*))$. such that 
\[\lim_{\alpha \to 0}W_2\big(\mathcal{L}(Y^{(\alpha
)}), \mathcal{L}(Y)\big) = 0.\]
Furthermore, we have $W_2\big(\mathcal{L}(Y^{(\alpha
)}), \mathcal{L}(Y)\big) \in \mathcal{O}(\alpha^\frac{1}{4})$, which implies that
\begin{align}\label{eq: q bias}
 \E[q^{(\alpha)}] = q^* + \sqrt{\alpha}\E[Y] + \mathcal{O}(\alpha^\frac{3}{4}).   
\end{align}
\end{theorem}

A few remarks are in order. Similar to the general SA setting, Theorem~\ref{thm: q limit} indicates that the steady-state bias of Q-learning, $\E[q^{(\alpha)}] -q^*$, is in general of order $O(\sqrt{\alpha})$ for small stepsize $\alpha.$ Again, this distinctive $\sqrt{\alpha}$-bias result is due to  the nonsmooth nature of the Q-learning dynamic; cf.\ equation \eqref{eq: theta}. Our next theorem provides a more precise characterization on the bias. 
Moreover, as a byproduct of our prelimit coupling, for the explicit Q-learning dynamic, we can obtain an $\mathcal{O}(\alpha^\frac{1}{4})$ convergence rate of $Y^{(\alpha)}$ to the limit $Y$.  The proof of Theorem \ref{thm: q limit} is provided in Section \ref{sec: q limit}.


To discuss the further properties of the limit $Y$,  we first introduce some definitions. 
We say that a state $s^\prime \in\mS$ is \textbf{rooted} if 
\[P\left(s^\prime | s, a\right) = 0, \quad \forall (s, a) \in \mS \times \mA.\]
Intuitively, a state $s'$ is rooted if it is not accessible from any other state in the MDP. Using the optimal Q-function $q^*$, we define $\mA^*(s):= \argmax_{a \in \mA}q^*(s,a)$ as the optimal action set for each state $s\in\mS$.  Note that the action distribution $\pi^*(\cdot |s)$ of the optimal policy is supported on the set 
$\mA^*(s)$ for each $s\in \mS.$ 
We say that a state $s\in\mS$ is \textbf{tied} if $|\mA^*(s)| > 1$, i.e., there are multiple optimal actions for $s$.

We classify all MDPs into two types:  
\begin{itemize}[leftmargin=*]
    \item \textbf{Type A}: There exists at least one state that is tied and not rooted. 
    \item \textbf{Type B} (i.e., not Type A): There is no tied state, or all tied states are rooted. 
\end{itemize}  

For each type of MDP, the following theorem provides a finer  characterization of 
the expectation of the limit $Y$. Recall that
$\E[Y]$ determines the order of the steady-state bias by equation \eqref{eq: q bias}.
\begin{theorem}
[Bias Characterization]\label{thm: q EY}
Under the same setting as Theorem \ref{thm: q limit}, we have
\begin{enumerate}
\item $\E[Y] \neq 0$ if the MDP is in Type A and $\var(\tmT(q^*,\{D_0,P_0,r_0\}))$ is positive definite.
\item $\E[Y] = 0$ if the  MDP is in Type B.
    \item If the MDP is in Type B and Assumption \ref{as: q noise}($\bm{n}$) holds for $n \geq 2$, then $\E[q^{(\alpha)}] = q^* + \mathcal{O}(\alpha^n).$
\end{enumerate} 
\end{theorem}

Note that for a Type-A MDP, the optimal policy is not unique due to the existence of multiple optimal actions for at least one state. In this case, Part (1) of the theorem implies $\E[Y]\neq 0$. Consequently, the asymptotic bias $\E[q^{(\alpha)}] - q^*$ of Q-learning is of $\sqrt{\alpha}$ order. As we will see in Section \ref{sec: averaging and extrapolation}, the precise characterization of order-$\sqrt{\alpha}$ bias allows one to use the Richardson-Romberg extrapolation for bias reduction. 

Parts (2) and (3) of the theorem imply that for Type-B MDPs (i.e., those with a unique optimal policy), the asymptotic bias can be controlled by the $n$-th order of the stepsize, as long as the noise has finite $2n$-th moment. For Q-learning, the random matrices $\{D_t, P_t\}_{t \geq 0}$ are bounded and thus all their moments are finite. If the rewards $\{r_t\}_{t \geq 0}$ also have finite arbitrary moments (e.g., they are Gaussian distributed or bounded), then the asymptotic bias is $\mathcal{O}(\alpha^n)$ for any $n\ge 1$, that is, the bias decays superpolynoimally with respect to the stepsize. Prior work in~\cite{zhang2024constant} considers the constant stepsize Q-learning algorithm with Type B MDPs and Markovian data. They prove that the bias is proportional to the stepsize~$\alpha$. We note that when the data is i.i.d., their results reduce to the case where the linear bias term vanishes, and the bias is of order~$\mathcal{O}(\alpha^2)$. In contrast, our Theorem~\ref{thm: q EY} leverages the lightness of the tail to provide a tighter upper bound on the bias. The proof of Theorem \ref{thm: q EY} is provided in Section \ref{sec:q-EY}.

\section{Iterate Averaging and Extrapolation}\label{sec: averaging and extrapolation}

In this section, we study the implications of our theoretical results for iterate averaging and extrapolation. In particular, we consider applying Polyak-Ruppert (PR) tail averaging \citep{ruppert1988efficient, polyak1992acceleration, jain2018parallelizing} and Richardson-Romberg (RR) extrapolation \citep{hildebrand1987introduction} to the iterates generated by contractive SA algorithms, and investigate the resulting estimation errors and biases in the presence of nonsmoothness.

To this end, we will first state two general results for PR averaging and RR extrapolation, respectively. We remark that these general results cover settings broader than those considered in this paper and may be of independent interest. We then apply these results to the nonsmooth contractive SA and Q-learning procedures studied in Section~\ref{sec:add-noise} and Section~\ref{sec:syn-q}. 


Let $\{\theta_t^{(\alpha)}\}_{t \geq 0}$ be a sequence of (raw) iterates in $\R^d$ generated by an SA procedure of the form
\begin{equation}\label{eq: markovian general SA}
\theta_{t+1}^{(\alpha)} = \theta_t^{(\alpha)} + \alpha \big(\tmT(x_t, \theta_t^{(\alpha)} )-\theta_t^{(\alpha)}\big)
\end{equation}
with a constant stepsize $\alpha > 0$.
We assume that the noise sequence $\{x_t\}_{t \geq 0}$ is a uniformly ergodic Markov chain defined on a general state space $\mathcal{X}$ with transition kernel $p$ and stationary distribution $\mu_{\mathcal{X}}$. Let $\tau_\alpha$ denote its $\alpha$-mixing time, i.e., 
$\tau_\alpha := \min\{t \geq 1: \max_{x \in \mathcal{X}} \|p^t(x,\cdot) -  \mu_{\mathcal{X}}\|_{\operatorname{TV}} \leq \alpha\},$ 
where $\|\cdot\|_{\operatorname{TV}}$ denotes the total variation norm. Note that a sequence of i.i.d.\ noise $\{x_t\}_{t \geq 0}$ is a uniformly ergodic Markov chain with $\tau_\alpha = 1$ for all $\alpha>0$.

We introduce two conditions on the raw SA iterates $\{\theta_t^{(\alpha)}\}_{t \geq 0}$, which allow us to quantify the performance of PR averaging and RR extrapolation with respect to a target vector $\theta^*$.

\begin{condition}[Distributional convergence]\label{condition: geo convergence}
There exist constants $C_0, C_1, \bar{\alpha}>0$ satisfying $0< 1-\bar{\alpha} C_1 < 1$ such that  for some random variable $\theta^{(\alpha)}$ it holds that
\[W^2_2(\mathcal{L}(\theta_t^{(\alpha)}), \mathcal{L}(\theta^{(\alpha)})) \leq C_0 \cdot (1-\alpha C_1 )^t, \quad \forall t \geq \tau_\alpha \text{ and } \forall \alpha \leq  \bar{\alpha}.\]
\end{condition}

\begin{condition}[Asymptotic bias and variance]\label{condition: bias}
There exist constants $\beta>0$ and $\delta \geq 0$ such that 
\begin{align}
\E[\theta^{(\alpha)}] = \theta^* + \alpha^\beta B + o(\alpha^{\beta+\delta}), \label{eq:condition_bias_general}
\end{align}
where $B \in \R^d$ is a vector independent of $t$ and $\alpha.$
Moreover, $\E[\|\theta^{(\alpha)} - \theta^*\|_2^2] \in \mathcal{O}(\alpha\tau_\alpha)$.
\end{condition}

Note that as the stepsize $\alpha$ gets larger, the iterates enjoys a faster geometric convergence by Condition~\ref{condition: geo convergence}, but admits a larger bias by Condition~\ref{condition: bias}. We will verify these conditions under general contractive SA and Q-learning settings.

\subsection{Polyak-Ruppert Tail Averaging}

Polyak-Ruppert (PR) averaging procedure \citep{ruppert1988efficient, polyak1992acceleration, jain2018parallelizing} is a popular method for reducing the variance of the SA iterates and accelerating the convergence. Specifically, given a burn-in period $k_0\geq 0$, we compute the tail-averaged iterates as: 
\begin{align}
\bar{\theta}^{(\alpha)}_{k_0, k}&:=\frac{1}{k-k_0} \sum_{t=k_0}^{k-1} \theta_t^{(\alpha)},\quad \text{for } k\geq k_0 +1. \label{eq:PR}
\end{align}

The following proposition provides non-asymptotic bounds for the first two moments of the tail-averaged iterate $\bar{\theta}^{(\alpha)}_{k_0, k}.$ 

\begin{proposition}\label{co: average}
Under Conditions \ref{condition: geo convergence} and \ref{condition: bias}, we have for all $k_0 \geq \frac{2}{\alpha C_1}\log\big(\frac{1}{\alpha\tau_\alpha}\big)$ and $k \geq k_0+\tau_\alpha$:
\begin{align}
\mathbb{E}\big[\bar{\theta}^{(\alpha)}_{k_0, k}\big]-\theta^*&= \alpha^\beta B  + o(\alpha^{\beta+\delta})+\mathcal{O}\Big(\frac{1}{\alpha(k - k_0)} \exp\big(-\frac{\alpha C_1 k_0}{2}\big)\Big),\label{eq: TA1}\\
\mathbb{E}\big[\left(\bar{\theta}_{k_0, k}-\theta^*\right)\left(\bar{\theta}_{k_0, k}-\theta^*\right)^{\top}\big] &= {\alpha^{2\beta}BB^\top + o(\alpha^{2\beta+\delta})} + {\mathcal{O}\Big(\frac{1}{\alpha\left(k-k_0\right)^2} \exp \big(-\frac{\alpha C_1k_0}{2}\big)\Big)} + {\mathcal{O}\Big(\frac{\tau_\alpha}{k-k_0}\Big)}.\label{eq: TA2} 
\end{align}
\end{proposition}

As a typical application of the above result, let us set the burn-in parameter as $k_0=k/2$ and parse the result for the second moment in Equation \eqref{eq: TA2}. 
The first two terms on the right-hand side of  \eqref{eq: TA2} correspond to the squared asymptotic bias of the raw iterates in Condition~\ref{condition: bias}, and cannot be reduced by averaging. The third term captures the optimization error, which decays geometrically in $k$ due to the geometric distributional convergence in Conditions \ref{condition: geo convergence}. The last term corresponds to the variance of averaged iterate $\bar{\theta}_{k/2, k}$, which decays at a rate $\mathcal{O}(1/k)$ due to averaging over $k/2$ raw iterates that are geometrically mixed. The proof of Proposition \ref{co: average}  generalizes the arguments on Linear SA in~\cite{huo2023bias}, and
is provided in Section \ref{sec:proof_average}.

\subsection{Richardson-Romberg Extrapolation}

With the fine-grained characterization of the asymptotic bias in Condition \ref{condition: bias}, one can use the RR extrapolation technique \citep{hildebrand1987introduction} to reduce the bias to a higher order term of the stepsize $\alpha$. In particular, we consider first-order RR extrapolation, where we run two SA recursions \eqref{eq: markovian general SA} in parallel under two different stepsizes $\alpha$ and $2\alpha$, with either the \emph{same} or \emph{different} sequence of noise $\{w_t\}_{t\geq0}.$ The resulting tail-averaged iterates $\bar{\theta}_{k_0, k}^{(\alpha)}$ and $\bar{\theta}_{k_0, k}^{(2\alpha)}$ are defined as \eqref{eq:PR}.
The RR extrapolated iterates are then computed as follows as a linear combination of the two averaged iterates:
\begin{align}
\tilde{\theta}_{k_0, k}^{(\alpha)}&=\frac{2^\beta}{2^\beta-1} \bar{\theta}_{k_0, k}^{(\alpha)}-\frac{1}{2^\beta-1}\bar{\theta}_{k_0, k}^{(2 \alpha)}. \label{eq:RR_def}
\end{align}
The coefficients of the above linear combination are chosen in the way that we can cancel out the dominating terms $\alpha^\beta$ and $(2\alpha)^\beta$ in the biases, as stated in the following proposition.

\begin{proposition}\label{co: RR}
Under Conditions \ref{condition: geo convergence} and \ref{condition: bias}, the RR extrapolated iterates defined in \eqref{eq:RR_def} satisfy the following bounds for all $k_0 \geq \frac{2}{\alpha C_1}\log\big(\frac{1}{\alpha\tau_\alpha}\big)$ and $k \geq k_0+\tau_\alpha$:
\begin{align}
\mathbb{E}\big[\tilde{\theta}^{(\alpha)}_{k_0, k}\big]-\theta^*&\in  o(\alpha^{\beta+\delta})+\mathcal{O}\Big(\frac{1}{\alpha(k - k_0)} \exp\big(-\frac{\alpha C_1 k_0}{2}\big)\Big),\label{eq:RR_1}\\
\mathbb{E}\big[(\tilde{\theta}_{k-k_0}-\theta^*)(\tilde{\theta}_{k-k_0}-\theta^*)^{\top}\big]&\in {o(\alpha^{2\beta+2\delta})} + {\mathcal{O}\Big(\frac{1}{\alpha\left(k-k_0\right)^2} \exp \big(-\frac{\alpha C_1k_0}{2}\big)\Big)}+{\mathcal{O}\Big(\frac{2^{2\beta}}{(2^\beta-1)^2}\frac{\tau_\alpha}{k-k_0}\Big)}.\label{eq:RR_2}
\end{align}
\end{proposition}
Proposition~\ref{co: RR} follows directly from Proposition~\ref{co: average} and the proof of~\citet[Corollary 4.7]{huo2023bias}; we omit its proof.
Again let us focus on the second moment bound~\eqref{eq:RR_2} with $k_0 = k/2$. Compared with the $\Theta(\alpha^{2\beta})$ squared asymptotic bias for the original iterates in \eqref{eq: TA2}, the squared bias of RR-extrapolated iterates is reduced to $o(\alpha^{2\beta+2\delta})$. Meanwhile, we retain the geometric convergence of the optimization error (the second right hand side term) and the $\mathcal{O}(1/k)$ rate of the variance (the last right hand side term). We remark that the effectiveness of RR extrapolation relies on identifying the precise order $\alpha^{\beta}$ of the bias; however, it does not require the knowledge of the coefficient $B,$ which is a problem instance dependent constant typically unknown.

\subsection{Applications to Contractive SA and Q-Learning}
\label{sec:application}

First consider the general contractive SA dynamic \eqref{eq: additive raw dynamic} from Section~\ref{sec:add-noise}. By Theorem~\ref{thm: additive convergence}, Condition \ref{condition: geo convergence} holds with $C_1=1-\sqrt{\gamma}$ and $\tau_{\alpha}=1$, By Theorem \ref{thm: additive limit}, 
Condition \ref{condition: bias} holds with $\tau_\alpha = 1, B=\E[Y], \beta = \frac{1}{2}$ and $\delta = 0$. 
Hence, Proposition \ref{co: RR} with $k_0=k/2$ implies the following MSE bound for the RR-extrapolated iterates:
\begin{align*}
\mathbb{E}\big\|\tilde{\theta}_{k/2}-\theta^*\big\|^2 
    &\in {o(\alpha)} + {\mathcal{O}\Big(\frac{1}{\alpha k^2} \exp \big(-\frac{\alpha (1-\sqrt{\gamma}) k}{4}\big)\Big)}+{\mathcal{O}\Big(\frac{1}{k}\Big)}.
\end{align*}

Similarly, for the Q-learning dynamic \eqref{eq: theta} in Section~\ref{sec:syn-q}, Condition \ref{condition: geo convergence} holds with $C_1=1-\sqrt{\gamma_0}$ and $\tau_{\alpha}=1$ by Corollary \ref{thm: q convergence}; Condition \ref{condition: bias} holds with $\tau_\alpha = 1, B=\E[Y], \beta = \frac{1}{2}$ and $\delta = 1/4$ by Corollary \ref{thm: q limit}. 
Consequently, we have the following MSE bound for the RR-extrapolated iterates:
\begin{align*}
\mathbb{E}\big\|\tilde{\theta}_{k/2}-\theta^*\big\|^2 
    &\in {o(\alpha^{3/2})} + {\mathcal{O}\Big(\frac{1}{\alpha k^2} \exp \big(-\frac{\alpha (1-\sqrt{\gamma_0}) k}{4}\big)\Big)}+{\mathcal{O}\Big(\frac{1}{k}\Big)}.\label{eq:RR_2}
\end{align*}
    
In both settings, the asymptotic bias of the raw iterate is on the order of $\sqrt{\alpha} \E[Y]$, which is reduced to $o(\sqrt{\alpha})$ or $o(\alpha^{3/4})$ by RR extrapolation. It is important to note that the bias order here differs from the typical $\Theta(\alpha)$ bias that arises in smooth SGD/SA dynamics~\citep{dieuleveut2020bridging,huo2023bias}. Correctly identifying the bias order, as established by our theoretical results, is crucial for RR extrapolation to be effective. 
We note that if $\E[Y] = 0$, the bias of the raw iterate is already $o(\sqrt{\alpha})$. In this case the above RR extrapolation scheme may not lead to further improvement, but it does not affect performance either (up to constant factors).  In Section \ref{sec:experiments}, we provide numerical experiments demonstrating the effectiveness of RR extrapolation for bias reduction.


\section{Proof Outline}
\label{sec:outline}

In this section, we outline the proofs of our main results for general contractive SA. 

A key ingredient for the analysis of SA dynamics is an appropriately constructed Lyapunov function. Recall that $\mT$ is contractive w.r.t.\ the norm  $\|\cdot\|_c$. As the standard Lyapunov function $\phi(\cdot) = \frac{1}{2}\|\cdot\|_c^2$ is not necessarily differentiable, our analysis makes use of its \emph{Moreau envelope}~\citep{parikh2014proximal, chen2023lyapunov,chen2020finite}, which can be thought of as a smooth surrogate of $\phi$. In particular, let $h(\cdot) = \frac{1}{2}\|\cdot\|_2^2$, which  is 1-smooth with respect to $\|\cdot\|_2$. Because all norms in $\R^d$ are equivalent \citep{folland1999real}, there exist two positive constants $l_{cs}$ and $u_{cs}$ such that $l_{cs}\|\cdot\|_2 \leq \|\cdot\|_c \leq u_{cs}\|\cdot\|_2$. The Moreau envelope $M_\eta:\R^d\to \R$ of $\phi$ with respect to $h$ is defined as
\begin{equation}\label{eq: Moreau envelope}
M_\eta(x) = \inf_{u \in \R^d} \Big\{\phi(u) + \frac{1}{\eta}h(x-u)\Big\},\qquad \forall x\in \R^d.
\end{equation}
The basic properties of $M_{\eta}$ are summarized below. 

\begin{proposition}\label{prop: Moreau envelope}
$M_\eta$ has the following properties: (1) $M_\eta$ is convex and $\frac{1}{\eta}$-smooth with respect to $\|\cdot\|_2$; (2) there exists a norm $\|\cdot\|_m$ such that $M_\eta(x) = \frac{1}{2}\|x\|_m^2$; (3) it holds that $l_{cm}\|\cdot\|_m\leq \|\cdot\|_c \leq u_{cm}\|\cdot\|_m$, where $l_{cm} = (1+\eta l_{cs}^2)^{\frac{1}{2}} $ and $u_{cm} = (1+\eta u_{cs}^2)^{\frac{1}{2}}$; (4) $\langle \nabla M_\eta(x), y\rangle \leq \|x\|_m \|y\|_m,$ $\forall x, y \in \R^d$; (5) $\nabla M_\eta(cx) = c\nabla M_\eta(x)$, $\forall c \geq 0, x\in \R^d$.
\end{proposition}
The proof of items (1)--(4) above can be found in \citet[Proposition 1]{chen2023lyapunov} and \citet[Lemma A.1]{chen2020finite}. Item (5) follows from (1), (2) and Euler’s homogeneous function theorem \citep{euler2000foundations}.

In this section, we omit the subscript $(\alpha)$ in $\theta_t^{(\alpha)}$ when the dependence on the stepsize $\alpha$ is clear.

\subsection{Proof Outline for Proposition~\ref{thm: additive 2n moment}  and Theorem \ref{thm: additive convergence}} \label{sec:prf_outline_thm_additive_2n}

\paragraph{Proposition~\ref{thm: additive 2n moment} (Moment Bounds).} To bound the $(2n)$-th moment $\E \|\theta_t^{(\alpha)} - \theta^*\|_c^{2n}$, we use the  Moreau envelope $M_{\eta}$ as a Lyapunov function and generalize the arguments for MSE bounds in~\cite{chen2020finite, chen2023lyapunov} to higher moments by induction on $n$. 
In particular, using the contractive property of $\mT$ and the properties of $M_{\eta}$, we can obtain
\begin{equation}\label{eq: outline moreau envelope}
\begin{aligned}
\E[M_\eta^{n}(\theta_{t+1} - \theta^*)] \leq& \E\bigg[\Big(\underbrace{\left(1 - \alpha(1-\sqrt{\gamma}) \right)M_\eta(\theta_t - \theta^*)}_{T_1} + \underbrace{\frac{\alpha^2}{\eta}\|\tmT(x_t, \theta_t) - \tmT(x_t,\theta^*)\|_2^2}_{T_2} \\
&+ \underbrace{(1-\alpha)\alpha\langle \nabla M_\eta (\theta_t - \theta^*), \tmT(x_t, \theta_t) - \mT(\theta_t)\rangle}_{T_3} + \underbrace{\frac{\alpha^2}{\eta}\|\tmT(x_t, \theta^*) - \mT(\theta^*)\|_2^2}_{T_4}\Big)^{n}\bigg].
\end{aligned}
\end{equation}
Expanding the right hand side and noting $\tmT(x_t, \theta_t) - \mT(\theta_t)$ is zero mean given $\theta_t$, we derive $\E[nT_1^{n-1}T_3] = 0$ and $\E[T_1^n] \leq \left(1 - \alpha(1 - \sqrt{\gamma})\right)\E[M_\eta^{n}(\theta_{t} - \theta^*)]$. A careful calculation using  induction hypothesis shows that all other cross terms are of order $ \mathcal{O}(\alpha^{n+1}).$ Combining these bounds yields 
\begin{align}
   \E[M_\eta^n(\theta_t - \theta^*)] &\leq \E[M_\eta^n(\theta_{t_\alpha} - \theta^*)](1- \alpha(1 - \sqrt{\gamma}))^{t-t_{\alpha,n}} + c_n \alpha^n,\quad\forall t\geq t_{\alpha,n}, \label{eq:Moreau_2n_moment}
\end{align}
from which the desired moment bounds follow in light of part (3) of Proposition \ref{prop: Moreau envelope}.

\paragraph{Theorem \ref{thm: additive convergence} (Distributional Convergence).}  
The key step in proving Theorem~\ref{thm: additive convergence} is establishing the convergence of $W_2^2\big(\mathcal{L}(\theta_t^{(\alpha)}), \mathcal{L}({\theta^\prime}_t^{(\alpha)})\big)$ for two iterate sequences $\{\theta_t^{(\alpha)}\}_{t\ge0}$ and $\{{\theta^\prime}_t^{(\alpha)}\}_{t\ge0}$ with different initialization. Coupling these two sequences by sharing the noise sequence $\{x_t\}_{t \geq 0}$, we further reduce the problem to bounding $\E\|{\theta}_t^{(\alpha)}-{\theta^\prime}_t^{(\alpha)}\|_c^2$ and, in turn, to bounding $\E[M_\eta(\theta_t^{(\alpha)}-{\theta^\prime}_t^{(\alpha)})]$. The latter can be done following a similar argument as that for equation~\eqref{eq: outline moreau envelope}.

\subsection{Proof Outline for Theorem \ref{thm: additive smooth} (Bias Characterization)} \label{sec:prf_outline_thm_additive_smooth}


Recall that Theorem \ref{thm: additive limit} has established the universality result that all the contractive SA asymptotically have the same steady-state limit $Y$. Therefore, it is sufficient to study whether $\E[Y] \neq 0$ on the SA dynamic \eqref{eq: iid approximate} with additive Gaussian noise. Theorem \ref{thm: additive convergence} states that the stochastic process $\{Z_t^{(\alpha)}\}_{t\geq 0}$ driven by dynamic \eqref{eq: iid approximate} converges weakly in $W_2$ to a random variable $Z^{(\alpha)}$ corresponding to its stationary distribution. At stationarity we have the following equation in distribution:
\begin{equation}\label{eq: outline equation in distribution}
Z^{(\alpha)} \overset{\textup{d}}{=} (1-\alpha)Z^{(\alpha)} + \sqrt{\alpha}\big(\mT(\sqrt{\alpha} Z^{(\alpha)} + \theta^*) - \mT(\theta^*) + w\big),
\end{equation}
where $w \sim N(0,\Sigma)$ is independent with $Z^{(\alpha)}$ and $\Sigma = \operatorname{Var}(\tilde{\mT}(x_0, \theta^*)-\mT(\theta^*))$. Taking the expectation on both sides of the above equation yields
\[\E[Z^{(\alpha)}] = \frac{1}{\sqrt{\alpha}}\E[\mT(\sqrt{\alpha} Z^{(\alpha)} + \theta^*) - \mT(\theta^*)].\]
Recall that the operator $\mT$ is one-sided directionally differentiable at $\theta^*$. We decompose the right-hand side of the above equation into two parts:
\begin{align}
\E[Z^{(\alpha)}] =&  {\frac{1}{\sqrt{\alpha}}\E\big[(\mT(\sqrt{\alpha} Z^{(\alpha)} + \theta^*) - \mT(\theta^*))\mathbbm{1}(\alpha^{\frac{1}{4}}Z^{(\alpha)} \notin B^d(0, 1))\big]} \tag{$T_1$} \\
&+ {\frac{1}{\sqrt{\alpha}}\E\big[(\mT(\sqrt{\alpha} Z^{(\alpha)} + \theta^*) - \mT(\theta^*))\mathbbm{1}(\alpha^{\frac{1}{4}}Z^{(\alpha)} \in B^d(0, 1))\big]} \tag{$T_2$}.
\end{align}

For the term $T_1$, we make use of the contraction property of $\mT$ and a concentration inequality to show that $\lim_{\alpha \to 0}T_1 = 0$.
Invoking Assumption~\ref{as: contraction0} we then deduce that  
\begin{equation}\label{eq:g}
\E[Y] = \lim_{\alpha \to 0}\E[T_2] =  \E[H(Y)], \quad \text{where }H\text{ is defined in equation \eqref{eq:h}}.
\end{equation}
We consider two cases.

\textbf{Case 1:}  If $\mathcal{T}$ is differentiable at $\theta^{*}$, Lemma~\ref{usefullemma2} implies that its one–sided directional derivative is linear; that is, $G(\theta)=J\theta$ for all $\theta\in\mathbb{S}^{d-1}$. Substituting this expression into~\eqref{eq:g} yields $(J-I)\mathbb{E}[Y]=0.$ Since $J-I$ is nonsingular by Lemma~\ref{usefullemma2}, it follows immediately that $\mathbb{E}[Y]=0.$
 
 \textbf{Case 2:} Without the loss of generosity, let us consider the case where the subdifferential of $H_1(\cdot)$ at 0 is not singleton. Then there exist $z_1, z_2 \in \R^d$ such that 
\[
H_1(Y) =H_1(Y) - H_1(0) \geq z_j^\top Y, ~~~j = 1,2.
\]
Below we argue that  $\E[Y] \neq 0$ by contradiction. 

Suppose that $\E[Y] = 0$, in which case $\E[H(Y)] = \E[Y] = 0$. Therefore, we have $\E[H_1(Y) - z_j^\top Y] = 0$ for $j = 1,2$. Because $H_1(Y) - z_j^\top Y $ is always non-negative, we must have $H_1(Y) - z_j^TY = 0$ almost surely for $j = 1,2$. Therefore, we have $z_1^\top Y = z_2^\top Y$ almost surely. Letting $\zeta = z_1 - z_2$, we have $\zeta^\top Y = 0$ almost surely, which implies 
$ \E[(\zeta^{\top}Y)^2]=0. $
By equation \eqref{eq: outline equation in distribution}, we obtain
\begin{align}
\E[(\zeta^{\top}Z^{(\alpha)})^2] =& (1-\alpha)^2\E\big[(\zeta^{\top}Z^{(\alpha)})^2\big] + 2\sqrt{\alpha}(1-\alpha)\E\big[\zeta^{\top}Z^{(\alpha)}\cdot \zeta^{\top} (\mT(\sqrt{\alpha} Z^{(\alpha)} + \theta^*) - \mT(\theta^*) )\big]\nonumber\\
&+ \alpha\E\big[\big(\zeta^{\top} (\mT(\sqrt{\alpha} Z^{(\alpha)} + \theta^*) - \mT(\theta^*) + w )\big)^2\big].\label{eq:prfoutline}
\end{align}
By properly scaling both sides of the above equation and then taking $\alpha \to 0$, we can finally obtain 
\[\E[(\zeta^{\top}Y)^2]\geq \E[(\zeta^{\top}w)^2] = \zeta^{\top}\var(w)\zeta > 0,\]
which contradicts with the equality $\E[(\zeta^{\top}Y)^2]=0$ established above. We conclude that $\E[Y] \neq 0.$


\section{Numerical Experiments}
\label{sec:experiments}

We present numerical experiments on asynchronous Q-learning~\citep{chen2023concentration} to demonstrate our distributional convergence results (Theorem \ref{thm: additive convergence} and Corollary \ref{thm: q convergence}), bias characterization results (Theorem \ref{thm: additive smooth} and Theorem \ref{thm: q EY}), and findings on iterate averaging and RR extrapolation in Section \ref{sec: averaging and extrapolation}.

First, we randomly generate a Markov Decision Process (MDP) with $|\mathcal{S}| = 20$ states and $|\mathcal{A}| = 2$ actions. The expected reward function $\bar{r}$ is sampled uniformly from the hypercube $[-1, 1]^{|\mathcal{S}|\times|\mathcal{A}|}$. Each row of the transition kernel $P$ are generated by drawing $|\mathcal{S}|$ independent uniform $[0,1]$ random variables and normalizing them such that each row sums to 1. Due to the random sampling of the transition matrix and the expected reward, the resulting MDP is likely to be a Type B MDP, where each state has a unique optimal action. Next, we construct a Type A MDP by manually setting two actions for each state to share the same transition probabilities and expected rewards, thereby introducing non-uniqueness in the optimal policy. For both Type A and B MDPs, the observed rewards are Gaussian distributed, with the variance being uniformly distributed in $[0,1]$.

We execute asynchronous Q-learning on our constructed Type A and B MDPs with a randomly sampled initialization $q_0 \sim \mathcal{N}\left(0, 0.5 \cdot I_{|\mathcal{S}|\times|\mathcal{A}|}\right)$, using stepsizes $\alpha \in \{0.02, 0.04, 0.08, 0.16\}$. We conduct 20 independent trials for each stepsize.

\begin{figure}[!htbp]
  \centering
   \subfigure[Type A MDPs.]{
    \includegraphics[width=0.45\textwidth]{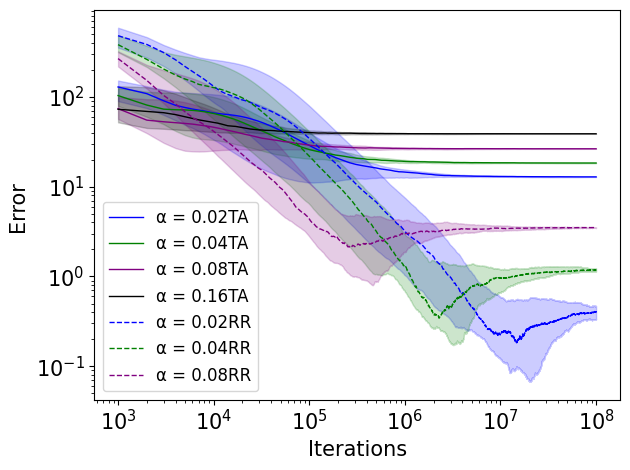}\label{fig:A}}
  \hfill
    \subfigure[Type B MDPs.]{              
    \includegraphics[width=0.45\textwidth]{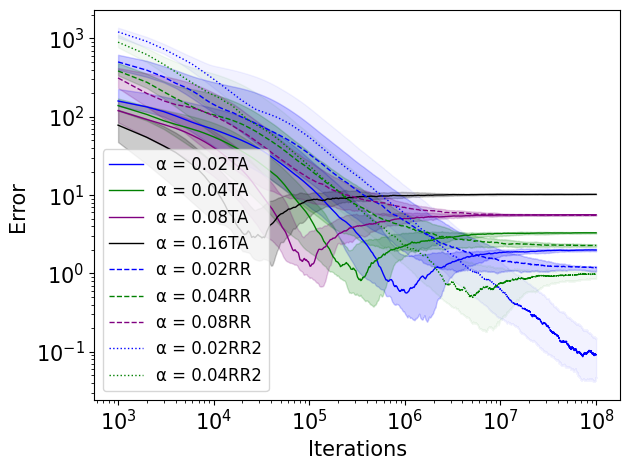}\label{fig:B}}
  \caption{The errors of tail-averaged (TA) and
RR extrapolated iterates with different stepsizes $\alpha$. In the legends, \textit{$\alpha=x$ RR} means RR extrapolation with two stepsizes $x$ and $2x$ and \textit{$\alpha=x$ RR2} means RR extrapolation with three stepsizes $x$, $2x$ and $4x$. The results show the median over 20 trials and the shaded region is the $95\%$ confidence interval.}
  \label{fig:TAandRR}
\end{figure}

Theorems \ref{thm: q limit} and \ref{thm: q EY} indicate that for Type A MDPs, the bias of the TA iterates is $\Theta(\sqrt{\alpha})$, whereas the RR extrapolation can reduce the bias to $o(\alpha^{3/4})$. This prediction is consistent with the results illustrated in Figure \ref{fig:A}. As discussed in  Section \ref{sec:application} for Theorem \ref{thm: q EY}, for Type B MDPs, the bias is already small, meaning that RR extrapolation with two stepsizes to eliminate $\sqrt{\alpha}$ bias may not lead to significant improvements. Figure \ref{fig:B} shows that RR extrapolation with two stepsizes either slightly reduces the bias or achieves performance comparable to the original iterates in this case. We further consider RR extrapolation with three stepsizes  $\tilde{\tilde{q}}^{(\alpha)}_{k_0,k} = (4+2\sqrt{2})\bar{q}_{k_0,k}^{(\alpha)}-(4+3\sqrt{2})\bar{q}_{k_0,k}^{(2\alpha)}+(1+ \sqrt{2})\bar{q}_{k_0,k}^{(4\alpha)}$ for this scenario, with the goal of eliminating both $\sqrt{\alpha}$ and $\alpha$ bias. As illustrated in Figure \ref{fig:B}, RR extrapolation with three stepsizes leads to a significant improvement, which implies that the bias of Type B MDP is at least in a order of $\mathcal{O}(\alpha)$, consistent with the results in Theorem \ref{thm: q EY}.

\begin{figure}[!htbp]
  \centering
  \subfigure[Ratio of Error for Type A MDPs.]{
    \includegraphics[width=0.45\textwidth]{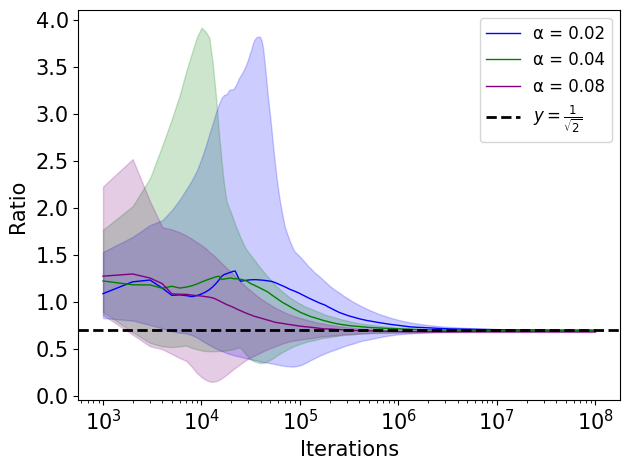}\label{fig:Ratio1}}
  \hfill
  \subfigure[Ratio of Error for Type A MDPs.]{
    \includegraphics[width=0.45\textwidth]{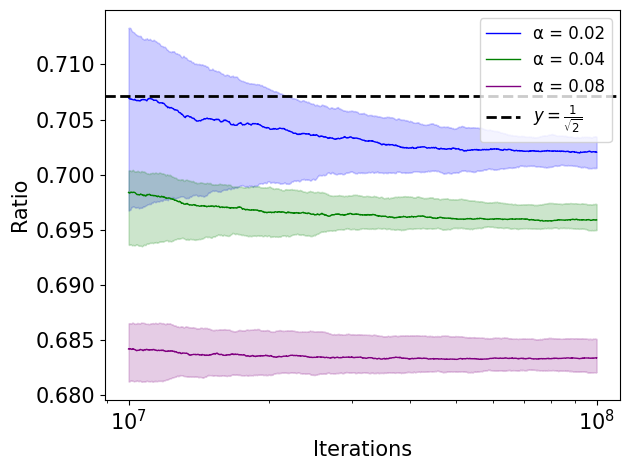}\label{fig:Ratio2}}
  
  \caption{The ratios of errors of simple averaged iterates with different stepsizes $\alpha$. In the legends, $\alpha = x$ means the ratio $\|\bar{q}^{(x)}_{0,k} - q^*\|/\|\bar{q}^{(2x)}_{0,k} - q^*\|$ with two stepsizes $x$ and $2x$. The results show the median over 20 MDP instances and the shaded region is the $95\%$ confidence interval.}
  \label{fig:Ratio}
\end{figure}

To further demonstrate our bias characterization results, we independently generate 20 Type A MDPs. We plot the ratio of the error of simple averaged iterates with stepsize $\alpha$ to that with stepsize $2\alpha$, for $\alpha \in \{0.02, 0.04, 0.08\}$, i.e., $\|\bar{q}^{(\alpha)}_{0,k} - q^*\|/\|\bar{q}^{(2\alpha)}_{0,k} - q^*\|$.  According to Proposition \ref{co: average}, for each MDP we have approximately $\bar{q}^{(\alpha)}_{0,k} - q^* \approx c\sqrt{\alpha},$ as $ k \to \infty,$ where $c$ only depends on the underlying MDP parameters. Therefore, the ratio of the errors for stepsizes $\alpha$ and $2\alpha$ should converge approximately to $1/\sqrt{2}$. The results in Figures \ref{fig:Ratio1} and \ref{fig:Ratio2} confirm this prediction. Furthermore, Figure \ref{fig:Ratio2} shows that with smaller stepsizes $\alpha$, the ratio converges to a value closer to $1/\sqrt{2}$, supporting our results in Proposition \ref{co: average}.

\section{Conclusion}

In this work, we study nonsmooth contractive SA with a constant stepsize. We develop prelimit coupling techniques for establishing steady-state convergence and characterizing the asymptotic bias, highlighting the impact of nonsmoothness on steady-state behavior. As an intermediate step, we investigate our coupling technique to bridge the gap between general SA and SA with only additive Gaussian Noise. The resulting universal guarantee demonstrates the potential of our technique for analyzing general SA. Our coupling techniques can also serve as a powerful tool for other nonsmooth dynamical systems such as piecewise smooth diffusion, stochastic differential equations and their discretization. 

There are several natural directions for future work:
\begin{itemize}
    \item One direction of immediate interest is to obtain more refined characterization of the steady-state distribution $Y^{(\alpha)}$ and its limit $Y$, including higher moment results,  other functionals of the distribution, as well as non-asymptotic results as a function of $\alpha$ and the level of nonsmoothness.
    \item Generalizing our results to general noise settings  is another interesting future direction. For additive martingale difference noise, we believe the current analysis can be combined with an appropriate martingale Berry-Esseen Central Limit Theorem to establish similar distributional and steady-state convergence results. For more general multiplicative Markovian noise, however, establishing such results would require extending our coupling techniques to exploit the structure of Markovian nonsmooth SA.
    \item In this work, we focus on the unprojected stochastic approximation update
\eqref{eq: additive raw dynamic}. Many interesting applications involve the projected version of stochastic approximation~\citep{andradottir1991projected}.
Recall that Assumption~\ref{as: contraction} imposes a local (potentially nonsmooth) structural condition at the fixed point \(\theta^*\). With a projection operator, the interaction between the
update map and the geometry of the set we project onto may introduce additional
nonsmoothness. Extending our main results in
Section~\ref{sec:add-noise} to the projected setting is an interesting direction for future work.

\end{itemize}  
\ACKNOWLEDGMENT{The authors thank the area editor, associate editor, and
reviewers for constructive comments and suggestions that
significantly improved the quality of this paper.}

\bibliographystyle{plainnat}
\bibliography{main_journal}
\noindent\textbf{Yixuan Zhang} is a Ph.D. candidate in the Department of Industrial \& Systems Engineering at the University of Wisconsin--Madison. His research interests include modern probabilistic methods for stochastic algorithms and online learning with applications to operations research. \medskip 

\noindent\textbf{Dongyan (Lucy) Huo} is an assistant professor in the Department of Industrial Engineering \& Decision Analytics at the Hong Kong University of Science and Technology. She received her Ph.D. from the School of Operations Research and Information Engineering at Cornell University. Her research interests include applied probability, queueing theory, stochastic optimization, and reinforcement learning. \medskip 

\noindent\textbf{Yudong Chen} is an associate professor in the Department of Computer Sciences at the University of Wisconsin--Madison. He received his Ph.D. in Electrical and Computer Engineering from the University of Texas at Austin. His research interests include reinforcement learning theory, non-convex and nonsmooth learning problems, stochastic optimization and approximation. \medskip 

\noindent\textbf{Qiaomin Xie} is an assistant professor in the Department of Industrial \& Systems Engineering and an affiliate faculty member in the Department of Computer Sciences at the University of Wisconsin--Madison. She received her Ph.D. in Electrical and Computer Engineering from the University of Illinois Urbana--Champaign. Her research interests include reinforcement learning, game theory, applied probability and stochastic networks, with applications to computer and communication networks.
\ECSwitch
\section{Additional Related Work}\label{sec: related work}

In this section, we discuss the existing results that are most relevant to our work.

\subsection{Results on SA and SGD}

The study of SA and SGD traces its origins to the seminal work by Robbins and Monro~\citep{robbins1951stochastic}. Classical results focus on 
diminishing stepsize regime~\citep{robbins1951stochastic, blum1954approximation} and have established almost sure asymptotic convergence for SA and SGD algorithms. Subsequent work~\citep{ruppert1988efficient, polyak90_average} proposes the iterate averaging technique, now known as Polyak-Ruppert (PR) averaging, to accelerate convergence. 
The asymptotic convergence theory of SA and SGD is well-developed and extensively addressed in many textbooks~\citep{kushner2003-yin-sa-book, benveniste2012adaptive,WrightRecht2022_OptBook}. Recent studies in \cite{chen2022finite, Chandak22-QLearn, lauand2024revisiting}
analyze non-asymptotic convergence under diminishing stepsizes. In particular,  \cite{lauand2024revisiting} provide an explicit
\(O(\alpha_n)\) bias bound for the linear SA setting.
The recent work \cite{chen2023concentration} establishes the high probability bound on the estimation error of contractive SA with diminishing stepsize.

SA and SGD with constant stepsizes have gained attention in recent work, most of which assumes i.i.d.\ data. When using a constant stepsize, one loses the almost sure convergence guarantee from the diminishing stepsize sequence regime, but distributional convergence can be established as shown in~\cite{dieuleveut2020bridging, yu2021analysis,chen2022stationary, huo2023bias, zhang2024constant}. 
A recurring observation in the literature is the presence of asymptotic bias when using constant stepsize in SA, i.e., $\E[\theta^{(\alpha)}]\neq \theta^*.$  When the SA update is locally smooth, the asymptotic bias has been demonstrated to be of $\Theta(\alpha)$ order in \cite{dieuleveut2020bridging, huo2023bias, zhang2024constant}. The work in \cite{yu2021analysis} considers nonsmooth SA but only provides an upper bound for the asymptotic bias, i.e. $|\E[\theta^{(\alpha)}] - \theta^*| \leq c\sqrt{\alpha} $. 
\citet{he2019asymmetric} examine an asymmetric-valley loss landscape that is conceptually close to our nonsmooth setting. Their result establishes the non-zero bias result $\E[\theta^{(\alpha)}] - \theta^* \neq 0 $ in the one-dimensional case, but does not provide an explicit characterization of how this bias depends on the stepsize~$\alpha$ nor consider the higher dimensional setting.
Many papers provide non-asymptotic MSE upper bounds. The work in \cite{Lakshminarayanan18-LSA-Constant-iid, Mou20-LSA-iid} studies linear SA under i.i.d.\ data and provides an upper bound on the MSE. There are also papers that analyze the MSE with Markovian data, such as \cite{srikant2019finite, Mou21-optimal-linearSA, durmus2021-LSA, durmus22-LSA}. The work in \cite{chen2023lyapunov, chen2020finite} introduces the Generalized Moreau Envelope (GME) to analyze the MSE of general contractive SA.  In our work, we extend the GME technique to analyze a different class of SA, specifically, establishing upper bounds for any $2n$-th moment and showing weak convergence of SA iterates. We also integrate GME with our prelimit coupling technique to prove steady-state convergence as stepsize $\alpha$ decreases to $0$.

\subsection{Applications in Reinforcement Learning}
Many widely employed iterative algorithms in reinforcement learning (RL) can be reformulated as SA problems \citep{Sutton18-RL-book,Bertsekas19-RL-book}. Among those, the two most well-known algorithms are the temporal-difference (TD) learning for policy evaluation \citep{Sutton1988-td, Dayan1994-tdlambda} and Q-learning for optimal policy learning \citep{Watkins92-QLearning}. The TD algorithms with linear function approximation can be cast into the framework of linear SA. 
Q-learning is a nonsmooth and nonlinear contractive SA, and has also been studied extensively in both classical work for asymptotic convergence \citep{Tsitsiklis1994-QLearn, Szepesvari97-QLearn-Rates} and recent work for non-asymptotic guarantee \citep{EvenDar04-QLearn-rates,chen2022finite, chen2023lyapunov}. 
The work \cite{zhang2024constant} studies the stationary distribution of asynchronous Q-learning with Markovian data and characterizes the asymptotic bias under the assumption that MDP has a unique optimal policy. 


\subsection{Nonsmooth Function Class}

Nonsmooth functions have been studied in many papers, including semi-smoothness \citep{mifflin1977semismooth},  identifiable surfaces \citep{wright1993identifiable},  $\mathcal{U}$$\mathcal{V}$-structures \citep{lemarechal2000, mifflin2005algorithm}, partly smoothness \citep{lewis2002active}, $g \circ F$ decomposition \citep{shapiro2003class, sagastizabal2013composite} and minimal identifiable sets \citep{drusvyatskiy2014optimality, davis2023asymptotic}. In this work, we follow the one-sided directional differentiability notion of \cite{rockafellar1998variational} to define our class of nonsmooth stochastic approximation operators.

\subsection{Steady-State Convergence}
The steady-state convergence is commonly studied in the realm of stochastic processes, with one well-known application being the steady-state convergence in queueing networks. 
As discussed in Section \ref{sec: challenge}, the classical method is through justifying the interchange of limits, as seen in \cite{GamaZeev2006, Gurv2014a, YeYao2016, YeYao2018}.
An alternative approach is through the basic adjoint relationship (BAR) approach, which studies the generator of the Markov process, i.e., $\E[Gf(Y^{(\alpha)})]=0$ as $\alpha\to0$ \citep{BravDaiMiya2017, BravDaiMiya2023, chen2022stationary}. Notably, this approach demands a careful construction of test functions $f$. 
Another line of work related to steady-state convergence focuses on the unadjusted Langevin algorithm (ULA)~\citep{durmus2017nonasymptotic, durmus2019high}. These papers take an approach similar to the justification of limit interchange in queueing networks, where one first demonstrates the convergence of ULA to the corresponding stochastic differential equation (SDE), and then relates the convergence to the stationary distribution of the SDE. 


\section{Technical Lemmas for Nonsmooth Operators}

This section presents several auxiliary lemmas for the nonsmooth contractive operator $\mT$ introduced in Assumptions \ref{as: contraction0} and \ref{as: contraction}.

\subsection{Example of Contractive Operator Violating Assumption \ref{as: contraction}}

The following lemma provides an example showing that not every contractive operator satisfies Assumption \ref{as: contraction}.
\begin{lemma}
\label{lem:counterexample}
Define the mapping $\mT:\R\to\R$ by
\[
\mT(x)=
\begin{cases}
0.4x +0.1x\sin\bigl(\ln|x|\bigr), & x\neq 0,\\
0, & x=0.
\end{cases}
\]
Then
\begin{enumerate}[label=(\roman*)]
    \item $\mT$ is a contraction on $(\R,|\cdot|)$ with Lipschitz constant $L=0.6$ and has the unique fixed point $\theta^{\ast}=0$.
    \item $\mT$ is not one-sided directionally differentiable at $\theta^{\ast}$.
\end{enumerate}
\end{lemma}

\begin{proof}{Proof of Lemma \ref{lem:counterexample}}

\begin{enumerate}[label=(\roman*)]
\item  
For $x,y\neq 0$,
\begin{align*}
    \mT(x)mT(y)&=0.4(x-y)+0.1\bigl[x\sin(\ln|x|)-y\sin(\ln|y|)\bigr]\\
    &= 0.4(x-y)+0.1\bigl[(x-y)\sin(\ln|x|)+y\,[\sin(\ln|x|)-\sin(\ln|y|)]\bigr].
\end{align*}
Using $|\sin t|\le1$ and $|\sin a-\sin b|\le |a-b|$, we have
\[
|x\sin(\ln|x|)-y\sin(\ln|y|)|\le |x-y|+|y|\bigl|\ln|x|-\ln|y|\bigr|.
\]
Assuming $|x|\ge|y|>0$ and setting $a:=|x|/|y|\ge1$, the inequality
$\ln a\le a-1$ yields
$|y|\bigl|\ln|x|-\ln|y|\bigr|\le |x|-|y|\le|x-y|$, so
\(
|x\sin(\ln|x|)-y\sin(\ln|y|)|\le 2|x-y|.
\)
Multiplying by $0.1$ and adding the linear part gives
\[
|\mT(x)-\mT(y)|\le (0.4+0.2)|x-y|=0.6\,|x-y|.
\]
If either argument is zero, the bound tightens to $0.5|x-y|$.  Hence
$\mT$ is globally Lipschitz with constant $L=0.6<1$ and is therefore a
contraction; Banach’s fixed-point theorem secures the unique fixed
point $\theta^{\ast}=0$.

\item 
For $w>0$,
\[
\frac{\mT(w)-\mT(0)}{w}=0.4+0.1\sin(\ln w).
\]
As $w \to 0^+$, $\sin(\ln w)$ oscillates between $-1$ and $1$, so the
limit does not exist.  Consequently, the one-sided directional
derivative of $\mT$ at $\theta^{\ast}$ in direction $+1$ (and likewise in
direction $-1$) fails to exist, proving that $\mT$ is not one-sided
directionally differentiable at its fixed point.
\end{enumerate}
\end{proof}

\subsection{Lemmas on One-Sided Directional Derivatives}

We present two lemmas on the one-sided directional derivative $G$ of $\mT$ as defined in Assumption~\ref{as: contraction}.

\begin{lemma}\label{usefullemma1}
    Under Assumptions \ref{as: contraction0} and \ref{as: contraction}, the following hold.
    \begin{enumerate}
    \item When $d \geq 2$, \(G\) is Lipschitz continuous on the unit sphere \(\mathbb{S}^{d-1}\).  In particular, \(\|G(\theta)\|_{c}<\infty\) for every \(\theta\in\mathbb{S}^{d-1}\).
    \item The convergence in Assumption~\ref{as: contraction} is uniform over the directions, i.e.,
    \begin{align*}
    \lim_{w \to 0^+}\sup_{\theta \in \mathbb{S}^{d-1}}\Big\|\frac{\mT( \theta^*+w\theta)-\mT(\theta^*)}{w} - G(\theta)\Big\|_2 = 0.
    \end{align*}
\end{enumerate}
\end{lemma}
\begin{proof}{Proof of Lemma \ref{usefullemma1}}
For each \(0<w\le 1\), define
\begin{align*}
F_w(u):=\frac{\mT\left(\theta^*+w u\right)-\mT\left(\theta^*\right)}{w}, \quad u \in \mathbb{S}^{d-1}.
\end{align*}
By Assumption \ref{as: contraction0}, for any \(\theta,\theta'\in\mathbb{S}^{d-1}\) we have
\begin{align*}
    \|G(\theta)-G(\theta')\|_c &\leq\|G(\theta)-F_w(\theta)\|_c+\|F_w(\theta)-F_w(\theta')\|_c+\|G(\theta')-F_w(\theta')\|_c \\
    &\leq \|G(\theta)-F_w(\theta)\|_c+\gamma\|\theta-\theta'\|_c+\|G(\theta')-F_w(\theta')\|_c\\
    &\overset{w \to 0}{\leq} \gamma\|\theta-\theta'\|_c.
\end{align*}
Thus \(G\) is \(\gamma\)-Lipschitz on the compact set \(\mathbb{S}^{d-1}\) when $d \geq 2$,
hence bounded.

Fix \(\varepsilon>0\) and set \(\delta:=\varepsilon/(3\gamma)\).
By the Lipschitz property of \(G\), 
\(\|G(\theta)-G(\theta')\|_{c}\leq\varepsilon/3\) whenever
\(\|\theta-\theta'\|_{c}\leq\delta\).
Because \(\mathbb{S}^{d-1}\) is compact, there exists a finite
\(\delta\)-net \(\{\theta_{1},\dots,\theta_{m}\}\subset\mathbb{S}^{d-1}\).
For each \(k\) choose \(w_{k}>0\) such that
\(\|F_{w}(\theta_{k})-G(\theta_{k})\|_{c}<\varepsilon/3\) whenever \(0<w<w_{k}\),
and set \(w(\varepsilon):=\min_{k} w_{k}\).

Now for each \(\theta\in\mathbb{S}^{d-1}\), there exists \(\theta_{k}\) from the net
satisfying \(\|\theta-\theta_{k}\|_{c}\leq\delta\).  For \(0<w<w(\varepsilon)\) we obtain
\begin{align*}
\left\|F_w(\theta)-G(\theta)\right\|_c & \leq \left\|F_w(\theta)-F_w\left(\theta_k\right)\right\|_c+\left\|F_w\left(\theta_k\right)-G\left(\theta_k\right)\right\|_c+\left\|G\left(\theta_k\right)-G(\theta)\right\|_c \\
& \leq \gamma \delta+\frac{\varepsilon}{3}+\gamma \delta=\varepsilon.
\end{align*}
This implies that 
\begin{align*}
\sup_{\theta \in \mathbb{S}^{d-1}}\Big\|\frac{\mT( \theta^*+w\theta)-\mT(\theta^*)}{w} - G(\theta)\Big\|_c \leq \varepsilon, \quad \forall 0<w<w(\varepsilon).
\end{align*}
As above holds for arbitrary $\varepsilon>0$, we obtain the desired limit.
Finally, equivalence of norms in finite-dimensional spaces allows the argument
to be carried out with any norm on \(\mathbb R^{d}\), completing the proof of Lemma \ref{usefullemma1}.
\end{proof}
\begin{lemma}\label{usefullemma2}
Under Assumptions \ref{as: contraction0} and \ref{as: contraction}, the following two statements are equivalent with the same matrix $J \in \R^{d \times d}$:
\begin{enumerate}
    \item $\mT$ is differentiable at $\theta^*$ with Jacobian $J \in \R^{d \times d}$, i.e.,
    \begin{align*}
     \lim_{\|h\|_2 \to 0}\frac{\|\mT(\theta^*+h)-\mT(\theta^*)-Jh\|_2}{\|h\|_2} = 0.
    \end{align*}
    \item For all $\theta \in \mathbb{S}^{d-1}$,  $G(\theta) = J\theta$.
\end{enumerate}
Furthermore, assuming the Jacobian $J$ exists, the matrix $J-I$ is nonsingular.
\end{lemma}
\begin{proof}{Proof of Lemma \ref{usefullemma2}}
If $\mT$ is differentiable at $\theta^*$ with Jacobian $J$, for all $\theta \in \mathbb{S}^{d-1}$, we have
\begin{align*}
\frac{\mT(w\theta+\theta^*)-\mT(\theta^*)}{w} = J\theta + o(1), \quad \text{ as }w \to 0,
\end{align*}
which implies $G(\theta) = J\theta$ for all $\theta \in \mathbb{S}^{d-1}$.

If $G(\theta) = J\theta$ for all $\theta \in \mathbb{S}^{d-1}$, for any $h \neq 0$, write $h = wu$ with $w = \|h\|_2$ and $u = h/\|h\|_2$. Then, by the equivalence of all norms in $\R^d$, 
\begin{align*}
    \frac{\|\mT(\theta^*+h)-\mT(\theta^*) - Jh\|_2}{\|h\|_2} = \|F_w(u)-J(u)\|_2 \leq \sup _{u \in \mathbb{S}^{d-1}}\left\|F_w(u)-J u\right\|_2 \overset{w \to 0+}{\longrightarrow} 0,
\end{align*}
which completes the proof of the equivalence.

For any $h \neq 0$, we notice that
\begin{align*}
    \|Jh\|_c =\|h\|_c\|Jh/\|h\|_c\|_c= \|h\|_c \lim_{w \to 0^+}\frac{\|\mT(wh/\|h\|_c+\theta^*)-\mT(\theta^*)\|_c}{w} \leq \gamma \|h\|_c.
\end{align*}
Therefore,
\begin{align*}
    \|(I-J)h\|_c \geq \|h\|_c - \|Jh\|_c \geq (1-\gamma)\|h\|_c>0,
\end{align*}
which implies $h \neq Jh$.
\end{proof}

\section{Proof of Proposition~\ref{thm: additive 2n moment}}\label{sec: addtive 2n moment}

Proposition~\ref{thm: additive 2n moment} follows from combining the following lemma and the property (3) in Proposition \ref{prop: Moreau envelope}.

\begin{lemma}\label{lemma: additive 2n moment}
For each integer $n \geq 1$, under Assumption~\ref{as: contraction0} and Assumption \ref{as: additive noise}(\textbf{n}), there exist $\eta>0$, $\bar{\alpha}>0$ such that for any $\alpha \leq \bar{\alpha}$, there exists $t_{\alpha,n}$ such that 
\[\E[M_\eta^n(\theta_t - \theta^*)] \leq \E[M_\eta^n(\theta_{t_\alpha} - \theta^*)](1- \alpha(1 - \sqrt{\gamma}))^{t-t_{\alpha,n}} + c_n \alpha^n\]
holds for all $t \geq t_{\alpha,n}$, where $M_\eta(\cdot)$ is defined in \eqref{eq: Moreau envelope} and $\{c_n\}_{n \geq 0}$ are universal constants that are independent with $\alpha$ and $t$. Moreover, $t_{\alpha,1} = 0$.
\end{lemma}

\subsection{Proof of Lemma \ref{lemma: additive 2n moment}}
\textbf{Base Case:} $n = 1$.
By subtracting $\theta^*$ from both side of equation \eqref{eq: additive raw dynamic}, we obtain
\begin{equation}\label{eq: additive modified dynamic}
\begin{aligned}
\theta_{t+1} - \theta^*&= \theta_t - \theta^* + \alpha(\tmT( x_t,\theta_t) - \theta_t)
=(1-\alpha)(\theta_t - \theta^*) + \alpha(\tmT(x_t, \theta_t) - \mT(\theta^*)),
\end{aligned}
\end{equation}
where the second equality holds because $\mT(\theta^*) = \theta^*$. Applying the  Moreau envelope $M_\eta(\cdot)$ defined in equation~\eqref{eq: Moreau envelope} to both sides of equation~\eqref{eq: additive modified dynamic} and by Proposition \ref{prop: Moreau envelope}, we obtain
\begin{align}
M_\eta(\theta_{t+1} - \theta^*) \leq& (1-\alpha)^2M_\eta(\theta_t - \theta^*) \nonumber\\
&+ (1-\alpha)\alpha\langle \nabla M_\eta (\theta_t - \theta^*), \tmT(x_t, \theta_t) - \mT(\theta^*)\rangle \label{eq: Moreau_T1}\\
&+ \frac{\alpha^2}{2\eta}\|\tmT(x_t, \theta_t) - \mT(\theta^*)\|_2^2.\label{eq: Moreau_T2}
\end{align}
The term in \eqref{eq: Moreau_T1} can be bounded as follows:
\begin{align*}
\eqref{eq: Moreau_T1} &= (1-\alpha)\alpha\big(\langle \nabla M_\eta (\theta_t - \theta^*), \mT(\theta_t) - \mT(\theta^*)\rangle + \langle \nabla M_\eta (\theta_t - \theta^*), \tmT(x_t, \theta_t) - \mT(\theta_t)\rangle\big)\\
&\overset{(\text{i})}{\leq} (1-\alpha)\alpha\big(\|\theta_t - \theta^*\|_m\|\mT( \theta_t ) - \mT(\theta^*)\|_m + \langle \nabla M_\eta (\theta_t - \theta^*), \tmT(x_t, \theta_t) - \mT(\theta_t)\rangle\big)\\
&\overset{(\text{ii})}{\leq} \frac{(1-\alpha)\alpha \gamma}{l_{cm}}\|\theta_t - \theta^*\|_m\|\theta_t - \theta^*\|_c + (1-\alpha)\alpha\langle \nabla M_\eta (\theta_t - \theta^*), \tmT(x_t, \theta_t) - \mT(\theta_t)\rangle\\
&\overset{(\text{iii})}{\leq} \frac{2\alpha(1-\alpha)\gamma u_{cm}}{l_{cm}} M_\eta(\theta_t - \theta^*) + (1-\alpha)\alpha\langle \nabla M_\eta (\theta_t - \theta^*), \tmT(x_t, \theta_t) - \mT(\theta_t)\rangle,
\end{align*}
where (i) holds due to property (4) of Proposition \ref{prop: Moreau envelope}, (ii) follows from property (3) of Proposition \ref{prop: Moreau envelope} and $\gamma$-contraction of $\mT(\cdot)$, and (iii) holds because of property (2) of Proposition \ref{prop: Moreau envelope}.

The term in \eqref{eq: Moreau_T2} can be bounded as follows:
\begin{equation*}
\begin{aligned}
\eqref{eq: Moreau_T2} \leq &  \frac{\alpha^2}{\eta}\big(\|\tmT(x_t, \theta_t) - \tmT(x_t,\theta^*)\|_2^2+\|\tmT(x_t, \theta^*) - \mT(\theta^*)\|_2^2\big).
\end{aligned}
\end{equation*}
Combining the above bounds, we obtain
\begin{equation*}
\begin{aligned}
M_\eta(\theta_{t+1} - \theta^*) \leq& \big(1 - 2\alpha(1-\frac{(1-\alpha)\gamma u_{cm}}{l_{cm}}) + \alpha^2\big)M_\eta(\theta_t - \theta^*) +  \frac{\alpha^2}{\eta}\|\tmT(x_t, \theta_t) - \tmT(x_t,\theta^*)\|_2^2 \\
&+ (1-\alpha)\alpha\langle \nabla M_\eta (\theta_t - \theta^*), \tmT(x_t, \theta_t) - \mT(\theta_t)\rangle + \frac{\alpha^2}{\eta}\|\tmT(x_t, \theta^*) - \mT(\theta^*)\|_2^2.
\end{aligned}
\end{equation*}

Recall that $\frac{u_{cm}}{l_{cm}} = \sqrt{\frac{1 + \eta u_{cs}^2}{1 + \eta l_{cs}^2}}$ by property (3) in Proposition \ref{prop: Moreau envelope}. We can always choose a sufficient small $\eta>0$ such that $\frac{u_{cm}}{l_{cm}} \leq \frac{1}{\sqrt{\gamma}}$, which implies $-2\alpha(1-\frac{(1-\alpha)\gamma u_{cm}}{l_{cm}}) \leq -2\alpha(1 - \sqrt{\gamma})$ and we have
\begin{equation}\label{eq: Moreau base}
\begin{aligned}
M_\eta(\theta_{t+1} - \theta^*) \leq& \big(1 -2\alpha(1 - \sqrt{\gamma}) + \alpha^2\big)M_\eta(\theta_t - \theta^*) +  \frac{\alpha^2}{\eta}\|\tmT(x_t, \theta_t) - \tmT(x_t,\theta^*)\|_2^2 \\
&+ (1-\alpha)\alpha\langle \nabla M_\eta (\theta_t - \theta^*), \tmT(x_t, \theta_t) - \mT(\theta_t)\rangle + \frac{\alpha^2}{\eta}\|\tmT(x_t, \theta^*) - \mT(\theta^*)\|_2^2.
\end{aligned}
\end{equation}

Taking expectation on both sides of equation \eqref{eq: Moreau base} yields
\begin{equation*}
\begin{aligned}
\E[M_\eta(\theta_{t+1} - \theta^*)] &\leq \big(1 - 2\alpha(1-\sqrt{\gamma}) + \alpha^2(1+\frac{2Lu_{cm}^2}{\eta l_{cs}^2})\big)\E[M_\eta(\theta_t - \theta^*)] + \frac{\alpha^2 c_x}{\eta },
\end{aligned}
\end{equation*}
where $c_x= \E[\|\tmT(x_0, \theta^*) - \mT(\theta^*)\|_c^2]$. The above inequality holds because $\tmT(x_t, \theta^*) - \mT(\theta^*)$ is zero mean and independent of $\theta_t - \theta^*$. Thus, there always exists $0<\bar{\alpha}_1 <1$ such that $\big(1 - 2\alpha(1 - \sqrt{\gamma})+ \alpha^2(1 + \frac{2L u_{cm}^2}{\eta l_{cs}^2})\big) \leq 1 - \alpha(1 - \sqrt{\gamma}) < 1$ when $\alpha \leq \bar{\alpha}_1$. Thus, for all $ \alpha \leq \bar{\alpha}_1$ and $t \geq 0$, we have
\begin{equation*}
\begin{aligned}
\E[M_\eta(\theta_{t+1} - \theta^*)] &\leq \left(1 - \alpha(1-\sqrt{\gamma})\right)\E[M_\eta(\theta_t - \theta^*)] + \frac{\alpha^2 c_x}{\eta }\\
&\leq \left(1 - \alpha(1 - \sqrt{\gamma})\right)^{t+1}\E[M_\eta(\theta_0 - \theta^*)] + \sum_{k = 0}^{t}\left(1 - \alpha(1 - \sqrt{\gamma})\right)^k\frac{\alpha^2 c_x}{\eta }\\
&\leq \left(1 - \alpha(1 - \sqrt{\gamma})\right)^{t+1}\E[M_\eta(\theta_0 - \theta^*)] + \frac{\alpha c_x}{\eta (1-\sqrt{\gamma})}.
\end{aligned}
\end{equation*}

\textbf{Induction Step:} 
Given a positive integer $k \geq 2$, assume Lemma \ref{lemma: additive 2n moment} holds for all $n \leq k-1$. When $n = k$, take $k$-th moment to both side of equation \eqref{eq: Moreau base} and we have
\begin{align*}
\E[M_\eta^{k}(\theta_{t+1} - \theta^*)] \leq& \E\Big[\big(\underbrace{(1 - 2\alpha(1-\sqrt{\gamma}) + \alpha^2)M_\eta(\theta_t - \theta^*)}_{T_1} + \underbrace{\frac{\alpha^2}{\eta}\|\tmT(x_t, \theta_t) - \tmT(x_t,\theta^*)\|_2^2}_{T_2} \\
&+ \underbrace{(1-\alpha)\alpha\langle \nabla M_\eta (\theta_t - \theta^*), \tmT(x_t, \theta_t) - \mT(\theta_t)\rangle}_{T_3} + \underbrace{\frac{\alpha^2}{\eta}\|\tmT(x_t, \theta^*) - \mT(\theta^*)\|_2^2}_{T_4}\big)^{k}\Big]  \\
=& \underbrace{\E\Big[\sum_{a+c = k}\tbinom{k}{a}\tbinom{k-a}{c} T_1^aT_3^c\Big]}_{S_1} + \underbrace{\E\Big[\sum_{a+b+c+d = k, b+d \geq 1} \!\!\!\!\!\! \tbinom{k}{a}\tbinom{k-a}{b}\tbinom{k-a-b}{c}T_1^aT_2^bT_3^cT_4^d\Big]}_{S_2}.
\end{align*}
The terms $S_1$ and $S_2$ are analyzed in the following Lemma \ref{lem: bound for T2}. Combining the bounds on $S_1, S_2$ and letting $\alpha$ to be sufficiently small, we obtain 
\[\E[M_\eta^k(\theta_{t+1} - \theta^*)] \leq \left(1 - \alpha(1-\sqrt{\gamma})\right)\E[M_\eta^k(\theta_t - \theta^*)] + \mathcal{O}(\alpha^{k+1}).\]
Therefore, there exists a threshold $\bar{\alpha}_k$ such that  for all $\alpha \leq \bar{\alpha}_k$, there exists a time $t_{\alpha, k} > 0$ such that the following holds for $\forall t \geq t_{\alpha, k}$:
\[\E[M_\eta^k(\theta_t - \theta^*)] \leq \E[M_\eta^k(\theta_{t_{\alpha, k}} - \theta^*)](1- \alpha(1 - \sqrt{\gamma}))^{t - t_{\alpha, k}} + c_k \alpha^k,\]
where $c_k$ is a  universal constant  that is  independent of $\alpha$ and $t$. We complete the proof of Lemma \ref{lemma: additive 2n moment}.

\begin{lemma}\label{lem: bound for T2}
Under the same setting as Lemma \ref{lemma: additive 2n moment}, we have
\begin{align*}
S_1 &\leq  \left( 1 - 2\alpha k (1-\sqrt{\gamma}) +  \mathcal{O}(\alpha^2)\right)\E[M_\eta^k(\theta_t - \theta^*)] + \mathcal{O}(\alpha^{k+1}),\\
S_2 &\leq  \mathcal{O}(\alpha^2) \cdot \E[M_\eta^k(\theta_t - \theta^*)] + \mathcal{O}(\alpha^{k+1}).
\end{align*}
\end{lemma}

\begin{proof}{Proof of Lemma \ref{lem: bound for T2}}
For $S_1$, by the independence between $x_t$ and $\theta_t$, we obtain
\begin{align*}
S_1 =& \left(1 - 2\alpha(1-\sqrt{\gamma}) + \alpha^2\right)^k\E[M_\eta^k(\theta_t - \theta^*)] + \E\big[\sum_{a+c = k,c \geq 2}\tbinom{k}{a}\tbinom{k-a}{c} T_1^aT_3^c\big]\\
\leq& \left(1 - 2\alpha(1-\sqrt{\gamma}) + \alpha^2\right)^k\E[M_\eta^k(\theta_t - \theta^*)]
+ \E\big[\sum_{a+c = k, c \geq 2}\mathcal{O}(\alpha^c) M_\eta^{a+\frac{c}{2}}(\theta_t - \theta^*)\|\tmT(x_t,\theta_t)-\tmT(x_t, \theta^*)\|_2^c\big]\\
&+ \E\big[\sum_{a+c = k, c \geq 2}\mathcal{O}(\alpha^c) M_\eta^{a+\frac{c}{2}}(\theta_t - \theta^*)\|\tmT(x_t, \theta^*) - \mT(\theta^*)\|_2^c\big]\\
&+ \E\big[\sum_{a+c = k, c \geq 2}\mathcal{O}(\alpha^c) M_\eta^{a+\frac{c}{2}}(\theta_t - \theta^*)\|\mT(\theta^*)-\mT( \theta_t)\|_2^c\big]\\
\leq& \left( 1 - 2\alpha k (1-\sqrt{\gamma}) +  \mathcal{O}(\alpha^2)\right)\E[M_\eta^k(\theta_t - \theta^*)]
+  \E\big[\sum_{a+c = k, c \geq 2}\mathcal{O}(\alpha^c) M_\eta^{a+\frac{c}{2}}(\theta_t - \theta^*) \big]\\
\overset{(\text{i})}{\leq}&  \left( 1 - 2\alpha k (1-\sqrt{\gamma}) +  \mathcal{O}(\alpha^2)\right)\E[M_\eta^k(\theta_t - \theta^*)] + \mathcal{O}(\alpha^{k+1}),
\end{align*}
where (i) holds by induction hypothesis and taking $t$ to be sufficiently large.

For $S_2$, we obtain
\begin{align*}
S_2\leq& \!\!\!\!\!\!\!\!\!\!\!\!\!\!\!\sum_{a+b+c +d= k, b+d \geq 1}   \!\!\!\!\!\!\!\!\!\!\!\!\!\!\!\!\!
\E\big[\mathcal{O}( \alpha^{2b+ c+2d})M_\eta^{a+\frac{c}{2}}(\theta_t - \theta^*)\|\tmT(x_t,\theta_t)-\tmT(x_t,\theta^*)\|_2^{2b}\|\tmT(x_t, \theta_t)-\mT(\theta_t)\|_2^c\|\tmT(x_t,\theta^*)-\mT(\theta^*)\|_2^{2d}\big] \\
\leq& \sum_{a+b+c +d= k, b+d \geq 1}\E\big[\mathcal{O}( \alpha^{2b+ c+2d})M_\eta^{a+\frac{c}{2}}(\theta_t - \theta^*)\|\tmT(x_t,\theta_t)-\tmT(x_t,\theta^*)\|_2^{2b+c}\|\tmT(x_t,\theta^*)-\mT(\theta^*)\|_2^{2d}\big] \\
&+ \sum_{a+b+c +d= k, b+d \geq 1}\E\big[\mathcal{O}( \alpha^{2b+ c+2d})M_\eta^{a+\frac{c}{2}}(\theta_t - \theta^*)\|\tmT(x_t,\theta_t)-\tmT(x_t,\theta^*)\|_2^{2b} \|\tmT(x_t,\theta^*)-\mT(\theta^*)\|_2^{2d+c}\big] \\
&+ \sum_{a+b+c +d= k, b+d \geq 1}\E\big[\mathcal{O}( \alpha^{2b+ c+2d})M_\eta^{a+c}(\theta_t - \theta^*)\|\tmT(x_t,\theta_t)-\tmT(x_t,\theta^*)\|_2^{2b} \|\tmT(x_t,\theta^*)-\mT(\theta^*)\|_2^{2d}\big] \\
\leq& \sum_{a+b+c +d= k, b+d \geq 1}\E\big[\mathcal{O}( \alpha^{2b+ c+2d})M_\eta^{a+b+c}(\theta_t - \theta^*)\big] \\
\overset{(\text{i})}{\leq}& \sum_{a+b+c +d= k, b \geq 1, d=0}\E\big[\mathcal{O}( \alpha^{2b+ c+2d})M_\eta^k(\theta_t - \theta^*)\big]+\sum_{a+b+c +d= k, b+d \geq 1} \mathcal{O}( \alpha^{a+3b+ 2c+2d}) \\
\leq& \mathcal{O}(\alpha^2)\E\big[M_\eta^k(\theta_t - \theta^*)\big] + \mathcal{O}(\alpha^{k+1}),
\end{align*}
where (i) holds by induction hypothesis and taking $t$ to be sufficiently large.
\end{proof}

\section{Proof of Theorem \ref{thm: additive convergence}}
\label{sec: additive convergence}

We prove the three properties stated in Theorem \ref{thm: additive convergence} in the next three subsections, respectively.

\subsection{Unique Limit Distribution}\label{sec:additive-unique-limit}

We consider a pair of coupled iterates, $\{\theta^{[1]}_t\}_{t \geq 0}$ and $\{\theta^{[2]}_t\}_{t \geq 0}$, defined as 
\begin{equation}\label{eq: additive couple}
\begin{aligned}
\theta_{t+1}^{[1]}  & = \theta_{t}^{[1]} +\alpha \big(\tmT(x_t, \theta_{t}^{[1]}) - \theta_{t}^{[1]}  \big)
\quad\text{and}\quad
\theta_{t+1}^{[2]}   = \theta_{t}^{[2]} +\alpha \big(\tmT(x_t, \theta_{t}^{[2]}) - \theta_{t}^{[2]}  \big).
\end{aligned}
\end{equation}
Here $\{\theta^{[1]}_t\}_{t \geq 0}$ and $\{\theta^{[2]}_t\}_{t \geq 0}$ are two iterates coupled by sharing the same noise sequence $\{x_t\}_{t \geq 0}$. The initial iterates $\theta^{[1]}_0$
and $\theta^{[2]}_0$ may depend on each other.

Taking the difference of the two equations in \eqref{eq: additive couple}, we obtain
\[\theta_{t+1}^{[1]}-\theta_{t+1}^{[2]} = (1-\alpha)(\theta_{t}^{[1]}-\theta_{t}^{[2]}) +  \alpha\big(\tmT(x_t, \theta_{t}^{[1]}) - \tmT(x_t, \theta_{t}^{[2]})\big).\]
Applying the  Moreau envelope $M_\eta(\cdot)$ defined in equation \eqref{eq: Moreau envelope} to both side of above equation and by Proposition \ref{prop: Moreau envelope}, we obtain
\begin{align*}
M_\eta(\theta_{t+1}^{[1]}-\theta_{t+1}^{[2]}) \leq& (1-\alpha)^2M_\eta(\theta_{t }^{[1]}-\theta_{t }^{[2]}) + \alpha(1-\alpha)\langle\nabla M_\eta(\theta_{t }^{[1]}-\theta_{t }^{[2]}), \tmT(x_t, 
 \theta_{t}^{[1]}) - \tmT(x_t, \theta_{t}^{[2]}) \rangle\\
&+ \frac{\alpha^2}{2\eta }\|\tmT(x_t, \theta_{t}^{[1]}) - \tmT(x_t, \theta_{t}^{[2]})\|_2^2.
\end{align*}
Taking expectation to both side of above equation, we have
\begin{align*}
\E[M_\eta(\theta_{t+1}^{[1]}-\theta_{t+1}^{[2]})] &\leq (1-2\alpha+\mathcal{O}(\alpha^2))\E[M_\eta(\theta_{t }^{[1]}-\theta_{t }^{[2]})] + \underbrace{ \alpha(1-\alpha)\E[\langle\nabla M_\eta(\theta_{t }^{[1]}-\theta_{t }^{[2]}),  \mT(\theta_{t}^{[1]}) - \mT(\theta_{t}^{[2]})\rangle]}_{T_1},
\end{align*}
where the above inequality holds by Assumption \ref{as:lip}($\bm{1}$). When $\alpha \leq 1$, we obtain
\begin{align*}
T_1 &\overset{(\text{i})}{\leq}  \alpha(1-\alpha)\E[\|\theta_{t }^{[1]}-\theta_{t }^{[2]}\|_m\|\mT(\theta_{t}^{[1]}) - \mT(\theta_{t}^{[2]})\|_m]\\
&\overset{(\text{ii})}{\leq} \frac{\alpha(1-\alpha)}{l_{cm}}\E[\|\theta_{t }^{[1]}-\theta_{t }^{[2]}\|_m\|\mT(\theta_{t}^{[1]}) - \mT(\theta_{t}^{[2]})\|_c]\\
&\leq \frac{\alpha(1-\alpha)\gamma}{l_{cm}}\E[\|\theta_{t }^{[1]}-\theta_{t }^{[2]}\|_m\|\theta_{t }^{[1]}-\theta_{t }^{[2]}\|_c]\\
&\overset{(\text{iii})}{\leq} \frac{2\alpha(1-\alpha)\gamma u_{cm}}{l_{cm}}\E[M_\eta(\theta_{t }^{[1]}-\theta_{t }^{[2]})] \overset{(\text{iv})}{\leq} 2\alpha\sqrt{\gamma} \E[M_\eta(\theta_{t }^{[1]}-\theta_{t }^{[2]})],
\end{align*}
where (i) follows from property (4) in Proposition \ref{prop: Moreau envelope}, (ii) and (iii) holds due to   (2) and (3) in Proposition \ref{prop: Moreau envelope}, and (iv) holds because of $\frac{u_{cm}}{l_{cm}} = \sqrt{\frac{1 + \eta u_{cs}^2}{1 + \eta l_{cs}^2}}$ by property (3) in Proposition \ref{prop: Moreau envelope} and that we can choose a sufficiently small $\eta>0$ such that $\frac{u_{cm}}{l_{cm}} \leq \frac{1}{\sqrt{\gamma}}$. Thus, there exists $\bar{\alpha}^\prime \leq \bar{\alpha}$ such that 
\begin{align*}
\E[M_\eta(\theta_{t+1}^{[1]}-\theta_{t+1}^{[2]})] &\leq (1-2\alpha(1-\sqrt{\gamma}) + \mathcal{O}(\alpha^2))\E[M_\eta(\theta_{t }^{[1]}-\theta_{t }^{[2]})] \\
&\leq (1-\alpha(1-\sqrt{\gamma}))\E[M_\eta(\theta_{t }^{[1]}-\theta_{t }^{[2]})],
\end{align*}
for all $\alpha \leq \bar{\alpha}^\prime$. Therefore, we have
\begin{equation}\label{eq: additive geo}
\begin{aligned}
W_2^2\big(\mathcal{L} (\theta_t^{[1]} ), \mathcal{L} (\theta_t^{[2]} )\big)  \leq 2u_{cm}^2\mathbb{E}\big[M_\eta(\theta_{t }^{[1]}-\theta_{t }^{[2]})\big] \leq 2u_{cm}^2\mathbb{E}\big[M_\eta(\theta_{0 }^{[1]}-\theta_{0 }^{[2]})\big](1-\alpha(1-\sqrt{\gamma}) )^t.
\end{aligned}    
\end{equation}
Note that equation \eqref{eq: additive geo} holds for any joint distribution of initial iterates $(  \theta^{[1]}_0, \theta^{[2]}_0)$. We use $\theta^{[2]}_{-1}$ to denote a random variable that satisfies $\theta^{[2]}_{-1} \overset{\textup{d}}{=} \theta^{[1]}_0,$ where $\overset{\textup{d}}{=}$ denotes equality in distribution, and $\theta^{[2]}_{-1}$ being independent of $\{x_t\}_{t \geq 0}$. Finally, we set $\theta^{[2]}_0$ as 
\begin{equation}\label{eq: additive watchback}
\theta_{0}^{[2]}   =  \theta_{-1}^{[2]} +\alpha\big(\tmT(x_{-1}, \theta_{-1}^{[2]}) - \theta_{-1}^{[2]}\big).
\end{equation}

Given that $\theta^{[2]}_{-1} \overset{\textup{d}}{=} \theta^{[1]}_0$ and $\theta^{[2]}_{-1}$ is independent with $\{x_t\}_{t \geq -1}$, we have $\theta^{[2]}_t \overset{\textup{d}}{=} \theta^{[1]}_{t+1}$ for all $t \geq 0$ by comparing the
dynamic of $(\theta^{[1]}_t)_{t \geq 0}$ and $(\theta^{[2]}_t)_{t \geq 0}$ as given in equations \eqref{eq: additive couple} and \eqref{eq: additive watchback}. We thus have  
\begin{align*}
W_2^2\big(\mathcal{L} ( \theta^{[1]}_t ), \mathcal{L} ( \theta^{[1]}_{t+1}  )\big)  =W_2^2\big(\mathcal{L} ( \theta^{[1]}_t ), \mathcal{L} ( \theta^{[2]}_t )\big)  \leq 2u_{cm}^2\mathbb{E}\big[M_\eta(\theta_{0 }^{[1]}-\theta_{0 }^{[2]})\big](1-\alpha(1-\sqrt{\gamma}) )^t,
\end{align*}
where the second inequality follows from equation \eqref{eq: additive geo}. It follows that
\begin{align*}
\sum_{t = 0}^{\infty} W_2^2\big(\mathcal{L} (\theta^{[1]}_t ), \mathcal{L} (\theta^{[1]}_{t+1} )\big) \leq    2u_{cm}^2\mathbb{E} \big[M_\eta(\theta_{0 }^{[1]}-\theta_{0 }^{[2]})\big] \sum_{t = 0}^{\infty} (1-\alpha(1-\sqrt{\gamma}) )^t <  \infty.
\end{align*}
Consequently, $\{\mathcal{L}(\theta^{[1]}_t )\}_{t \geq 0}$ forms a Cauchy sequence under the metric $W_2$. With the similar arguements in \cite[Appendix 2.2]{huo2023bias}, we conclude that the sequence $\{\mathcal{L}( \theta^{[1]}_t)\}_{t \geq 0}$ converges weakly to a limit distribution $\Bar{\mu} \in \mathcal{P}_2(\mathbb{R}^d  )$.

Suppose there exists another sequence $\{\widetilde{\theta}_t^{[1]}\}_{t \geq 0}$ with a different initial distribution that converges to a limit $\widetilde{\mu}$. By triangle inequality, we have
$$
W_2(\bar{\mu}, \tilde{\mu}) \leq W_2\big(\bar{\mu}, \mathcal{L} ( \theta_t^{[1]} )\big)+W_2\big(\mathcal{L} (\theta_t^{[1]} ), \mathcal{L} (\widetilde{\theta}_t^{[1]} )\big)+W_2\big(\mathcal{L} (\widetilde{\theta}_t^{[1]} ), \tilde{\mu}\big) \stackrel{t \rightarrow \infty}{\longrightarrow} 0.
$$
Note that the last step holds since $W_2\big(\mathcal{L} ( \theta_t^{[1]} ), \mathcal{L} (\widetilde{\theta}_t^{[1]} )\big) \stackrel{t \rightarrow \infty}{\longrightarrow} 0$ by equation \eqref{eq: additive geo}. We thus have $W_2(\Bar{\mu}, \widetilde{\mu}) = 0,$ which implies the uniqueness of the limit $\Bar{\mu}$.

Finally, the following lemma bounds the second moment of the limit random vector $\theta^{(\alpha)}$.
\begin{lemma}\label{lemma: additive variance}
Under Assumption \ref{as: additive noise}(\textbf{1}), when $\alpha \leq \bar{\alpha}^\prime$, we obtain 
\[\mathbb{E}[\Vert \theta^{(\alpha)} - \theta^*\Vert_2^2] \in \mathcal{O}(\alpha) \quad\text{and}\quad \mathbb{E}[\Vert \theta^{(\alpha)}\Vert_2^2] \in \mathcal{O}(1).\]

\end{lemma}

\begin{proof}{Proof of Lemma \ref{lemma: additive variance}}
We have shown that the sequence $\{\theta_t\}_{t \geq 0}$ converges weakly to $\theta^{(\alpha)}$ in $\mathcal{P}_2(\mathbb{R}^d)$. Since weak convergence in $\mathcal{P}_2(\mathbb{R}^d)$ is equivalent to convergence in distribution and the convergence of the first two moments, we obtain the limit
$
\mathbb{E}\big[\lVert \theta^{(\alpha)} - \theta^*\rVert_c^2\big] = \lim_{t\to\infty} \mathbb{E}\left[\lVert \theta_t - \theta^*\rVert_c^2\right]. 
$
Taking $t \rightarrow \infty$ on both sides of equation \eqref{eq: additive 2n moment} in Proposition~\ref{thm: additive 2n moment} with $n = 1$, and combining with the above limit gives
$
\mathbb{E}[\Vert \theta^{(\alpha)} - \theta^*\Vert_2^2] \leq \frac{1}{l_{cs}^2}\mathbb{E}[\lVert \theta^{(\alpha)} - \theta^*\rVert^2_c]\in \mathcal{O}(\alpha). 
$
Therefore, we have
$
\mathbb{E}[\Vert \theta^{(\alpha)}\Vert^2_2] 
\leq 2\mathbb{E}(\Vert \theta^{(\alpha)} - \theta^*\Vert^2_2)+ 2\Vert \theta^*\Vert^2_2 
\in \mathcal{O}(1).
$
\end{proof}

\subsection{Invariance}

Next, we will show that the unique limit distribution $\Bar{\mu}$ is also a stationary distribution for the Markov chain $\{\theta_t\}_{t \geq 0}$, as stated in the following lemma.
\begin{lemma}\label{lemma: additive invariance}
Let $\{\theta_t\}_{t \geq 0}$ and $\{\theta_t^{\prime}\}_{t \geq 0}$ be two trajectories of iterates coupled by equation \eqref{eq: additive couple}, where $\mathcal{L}(\theta_0) = \Bar{\mu}$ and $\mathcal{L}(\theta_0^{\prime}) \in  \mathcal{P}_2(  \mathbb{R}^{|\mS||\mA|}  )$ is arbitrary. We have
$
W^2_2 (\mathcal{L} (  \theta_1 ), \mathcal{L}(\theta_1^{\prime}) ) \leq \rho W^2_2 (\mathcal{L} (\theta_0 ), \mathcal{L}(\theta_0^{\prime}) ),
$
where the quantity $\rho: = \frac{u_{cm}^2}{l_{cm}^2}(1-\alpha(1-\sqrt{\gamma}))$ is independent of   $\mathcal{L}( \theta_0^{\prime})$. In particular, for any $t \geq 0$, if we set $\mathcal{L}( \theta_0^{\prime}) = \mathcal{L}(\theta_t)$, then
$
W^2_2 (\mathcal{L} (\theta_1 ), \mathcal{L}( \theta_{t+1}) ) \leq \rho W^2_2 (\Bar{\mu}, \mathcal{L}( \theta_t ) ).
$
\end{lemma}

\begin{proof}{Proof of Lemma \ref{lemma: additive invariance}}
We couple the two processes $\{ \theta_t\}_{t \geq 0}$ and $\{  \theta_t^{\prime} \}_{t \geq 0}$ such that 
$
W^2_2 (\mathcal{L} ( \theta_0 ), \mathcal{L}(  \theta_0^{\prime}) ) 
= \mathbb{E} [ \Vert \theta_0 - \theta^{\prime}_0\Vert_c^2 ].
$
Since $W_2$ is defined by infimum over all couplings, by defining $\rho:= \frac{u_{cm}^2}{l_{cm}^2}(1-\alpha(1-\sqrt{\gamma}))$, we have 
\begin{align*}
W^2_2 (\mathcal{L} ( \theta_1 ), \mathcal{L}( \theta_1^{\prime}) ) &\leq \mathbb{E} [  \Vert \theta_1 - \theta^{\prime}_1\Vert_c^2 ] \leq 2u_{cm}^2\mathbb{E} [  M_\eta( \theta_1 - \theta^{\prime}_1) ]\leq 2u_{cm}^2(1-\alpha(1-\sqrt{\gamma}) )\mathbb{E} [  M_\eta( \theta_0 - \theta^{\prime}_0) ]\\
&\leq \frac{u_{cm}^2}{l_{cm}^2}(1-\alpha(1-\sqrt{\gamma}))\mathbb{E} [  \Vert \theta_0 - \theta^{\prime}_0\Vert_c^2 ]=\rho W^2_2 (\mathcal{L} ( \theta_0 ), \mathcal{L}(  \theta_0^{\prime}) ).
\end{align*}
\end{proof}

By triangle inequality, we obtain
\begin{align*}
W_2 (\mathcal{L} (\theta_1 ), \bar{\mu} ) &\leq W_2 (\mathcal{L} (\theta_1 ), \mathcal{L} ( \theta_{t+1} ) )+W_2 (\mathcal{L} (\theta_{t+1} ), \bar{\mu} )
\leq \sqrt{\rho} W^2 (\Bar{\mu}, \mathcal{L}(\theta_t) )+W_2 (\mathcal{L} (\theta_{t+1} ), \bar{\mu} ) \stackrel{t \rightarrow \infty}{\longrightarrow} 0,
\end{align*}
where the second step holds by Lemma \ref{lemma: additive invariance} and last step follows from the weak convergence result. Therefore, we complete the proof that $\{\theta_t\}_{t \geq 0}$ converges to a unique stationary distribution $\bar{\mu}.$

\subsection{Convergence rate}

Consider the coupled processes defined in \eqref{eq: additive couple}. Suppose the initial iterate $\theta_0^{[2]}$ follows the stationary distribution $\Bar{\mu}$, thus $\mathcal{L}(\theta_t^{[2]}) = \Bar{\mu}=\mathcal{L}(\theta^{(\alpha)})$  for all $t \geq 0$. By equation \eqref{eq: additive geo}, we have for all $t \geq 0:$
\begin{align*}
W_2^2 (\mathcal{L}(\theta_t^{[1]}), \Bar{\mu} ) 
= W_2^2 (\mathcal{L}(\theta_t^{[1]}), \mathcal{L}(\theta_t^{[2]}) )
&\leq 2u_{cm}^2\mathbb{E} [M_\eta(\theta_{0 }^{[1]}-\theta_{0 }^{[2]}) ](1-\alpha(1-\sqrt{\gamma}) )^t\\
&= 2u_{cm}^2\mathbb{E} [M_\eta(\theta_{0 }^{[1]}-\theta^{(\alpha)}) ](1-\alpha(1-\sqrt{\gamma}) )^t .
\end{align*}
Lemma \ref{lemma: additive variance} states that the second moment of $\theta^{(\alpha)}$ is bounded by a constant. Combining this bound with above equation, we obtain the desired bound
$W^2_2(\mathcal{L}(\theta_t), \mu) \leq c \cdot (1-\alpha(1-\sqrt{\gamma}) )^t, $
where $c$ is a universal constant that is independent with $\alpha$ and $t$.

\section{Proof of Proposition \ref{prop:gaussian}}\label{sec: additive limit different noise}

We consider two sequences $\{Y^{(\alpha)}_t\}_{t\geq 0}$ and $\{Z_t^{(\alpha)}\}_{t\geq 0}$ derived by $Y^{(\alpha)}_t = \frac{\theta_t^{(\alpha)}-\theta^*}{\sqrt{\alpha}}$ and $Z_t^{(\alpha)}=\frac{\beta_t^{(\alpha)}-\theta^*}{\sqrt{\alpha}}$, where $\theta_t^{(\alpha)}$ and $\beta_t^{(\alpha)}$ are defined in Observation \ref{ob:bridging}. Therefore, $\{Y^{(\alpha)}_t\}_{t\geq 0}$ is associated with general noise $\{x_t\}_{t\geq0}$, and  $\{{Z_t}^{(\alpha)}\}_{t\geq 0}$ is associated with additive Gaussian distributed noise $\{w^\prime_t\}_{t\geq 0}.$ When the context is clear, we drop the superscript $^{(\alpha)}$ for the ease of exposition. By dynamics \eqref{eq: iid general SA2} and \eqref{eq: iid approximate}, We derive the dynamics for $\{Y_t\}_{t\geq 0}$ and $\{{Z_t}\}_{t\geq 0}$ as follows:
\begin{equation}\label{eq: additive different noise}
\begin{aligned}
Y_{t+1} &= (1-\alpha)Y_t + \sqrt{\alpha}\big(\tmT(x_t,\sqrt{\alpha} Y_t + \theta^*) - \tmT(x_t, \theta^*) + w_t(\theta^*)\big),\\
Z_{t+1}  &= (1-\alpha)Z_t  + \sqrt{\alpha}\big(\mT(\sqrt{\alpha} Z_t  + \theta^*) - \mT(\theta^*) + w_t^\prime\big),
\end{aligned}
\end{equation}
where $w_t(\theta^*)=\tmT(x_t, \theta^*)-\mT(\theta^*)$ and $ w_t^\prime$ are zero mean and have the same variance. When the context is clear, we use $w_t$ to denote $w_t(\theta^*)$ for the ease of exposition. Here $w_t$ and $w_t^\prime$ are not necessarily independent of each other. 
Note that $w_t$ has finite fourth moment by Assumption \ref{as: additive noise}($\bm{2}$). The specific coupling between $\{w^\prime_t\}_{t\geq 0}$ and $\{w_t\}_{t\geq 0}$ will be defined later.

Let $\kappa = \lfloor \alpha^{-\frac{1}{2}}\rfloor $. Direct calculation gives
\begin{align*}
&Y_{\kappa t+\kappa}   
= (1-\alpha )^\kappa Y_{\kappa t} + \sqrt{\alpha}\sum_{j = 1}^{\kappa}\big(\tmT(x_{\kappa t+\kappa-j}, \sqrt{\alpha} Y_{\kappa t} + \theta^*) - \tmT(x_{\kappa t+\kappa-j}, \theta^*)\big)\\
&+\sqrt{\alpha} \sum_{j=1}^\kappa \big(\tmT(x_{\kappa t+\kappa - j},\sqrt{\alpha} Y_{\kappa t+\kappa-j} + \theta^*)  - \tmT(x_{\kappa t+\kappa-j},\sqrt{\alpha} Y_{\kappa t} + \theta^*)  \big)\\
&+\sqrt{\alpha} \sum_{j=1}^\kappa \big((1-\alpha )^{j-1}-1\big)\big(\tmT(x_{\kappa t+\kappa - j},\sqrt{\alpha} Y_{\kappa t+\kappa-j} + \theta^*) - \tmT(x_{\kappa t+\kappa - j}, \theta^*) \big) 
+\sqrt{\alpha} \sum_{j=1}^\kappa (1-\alpha )^{j-1}w_{\kappa t+\kappa-j}.\\
&Z_{{\kappa} t+{\kappa}}   = (1-\alpha )^{\kappa} Z_{{\kappa} t}   + \sqrt{\alpha}{\kappa}\big(\mT(\sqrt{\alpha} Z_{{\kappa}t}   + \theta^*) - \mT(\theta^*)\big)\\
&+\sqrt{\alpha} \sum_{j=1}^{\kappa} \big(\mT(\sqrt{\alpha} Z_{{\kappa}t+{\kappa}-j}  + \theta^*) - \mT(\sqrt{\alpha} Z_{{\kappa}t}   + \theta^*) \big)\\
&+\sqrt{\alpha} \sum_{j=1}^{\kappa} \big((1-\alpha )^{j-1}-1\big)\big(\mT(\sqrt{\alpha} Z_{{\kappa}t+{\kappa}-j}   + \theta^*) - \mT(\theta^*) \big) 
+\sqrt{\alpha} \sum_{j=1}^{\kappa} (1-\alpha )^{j-1}w_{{\kappa}t+{\kappa}-j}^\prime .
\end{align*}
Taking the difference of the last two equations, we get
\begin{align*}
Y_{{\kappa} t+{\kappa}}-Z_{{\kappa} t+{\kappa}} = & (1-\alpha )^{\kappa}(Y_{{\kappa} t} - Z_{{\kappa} t} ) \\
&+ \sqrt{\alpha}\sum_{j = 1}^{\kappa}\tmT(x_{\kappa t+\kappa-j}, \sqrt{\alpha} Y_{\kappa t} + \theta^*) - \tmT(x_{\kappa t+\kappa-j}, \theta^*) - \mT(\sqrt{\alpha} Z_{{\kappa}t} + \theta^*) + \mT( \theta^*)\\
&+\sqrt{\alpha} \sum_{j=1}^\kappa \tmT(x_{\kappa t+\kappa - j},\sqrt{\alpha} Y_{\kappa t+\kappa-j} + \theta^*)  - \tmT(x_{\kappa t+\kappa-j},\sqrt{\alpha} Y_{\kappa t} + \theta^*)  \\
&-\sqrt{\alpha} \sum_{j=1}^{\kappa}  \mT(\sqrt{\alpha} Z_{{\kappa}t+{\kappa}-j}   + \theta^*) - \mT(\sqrt{\alpha} Z_{{\kappa}t}   + \theta^*) \\
&+\sqrt{\alpha} \sum_{j=1}^\kappa \big((1-\alpha )^{j-1}-1\big)\big(\tmT(x_{\kappa t+\kappa - j},\sqrt{\alpha} Y_{\kappa t+\kappa-j} + \theta^*) - \tmT(x_{\kappa t+\kappa - j}, \theta^*) \big) \\
&-\sqrt{\alpha} \sum_{j=1}^{\kappa} ((1-\alpha )^{j-1}-1)\left(\mT(\sqrt{\alpha} Z_{{\kappa}t+{\kappa}-j}   + \theta^*) - \mT(\theta^*)  \right) \\
&+\sqrt{\alpha} \sum_{j=1}^{\kappa} (w_{{\kappa}t+{\kappa}-j} - w_{{\kappa}t+{\kappa}-j}^\prime)
+\sqrt{\alpha} \sum_{j=1}^{\kappa} \big( (1-\alpha )^{j-1}-1\big)(w_{{\kappa}t+{\kappa}-j} - w_{{\kappa}t+{\kappa}-j}^\prime)\\
:= &(1-\alpha )^{\kappa}(Y_{{\kappa} t} - Z_{{\kappa} t} ) + A,
\end{align*}
where we collect in $A$ all but the first term on the RHS. Applying the  Moreau envelope $M_\eta(\cdot)$ defined in equation \eqref{eq: Moreau envelope} to both side of above equation and by Proposition \ref{prop: Moreau envelope}, we obtain
\begin{equation}\label{eq:additive-noise-M}
\mathbb{E}[M(Y_{{\kappa} t+{\kappa}}\!-\!Z_{{\kappa} t+{\kappa}} )] 
\leq (1\!-\!\alpha )^{2{\kappa}}\mathbb{E}[M(Y_{{\kappa} t} \!-\! Z_{{\kappa} t} )] 
+ (1\!-\!\alpha )^{\kappa}\underbrace{\mathbb{E}\langle \nabla M(Y_{{\kappa} t} \!-\! Z_{{\kappa} t} ), A\rangle}_{T_1} 
+ \frac{1}{2\eta}\underbrace{\mathbb{E}\|A\|_2^2}_{T_2}.
\end{equation}

The following lemmas, proved in Sections~\ref{sec:additive-limit-noise-T1} and~\ref{sec:additive-limit-noise-T2}, bound the $T_1$ and $T_2$ terms above.
\begin{lemma}
    \label{lem:additive-limit-noise-T1}
    Under the setting of Theorem~\ref{thm: additive limit}, there exists $t_\alpha'>0$ such that
    $
    T_1 \leq (1+\sqrt{\gamma})\alpha {\kappa} \E[M_\eta(Y_{{\kappa} t} - Z_{{\kappa} t} )] + {\kappa}^{2}\cdot\mathcal{O}(\alpha^{2})
    $ for  sufficiently large $t \geq t_\alpha'$.
\end{lemma}
\begin{lemma}
    \label{lem:additive-limit-noise-T2}
    Under the setting of Theorem~\ref{thm: additive limit}, there exist $t_\alpha''>0$ and a proper coupling between $\{w_t\}_{t \geq 0}$ and $\{w_t'\}_{t \geq 0}$ such that
    $
    T_2 \in \mathcal{O}(\alpha)
    $ for sufficiently large $t \geq t_\alpha''$.
\end{lemma}

Plugging the above bounds on $T_1$ and $T_2$ into equation \eqref{eq:additive-noise-M}, there exist an $\alpha_0$ such that for any $\alpha \leq  \alpha_0$, there exist $t_\alpha$ such
that for any $t \geq  t_\alpha$,  it holds that
\begin{align*}
\E[M_\eta(Y_{{\kappa}t+{\kappa}} -  Z_{{\kappa}t+{\kappa}}  ) ] &\leq \big((1-\alpha)^{2\kappa} + (1+\sqrt{\gamma})\alpha {\kappa} \big)\E[M_\eta(Y_{{\kappa}t} -  Z_{{\kappa}t}  ) ] + \mathcal{O}(\alpha)\\
&\leq \big(1- (1-\sqrt{\gamma})\alpha {\kappa} \big)\E[M_\eta(Y_{{\kappa}t} -  Z_{{\kappa}t}  ) ] + \mathcal{O}(\alpha).
\end{align*}
Therefore, we obtain
$
\lim_{t \to \infty} \E[M_\eta(Y_{{\kappa}t} -  Z_{{\kappa}t}  ) ] \in \mathcal{O}(\alpha^{\frac{1}{2}}).
$
By triangle inequality, we have
\begin{align*}
W_2\big(\mathcal{L}(Y^{(\alpha
)}), \mathcal{L}({Z }^{(\alpha
)})\big) &\leq  \lim_{t \to \infty}W_2\big(\mathcal{L}(Y^{(\alpha
)}), \mathcal{L}(Y_{{\kappa}t})\big) +W_2\big(\mathcal{L}(Y_{{\kappa}t}), \mathcal{L}(Z _{{\kappa}t})\big) +W_2\big(\mathcal{L}(Z _{{\kappa}t}), \mathcal{L}({Z }^{(\alpha
)})\big)\\
&\leq \lim_{t \to \infty} \sqrt{\mathbb{E}[\|Y_{{\kappa}t} - Z_{{\kappa}t}\|_c^2]}\leq \lim_{t \to \infty} \sqrt{2u_{cm}^2\mathbb{E}[M(Y_{{\kappa}t} - Z_{{\kappa}t})]} \in \mathcal{O}(\alpha^{\frac{1}{4}}),
\end{align*}
which completes the proof of Proposition \ref{prop:gaussian}.

\subsection{Proof of Lemma \ref{lem:additive-limit-noise-T1} on $T_1$}\label{sec:additive-limit-noise-T1}
By property (4) in Proposition \ref{prop: Moreau envelope}, and the fact that $w_{{\kappa}t+{\kappa}-j}$ and $ w_{{\kappa}t+{\kappa}-j}^\prime$ are zero mean noise and independent of $Y_{{\kappa} t}$ and $ Z_{{\kappa} t}$, we obtain that $T_1$ is upper bounded by
\begin{align}
& {\E[\|Y_{{\kappa} t} - Z_{{\kappa} t} \|_{m}\|\sqrt{\alpha}{\kappa}(\mT(\sqrt{\alpha} Y_{{\kappa}t} + \theta^*) - \mT(\sqrt{\alpha} Z_{{\kappa}t}  + \theta^*))\|_m]} \tag{$T_{11}$}\\
& + {\E\bigg[\|Y_{{\kappa} t} - Z_{{\kappa} t} \|_{m}\|\sqrt{\alpha} \sum_{j=1}^{\kappa} \left(\mT(\sqrt{\alpha} Y_{{\kappa}t+{\kappa}-j}  + \theta^*) - \mT(\sqrt{\alpha} Y_{{\kappa}t}  + \theta^*) \right)\|_m\bigg]} \tag{$T_{12}$}\\
&+ {\E\bigg[\|Y_{{\kappa} t} - Z_{{\kappa} t} \|_m\|\sqrt{\alpha} \sum_{j=1}^{\kappa} \left(  \mT(\sqrt{\alpha} Z_{{\kappa}t+{\kappa}-j}   + \theta^*) - \mT(\sqrt{\alpha} Z_{{\kappa}t}   + \theta^*) \right)\|_m\bigg]} \tag{$T_{13}$}\\
&+ {\E\bigg[\|Y_{{\kappa} t} - Z_{{\kappa} t} \|_m\|\sqrt{\alpha} \sum_{j=1}^{\kappa} ((1-\alpha )^{j-1}-1)\left(\mT(\sqrt{\alpha} Y_{{\kappa}t+{\kappa}-j}  + \theta^*) - \mT(\sqrt{\alpha} Z_{{\kappa}t+{\kappa}-j}   + \theta^*)  \right)\|_m \bigg]} \tag{$T_{14}$}.
\end{align}

\textbf{The $T_{11}$ term: }
By property (3) in Proposition \ref{prop: Moreau envelope}, we have
\begin{align*}
T_{11} &\leq \frac{\sqrt{\alpha}\kappa}{l_{cm}}{\E[\|Y_{{\kappa} t} - Z_{{\kappa} t} \|_{m}\|\mT(\sqrt{\alpha} Y_{{\kappa}t} + \theta^*) - \mT(\sqrt{\alpha} Z_{{\kappa}t}  + \theta^*)\|_c]} \leq \frac{\alpha\kappa\gamma}{l_{cm}}{\E[\|Y_{{\kappa} t} - Z_{{\kappa} t} \|_{m}\|Y_{{\kappa} t} - Z_{{\kappa} t} \|_{c}]}\\
&\leq \frac{\alpha\kappa\gamma u_{cm}}{l_{cm}}{\E[\|Y_{{\kappa} t} - Z_{{\kappa} t} \|_{m}^2]}= \frac{2\alpha {\kappa} \gamma u_{cm}}{l_{cm}} \E[M_\eta(Y_{{\kappa} t} - Z_{{\kappa} t} )]
\leq 2\alpha {\kappa} \sqrt{\gamma}\E[M_\eta(Y_{{\kappa} t} - Z_{{\kappa} t} )],
\end{align*}
where the last inequality holds because we can always choose a proper $\eta$ such that $\frac{u_{cm}}{l_{cm}} \leq \frac{1}{\sqrt{\gamma}}$.

\textbf{The $T_{12}$ and $T_{13}$ terms: }
We begin with the bound
$
T_{12} \leq \frac{\alpha \gamma }{l_{cm}} \sum_{j=1}^{\kappa} \E[\|Y_{{\kappa} t} - Z_{{\kappa} t} \|_m\|Y_{{\kappa}t+{\kappa}-j} - Y_{{\kappa}t}\|_c].
$
By equation \eqref{eq: additive different noise}, we obtain
\begin{align*}
&\|Y_{{\kappa}t+{\kappa}-j} - Y_{{\kappa}t} \|_c\\
=& \Big\|\big((1-\alpha)^{{\kappa}-j} - 1\big)Y_{{\kappa}t} + \sqrt{\alpha}\sum_{l=1}^{{\kappa}-j}(1-\alpha)^{l-1}\big(\tmT(x_{{\kappa}t+{\kappa}-j-l}, \sqrt{\alpha}Y_{{\kappa}t+{\kappa}-j-l} + \theta^*) - \tmT(x_{{\kappa}t+{\kappa}-j-l},\theta^*) + w_{{\kappa}t+{\kappa}-j-l}  \big)\Big\|_c.
\end{align*}
Therefore, $T_{12}$  can be bounded by the sum of three terms as follows:
\begin{align}
T_{12} \leq& {\frac{\alpha\gamma}{l_{cm}}\sum_{j = 1}^{{\kappa}}\E[\|Y_{{\kappa} t} - Z_{{\kappa} t} \|_m\|((1-\alpha)^{{\kappa}-j} - 1)Y_{{\kappa}t} \|_c]} \tag{$T_{121}$}\\
&+ {\frac{\alpha\gamma}{ l_{cm}}\sum_{j = 1}^{{\kappa}}\E\bigg[\|Y_{{\kappa} t} - Z_{{\kappa} t}  \|_m\|\sqrt{\alpha}\sum_{l=1}^{{\kappa}-j}(1-\alpha)^{l-1}\big(\tmT(x_{{\kappa}t+{\kappa}-j-l}, \sqrt{\alpha}Y_{{\kappa}t+{\kappa}-j-l} + \theta^*) - \tmT(x_{{\kappa}t+{\kappa}-j-l},\theta^*) \big)\|_c\bigg]} \tag{$T_{122}$}\\
&+ {\frac{\alpha\gamma}{ l_{cm}}\sum_{j = 1}^{{\kappa}}\E\bigg[\|Y_{{\kappa} t} - Z_{{\kappa} t} \|_m\|\sqrt{\alpha}\sum_{l=1}^{{\kappa}-j}(1-\alpha)^{l-1} w_{{\kappa}t+{\kappa}-j-l}\|_c\bigg]} \tag{$T_{123}$}.
\end{align}
To proceed, we invoke Proposition~\ref{thm: additive 2n moment}, which guarantees that for any norm $\|\cdot\|$ and each fixed $\alpha>0$, we have $\E[\|Y_t\|^n]\in \mathcal{O}(1)$ and $\E[\|Z_t\|^n] \in \mathcal{O}(1)$ for all sufficiently large $t\ge t_{\alpha,n}$. 
\begin{remark}\label{rem:uniform_in_t_appendix}
In this proof and those for Theorems~\ref{thm: additive limit} and \ref{thm: additive smooth}, we work in the asymptotic regime where we first set \(t \to \infty\) with $\alpha$ fixed, and then $\alpha\to0$. Therefore, Proposition~\ref{thm: additive 2n moment} suffices even though $t_{\alpha,n}$ may depend on $\alpha$. Throughout the rest of the e-companion, we always take \(t\) to be large enough and do not explicit state the threshold $t_{\alpha,n}$. Also see Remark~\ref{rem:unifrom_in_t}.
\end{remark}
Therefore, we have
\begin{align*}
T_{121}&\leq  \frac{\alpha \gamma u_{cm}}{ l_{cm}}\E[\|Y_{{\kappa} t} - Z_{{\kappa} t} \|_m\|Y_{{\kappa}t} \|_m]\sum_{j = 1}^{{\kappa}}(1-(1-\alpha)^{{\kappa}-j} ) \\
& \in \mathcal{O}(\alpha \sum_{j = 1}^{{\kappa}}(1-(1-\alpha)^{{\kappa}-j}  )) \sqrt{\E[M_\eta(Y_{{\kappa} t} - Z_{{\kappa} t} )]} \\
&\in \mathcal{O}(\alpha {\kappa} - 1 +(1-\alpha  )^{\kappa}) \sqrt{\E[M_\eta(Y_{{\kappa} t} - Z_{{\kappa} t} )]}\\
&\in \mathcal{O}(\alpha {\kappa} - 1 +(1-\frac{\alpha {\kappa}}{{\kappa}} )^{\kappa}) \sqrt{\E[M_\eta(Y_{{\kappa} t} - Z_{{\kappa} t} )]} \overset{(\text{i})}{\in} \mathcal{O}(\alpha^2) \cdot {\kappa}^2\sqrt{\E[M_\eta(Y_{{\kappa} t} - Z_{{\kappa} t} )]},
\end{align*}
where (i) holds by equation \eqref{eq: alpha order}.
We also have
\begin{align*}
T_{122} &\leq \frac{\alpha\gamma}{ l_{cm}}\sum_{j = 1}^{{\kappa}}\sum_{l=1}^{{\kappa}-j}\E[\|Y_{{\kappa} t} - Z_{{\kappa} t}  \|_m\|\sqrt{\alpha}(1-\alpha)^{l-1}\big(\tmT(x_{{\kappa}t+{\kappa}-j-l}, \sqrt{\alpha}Y_{{\kappa}t+{\kappa}-j-l} + \theta^*) - \tmT(x_{{\kappa}t+{\kappa}-j-l},\theta^*)  \big)\|_c]\\
&\leq \frac{\alpha\gamma L}{ l_{cm}}\sum_{j = 1}^{{\kappa}}\sum_{l=1}^{{\kappa}-j}\alpha(1-\alpha)^{l-1}\E[\|Y_{{\kappa} t} - Z_{{\kappa} t}  \|_m\|Y_{{\kappa}t+{\kappa}-j-l}\|_c]\\
&\leq \mathcal{O}(\frac{\alpha^2\gamma L}{ l_{cm}}\sum_{j = 1}^{{\kappa}}\sum_{l=1}^{{\kappa}-j}(1-\alpha)^{l-1})\sqrt{\E[M_\eta(Y_{{\kappa} t} - Z_{{\kappa} t} )]}\\
&\overset{\text{(i)}}{\in} \mathcal{O}( \alpha^2\sum_{j = 1}^{{\kappa}}\sum_{l=1}^{{\kappa}-j}(1-\alpha)^{l-1})\sqrt{\E[M_\eta(Y_{{\kappa} t} - Z_{{\kappa} t} )]}\in \mathcal{O}(\alpha^2)\cdot {\kappa}^2\sqrt{\E[M_\eta(Y_{{\kappa} t} - Z_{{\kappa} t} )]}.
\end{align*}
Lastly, for term $T_{123}$, we have 
\begin{align*}
T_{123} &\overset{\text{(i)}}{=} \frac{\alpha\gamma}{ l_{cm}}\sum_{j = 1}^{{\kappa}}\E[\|Y_{{\kappa}t} - Z_{{\kappa}t}  \|_m] \E[\|\sqrt{\alpha}\sum_{l=1}^{{\kappa}-j}(1-\alpha)^{l-1} w_{{\kappa}t+{\kappa}-j-l}\|_c]\\
&\leq \mathcal{O}(1) \cdot \alpha^{\frac{3}{2}} \sum_{j = 1}^{{\kappa}}\E[\|\sum_{l=1}^{{\kappa}-j}(1-\alpha)^{l-1} w_{{\kappa}t+{\kappa}-j-l}\|_2]\sqrt{\E[M_\eta(Y_{{\kappa} t} - Z_{{\kappa} t} )]}\\
&\leq \mathcal{O}(1) \cdot \alpha^{\frac{3}{2}} \sum_{j = 1}^{{\kappa}}\sqrt{\E[\|\sum_{l=1}^{{\kappa}-j}(1-\alpha)^{l-1} w_{{\kappa}t+{\kappa}-j-l}\|_2^2]}\sqrt{\E[M_\eta(Y_{{\kappa} t} - Z_{{\kappa} t} )]}\\
&\leq  \mathcal{O}(1) \cdot \alpha^{\frac{3}{2}} \sum_{j =1}^{{\kappa}}\sqrt{\sum_{l=1}^{{\kappa}-j}(1-\alpha)^{2l-2}}\sqrt{\E[M_\eta(Y_{{\kappa} t} - Z_{{\kappa} t} )]}\\
&\leq \mathcal{O}(1) \cdot \alpha^{\frac{3}{2}} {\sum_{j = 1}^{{\kappa}}\sqrt{{\kappa}-j}}\sqrt{\E[M_\eta(Y_{{\kappa} t} - Z_{{\kappa} t} )]}\\
&\overset{\text{(ii)}}{\in} \mathcal{O}(\alpha^{\frac{3}{2}})\cdot {\kappa}^{\frac{3}{2}}\sqrt{\E[M_\eta(Y_{{\kappa} t} - Z_{{\kappa} t} )]},
\end{align*}
where (i) holds because $Y_{{\kappa}t}$ and $ Z_{{\kappa}t} $ are independent with $w_{{\kappa}t+{\kappa}-j-l}$ and (ii) holds since ${\sum_{j = 1}^{{\kappa}}\sqrt{{\kappa}-j}} {\in \mathcal{O}({\kappa}^{\frac{3}{2}})}.$ Combining the bounds on $T_{121}, T_{122}$ and $T_{123}$, we obtain $T_{12} \in \mathcal{O}(\alpha^{\frac{3}{2}})\cdot {\kappa}^{\frac{3}{2}}\sqrt{\E[M_\eta(Y_{{\kappa} t} - Z_{{\kappa} t} )]}$. By Cauchy inequality, $T_{12} \leq \frac{1}{2}(1-\gamma)\alpha\kappa \E[M_\eta(Y_{{\kappa} t} - Z_{{\kappa} t} )] + \mathcal{O}(\alpha^2)\cdot {\kappa}^2 $. Similarly, we have $T_{13}\leq \frac{1}{2}(1-\gamma)\alpha\kappa \E[M_\eta(Y_{{\kappa} t} - Z_{{\kappa} t} )] + \mathcal{O}(\alpha^2)\cdot {\kappa}^2$.

\textbf{The $T_{14}$ Term:} 
We have the bound
\begin{align*}
T_{14} &\leq \frac{\alpha}{l_{cm}}\sum_{j=1}^{\kappa} ((1-\alpha )^{j-1}-1) {\E[\|Y_{{\kappa} t} - Z_{{\kappa} t} \|_m\|Y_{{\kappa}t+{\kappa}-j}  -  Z_{{\kappa}t+{\kappa}-j} \|_c]} \\
& \overset{\text{(i)}}{\leq} \mathcal{O}(1) \cdot \alpha \sum_{j=1}^{\kappa} (1-(1-\alpha )^{j-1})\leq \mathcal{O}(1) \cdot (1-\alpha {\kappa} -(1-\alpha  )^{\kappa}) \in \mathcal{O}(\alpha^2 )\cdot {\kappa}^2,
\end{align*}
where in (i) we use ${\E[\|Y_{{\kappa} t} - Z_{{\kappa} t} \|_m\|Y_{{\kappa}t+{\kappa}-j}  -  Z_{{\kappa}t+{\kappa}-j}  \|_c]} {\in \mathcal{O}(1)}$. Combining the bound for $T_{11} \sim T_{14}$, we obtain
$
T_1 \leq (1+\sqrt{\gamma})\alpha {\kappa} \E[M_\eta(Y_{{\kappa} t} - Z_{{\kappa} t} )] + \mathcal{O}(\alpha^2 )\cdot {\kappa}^2\
$ and finish the proof of Lemma \ref{lem:additive-limit-noise-T1}.

\subsection{Proof of Lemma \ref{lem:additive-limit-noise-T2} on $T_2$}\label{sec:additive-limit-noise-T2}
Observe that
\begin{align}
T_2\leq &8\alpha\E[\|\sum_{j = 1}^{\kappa}\tmT(x_{\kappa t+\kappa-j}, \sqrt{\alpha} Y_{\kappa t} + \theta^*) - \tmT(x_{\kappa t+\kappa-j}, \theta^*)\|_2^2] \tag{$T_{21}$}\\
&+ {8 \alpha {\kappa}^2 \E[\|\mT(\sqrt{\alpha} Z_{{\kappa}t} + \theta^*) - \mT( \theta^*)\|_2^2]} \tag{$T_{22}$}\\
&+ {8 \alpha \E[\|\sum_{j=1}^{\kappa} \tmT(x_{{\kappa}t+{\kappa}-j}, \sqrt{\alpha} Y_{{\kappa}t+{\kappa}-j}  + \theta^*) - \tmT(x_{{\kappa}t+{\kappa}-j}, \sqrt{\alpha} Y_{{\kappa}t}  + \theta^*) \|_2^2]} \tag{$T_{23}$}\\
&+ {8 \alpha \E[\|\sum_{j=1}^{\kappa} \mT(\sqrt{\alpha} Z_{{\kappa}t+{\kappa}-j}  + \theta^*) - \mT(\sqrt{\alpha} Z_{{\kappa}t}  + \theta^*) \|_2^2]} \tag{$T_{24}$}\\
&+ {8 \alpha \E[\| \sum_{j=1}^{\kappa} ((1-\alpha )^{j-1}-1)\left(\tmT(x_{{\kappa}t+{\kappa}-j}, \sqrt{\alpha} Y_{{\kappa}t+{\kappa}-j}  + \theta^*) - \tmT(x_{{\kappa}t+{\kappa}-j},  \theta^*)  \right)\|_2^2]} \tag{$T_{25}$} \\
&+ {8 \alpha \E[\| \sum_{j=1}^{\kappa} ((1-\alpha )^{j-1}-1)\left(\mT(\sqrt{\alpha} Z_{{\kappa}t+{\kappa}-j}    + \theta^*) - \mT(\theta^*)  \right)\|_2^2]} \tag{$T_{26}$} \\
&+ {8 \alpha \E[\|\sum_{j=1}^{\kappa} (w_{{\kappa}t+{\kappa}-j} - w_{{\kappa}t+{\kappa}-j}^\prime)\|_2^2]} \tag{$T_{27}$} \\
& +{8 \alpha \E[\| \sum_{j=1}^{\kappa} ( (1-\alpha )^{j-1}-1)(w_{{\kappa}t+{\kappa}-j} - w_{{\kappa}t+{\kappa}-j}^\prime)\|_2^2]} \tag{$T_{28}$}.
\end{align}

\textbf{The $T_{21}$ - $T_{26}$ Terms:}
We have
$
T_{21} \leq \frac{8 \alpha^2 {\kappa}^2 L^2}{l_{cm}^2}\E[\|Y_{{\kappa}t} \|_2^2] \in \mathcal{O}(\alpha^2 )\cdot {\kappa}^2
$
and
$
T_{22} \leq \frac{8 \alpha^2 {\kappa}^2 \gamma^2}{l_{cm}^2}\E[\|Z_{{\kappa}t}   \|_c^2]
\in \mathcal{O}(\alpha^2 )\cdot {\kappa}^2.
$
\begin{align*}
T_{23} &\leq 8 \alpha {\kappa} \sum_{j=1}^{\kappa}\E[\| \tmT(x_{{\kappa}t+{\kappa}-j}, \sqrt{\alpha} Y_{{\kappa}t+{\kappa}-j}  + \theta^*) - \tmT(x_{{\kappa}t+{\kappa}-j}, \sqrt{\alpha} Y_{{\kappa}t}  + \theta^*)\|_2^2]\\
&\leq \frac{8 \alpha^2 {\kappa} \gamma^2}{l_{cs}^2} \sum_{j=1}^{\kappa} {\E[\|Y_{{\kappa}t+{\kappa}-j}-Y_{{\kappa}t} \|_c^2]}\in \mathcal{O}(\alpha^2 )\cdot {\kappa}^2.\\
T_{24} &\leq 8 \alpha {\kappa} \sum_{j=1}^{\kappa}\E[\| \mT( \sqrt{\alpha} Z_{{\kappa}t+{\kappa}-j}   + \theta^*) - \mT(\sqrt{\alpha} Z_{{\kappa}t}   + \theta^*)\|_2^2]\\
&\leq \frac{8 \alpha^2 {\kappa} \gamma^2}{l_{cs}^2} \sum_{j=1}^{\kappa} {\E[\|Z_{{\kappa}t+{\kappa}-j} -Z_{{\kappa}t}  \|_c^2]}\in \mathcal{O}(\alpha^2 )\cdot {\kappa}^2.\\
T_{25} &\leq 8 \alpha {\kappa} \textstyle\sum_{j=1}^{\kappa} \E[\|  ((1-\alpha )^{j-1}-1)\big(\tmT(x_{{\kappa}t+{\kappa}-j}, \sqrt{\alpha} Y_{{\kappa}t+{\kappa}-j}  + \theta^*) - \tmT(x_{{\kappa}t+{\kappa}-j},  \theta^*)  \big)\|_2^2]\\
&\leq \frac{8 \alpha^2L^2 {\kappa}}{l_{cs}^2}\sum_{j=1}^{\kappa}  \E[\|  ((1-\alpha )^{j-1}-1)  Y_{{\kappa}t+{\kappa}-j}   \|_c^2] \leq \mathcal{O}(\alpha^2 ) \cdot {\kappa} \sum_{j=1}^{\kappa}((1-\alpha )^{j-1}-1) \in \mathcal{O}(\alpha^3 )\cdot {\kappa}^3.\\
T_{26} &\leq 8 \alpha {\kappa} \sum_{j=1}^{\kappa} \E[\|  ((1-\alpha )^{j-1}-1)\left(\mT(\sqrt{\alpha} Z_{{\kappa}t+{\kappa}-j}   + \theta^*) - \mT(\theta^*)  \right)\|_2^2]\\
&\leq \frac{8 \alpha^2\gamma^2 {\kappa}}{l_{cs}^2}\sum_{j=1}^{\kappa}  \E[\|  ((1-\alpha )^{j-1}-1)  Z_{{\kappa}t+{\kappa}-j}    \|_c^2]\leq \mathcal{O}(\alpha^2 ) \cdot {\kappa} \sum_{j=1}^{\kappa}((1-\alpha )^{j-1}-1) \in \mathcal{O}(\alpha^3 )\cdot {\kappa}^3.
\end{align*}

\textbf{The $T_{27}$ Term:}
First observe that
$
T_{27} = 8 \alpha {\kappa} \E[\| \frac{1}{\sqrt{{\kappa}}}\sum_{j=1}^{\kappa} w_{{\kappa}t+{\kappa}-j} - \frac{1}{\sqrt{{\kappa}}}\sum_{j=1}^{\kappa} w_{{\kappa}t+{\kappa}-j}^\prime)\|_2^2].
$
To proceed, we use Theorem 1 in \cite{bonis2020stein}, which is restated in the following lemma.
\begin{lemma}[Theorem 1 in \cite{bonis2020stein}]\label{lemma: Berry-Esseen}
Let $X_1, \dots, X_n$ be n i.i.d random variables taking values in $\R^d$ with zero mean and identity variance matrix. Let $\nu$ be the d-dimensional standard Gaussian measure and $X_1^\prime, \dots, X_n^\prime$ be n i.i.d random variables distributed as $\nu$. Assume that $\E[\|X_1\|_2^4] < \infty$. Let $S_n = \frac{X_1 + \dots + X_n}{\sqrt{n}}$ and $S_n^\prime = \frac{X_1^\prime + \dots + X_n^\prime}{\sqrt{n}}$ Then, we have
$
W_{2,2}\big(\mathcal{L}(S_n), \mathcal{L}(S_n^\prime)\big) = W_{2,2}\big(\mathcal{L}(S_n), \nu\big) \in \mathcal{O}(\frac{1}{\sqrt{n}}),
$
where $W_{2,2}$ denotes the Wasserstein distances of order 2 with $\ell_2$-norm.
\end{lemma}
Let $\Sigma := \operatorname{Var}(w_1)$. Since $\Sigma\succeq 0$, let $r:=\operatorname{rank}(\Sigma)\le d$
and take a eigen-decomposition
\[
\Sigma = U V U^\top,
\qquad
U\in\R^{d\times r},\ \ U^\top U = I_r,
\qquad
V=\operatorname{diag}(\lambda_1,\ldots,\lambda_r)\succ 0 .
\]
For each $j\in[\kappa]$, define
\[
X_j \;:=\; V^{-1/2}U^\top w_{\kappa t+\kappa-j}\in\R^r .
\]
Then, writing $P:=UU^\top$ for the orthogonal projector onto $\operatorname{Range}(\Sigma)$,
we have the identity
\[
U V^{1/2}X_j
= U V^{1/2}V^{-1/2}U^\top w_{\kappa t+\kappa-j}
= P\, w_{\kappa t+\kappa-j}.
\]
Consequently,
\begin{equation}\label{eq:proj-sum}
\frac{1}{\sqrt{\kappa}}\sum_{j=1}^{\kappa} U V^{1/2} X_{j}
=
P\,\frac{1}{\sqrt{\kappa}}\sum_{j=1}^{\kappa} w_{\kappa t+\kappa-j}.
\end{equation}
Moreover, for any $v\in\ker(\Sigma)$,
\[
\operatorname{Var}(v^\top w_1)=v^\top \Sigma v = 0
\quad\Longrightarrow\quad
v^\top w_1 = \E[v^\top w_1]=0 \ \ \text{a.s.}
\]
and hence
\[
v^\top w_n = 0 \ \ \text{a.s. for all }n.
\]
Therefore $w_n\in \ker(\Sigma)^\perp = \operatorname{Range}(\Sigma)$ almost surely, so
$P w_n = w_n$ a.s. In particular,
\[
P\,\frac{1}{\sqrt{\kappa}}\sum_{j=1}^{\kappa} w_{\kappa t+\kappa-j}
=
\frac{1}{\sqrt{\kappa}}\sum_{j=1}^{\kappa} w_{\kappa t+\kappa-j}
\quad\text{a.s.}
\]
Combining with \eqref{eq:proj-sum} gives the identity
\[
\frac{1}{\sqrt{\kappa}}\sum_{j=1}^{\kappa} U V^{1/2} X_{j}
=
\frac{1}{\sqrt{\kappa}}\sum_{j=1}^{\kappa} w_{\kappa t+\kappa-j}
\quad\text{a.s.}
\]
and hence, in particular,
\[
\mathcal L(\frac{1}{\sqrt{\kappa}}\sum_{j=1}^{\kappa} U V^{1/2} X_{j})
=
\mathcal L(\frac{1}{\sqrt{\kappa}}\sum_{j=1}^{\kappa} w_{\kappa t+\kappa-j}).
\]
Similarly, for each $j\in[\kappa]$, define the Gaussian variables $X_j':= V^{-1/2}U^\top w_{\kappa t+\kappa-j}'\in\R^r$ for any $j \in [\kappa]$ and have
\[
\mathcal L(\frac{1}{\sqrt{\kappa}}\sum_{j=1}^{\kappa} U V^{1/2} X_{j}')
=
\mathcal L(\frac{1}{\sqrt{\kappa}}\sum_{j=1}^{\kappa} w_{\kappa t+\kappa-j}').
\]
Then, by Lemma~\ref{lemma: Berry-Esseen} and the definition of $W_{2,2}$, there
exists a coupling of 
$\{w_{\kappa t+\kappa-j}\}_{j\in[\kappa]}$ and $\{w'_{\kappa t+\kappa-j}\}_{j\in[\kappa]}$
such that
\begin{align*}
T_{27} 
&= 8 \alpha {\kappa} \cdot W_{2,2}^2 \big(\law(\frac{1}{\sqrt{{\kappa}}}\sum_{j=1}^{\kappa} w_{{\kappa}t+{\kappa}-j}), \law( \frac{1}{\sqrt{{\kappa}}}\sum_{j=1}^{\kappa} w_{{\kappa}t+{\kappa}-j}^\prime)\big)\\
&= 8 \alpha {\kappa} \cdot  W_{2,2}^2 \big(\law(\frac{1}{\sqrt{{\kappa}}}\sum_{j=1}^{\kappa} UV^{1/2}X_j), \law(\frac{1}{\sqrt{{\kappa}}}\sum_{j=1}^{\kappa} UV^{1/2}X_j')\big)\\
&\leq 8 \alpha {\kappa} \|U\|_2^2\|V^{\frac{1}{2}}\|_2^2  \cdot W_{2,2}^2 \big(\law(\frac{1}{\sqrt{{\kappa}}}\sum_{j=1}^{\kappa} X_j), \law(\frac{1}{\sqrt{{\kappa}}}\sum_{j=1}^{\kappa} X_j')\big)\in \mathcal{O}(\alpha).
\end{align*}
\textbf{The $T_{28}$ Term:}
We have
\begin{align*}
T_{28} &= 8 \alpha \sum_{j=1}^{\kappa} \E[\|  ( (1-\alpha )^{j-1}-1)(w_{{\kappa}t+{\kappa}-j} - w_{{\kappa}t+{\kappa}-j}^\prime)\|_2^2]\\
&\leq \mathcal{O}(\alpha) \cdot \sum_{j=1}^{\kappa} ( (1-\alpha )^{j-1}-1)^2\leq \mathcal{O}(\alpha) \cdot \sum_{j=1}^{\kappa} ( (1-\alpha )^{j-1}-1) \in \mathcal{O}(\alpha^2 )\cdot {\kappa}^2.
\end{align*}

Combining these bounds and recalling ${\kappa} = \lfloor \alpha^{-\frac{1}{2}}\rfloor $, we obtain
$T_2 \in \mathcal{O}(\alpha),$
proving Lemma~\ref{lem:additive-limit-noise-T2}.

\section{Proof of Theorem \ref{thm: additive limit}}
\label{sec: additive limit}

In this section, we prove Theorem \ref{thm: additive limit}, which establishes steady-state convergence for general SA update \eqref{eq: iid general SA}. By Proposition \ref{prop:gaussian}, we have established Step 1 of the prelimit coupling strategy outlined in Section \ref{sec: technique}. In the following, we complete Steps 2 and 3 in Subsections \ref{sec: additive limit rational} and \ref{sec: additive limit continuity}, respectively.

\subsection{Step 2: Gaussian Noise and Rational Stepsize}\label{sec: additive limit rational}

As outlined in Section~\ref{sec: technique}, 
we start with equation \eqref{eq: iid approximate} to  obtain the following dynamic for $Z_t$:
\begin{equation}\label{eq: additive centralized dynamic}
Z_{t+1} = (1-\alpha)Z_t + \sqrt{\alpha}\left(\mT(\sqrt{\alpha} Z_t + \theta^*) - \mT(\theta^*) + w_t'\right),
\end{equation}
where $\{w_t^\prime\}_{t \geq 0}$ are i.i.d zero-mean noise with normal distribution, and sharing the same variance as $\{w_t(\theta^*)\}_{t \geq 0}$. Note that dynamic \eqref{eq: iid approximate} is a special case of dynamic \eqref{eq: iid general SA}, hence the moment bounds and convergence to the stationary distribution for \eqref{eq: iid general SA} also hold for \eqref{eq: iid approximate}.

We consider the coupling given by equation \eqref{eq:step2couple} with $\alpha' = \alpha/k$. Direct calculation gives
\begin{align}
Z_{kt+k}^\prime 
=& (1-\frac{\alpha}{k})^k Z_{kt}^\prime + \sqrt{\frac{\alpha}{k}}\sum_{j = 0}^{k-1} \big( (1-\frac{\alpha}{k} )^j - 1 \big) \big(\mT (\sqrt{\frac{\alpha}{k}} Z_{kt+k-1-j}^\prime + \theta^* ) - \mT(\theta^*) + w_{kt+k-1-j}^\prime \big) \notag \\
&+ \sqrt{\frac{\alpha}{k}}\sum_{j = 0}^{k-1} \big(\mT (\sqrt{\frac{\alpha}{k}} Z_{kt+k-1-j}^\prime + \theta^* ) - \mT (\sqrt{\frac{\alpha}{k}} Z_{kt}^\prime + \theta^* ) \big) \notag \\
&+ \sqrt{\alpha k} \big(\mT (\sqrt{\frac{\alpha}{k}} Z_{kt}^\prime + \theta^* ) - \mT(\theta^*)  \big) + \sqrt{\alpha}\cdot \frac{w_{kt}^\prime + \dots + w_{kt+k-1}^\prime}{\sqrt{k}}.
\label{eq: alpha kt+k}
\end{align}

Combining equations \eqref{eq:step2couple} and \eqref{eq: alpha kt+k}, we obtain
\begin{align*}
Z_{t+1} - Z_{kt+k}^\prime =& (1-\alpha)(Z_{t} - Z_{kt}^\prime) \\
& + \big(1-\alpha -  (1-\frac{\alpha}{k} )^k \big) Z_{kt}^\prime
  + \sqrt{\alpha}\left(\mT(\sqrt{\alpha}Z_t + \theta^*) - \mT(\sqrt{\alpha}Z_{kt}^\prime + \theta^*)\right) \\
&+ \sqrt{\alpha}\Big( \big(\mT (\sqrt{\alpha} Z_{kt}^\prime + \theta^* ) - \mT(\theta^*) \big) - \sqrt{k}\big(\mT (\sqrt{\frac{\alpha}{k}} Z_{kt}^\prime + \theta^* ) - \mT(\theta^*) \big)\Big)\\
&+\sqrt{\frac{\alpha}{k}}\sum_{j = 0}^{k-1}\big(1-(1-\frac{\alpha}{k})^j \big)\big(\mT (\sqrt{\frac{\alpha}{k}} Z_{kt+k-1-j}^\prime + \theta^* ) - \mT(\theta^*) + w_{kt+k-1-j}^\prime\big)\\
&+\sqrt{\frac{\alpha}{k}}\sum_{j = 0}^{k-1}\big(\mT (\sqrt{\frac{\alpha}{k}} Z_{kt}^\prime + \theta^* ) - \mT (\sqrt{\frac{\alpha}{k}} Z_{kt+k-1-j}^\prime + \theta^* ) \big) := (1-\alpha)(Z_{t} - Z_{kt}^\prime) + A,
\end{align*}
where $A$ collects all but the first term on the RHS.
Applying the  Moreau envelope $M_\eta(\cdot)$ defined in equation \eqref{eq: Moreau envelope} to both sides of above equation and by property (1) in Proposition \ref{prop: Moreau envelope}, we obtain
\begin{equation}\label{eq: different alpha}
M_\eta(Z_{t+1} - Z_{kt+k}^\prime ) \leq (1-\alpha)^2M_\eta(Z_{t} - Z_{kt}^\prime )  + (1-\alpha)\underbrace{\langle \nabla M_\eta(Z_{t} - Z_{kt}^\prime), A \rangle}_{T_1} + \underbrace{\frac{1}{2\eta}\|A\|_2^2}_{T_2}.
\end{equation}
The following lemmas, proved in Sections~\ref{sec: additive limit rational-T1} and~\ref{sec: additive limit rational-T2}, bound the $T_1$ and $T_2$ terms above.
\begin{lemma}
    \label{lem:additive-limit-rational-T1}
    Under the setting of Theorem~\ref{thm: additive limit},  there exist $t_\alpha'''>0$ such that 
    $
    \E[T_1] \leq \frac{2\alpha\gamma u_{cm}}{l_{cm}}\E[M_\eta(Z_t - Z_{kt}^\prime)] + o(\alpha)
    $ for sufficiently large $t \geq t_\alpha'''$.
\end{lemma}
\begin{lemma}
    \label{lem:additive-limit-rational-T2}
    Under the setting of Theorem~\ref{thm: additive limit}, there exist $t_\alpha''''>0$ such that 
    $
    \E[T_2] \leq \frac{5\alpha^2u_{cm}^2\gamma^2}{\eta l_{cs}^2}\E[M_\eta(Z_t - Z_{kt}^\prime)] + \mathcal{O}(\alpha^2)
    $ for sufficiently large $t \geq t_\alpha''''$.
\end{lemma}

Plugging the above bounds for $T_1$ and $T_2$ into equation \eqref{eq: different alpha}, we obtain
\begin{equation*}
\E[M_\eta(Z_{t+1} - Z_{kt+k}^\prime )] \leq \Big(1 - 2\alpha \big(1-\frac{(1-\alpha)\gamma u_{cm}}{l_{cm}} \big) + \mathcal{O}(\alpha^2)\Big)\E[M_\eta(Z_{t} - Z_{kt}^\prime )]  + o(\alpha).
\end{equation*}

By the similar argument as in the proof of Lemma \ref{lemma: additive 2n moment}, we can always choose proper $\eta, \bar{\alpha}$ such that for $ \forall \alpha \leq \bar{\alpha}$, there exist $t_\alpha$ such that for all $t \geq t_\alpha$, we obtain
\begin{equation*}
\E[M_\eta(Z_{t+1} - Z_{kt+k}^\prime )] \leq \left(1 - \alpha(1-\sqrt{\gamma})\right)\E[M_\eta(Z_{t} - Z_{kt}^\prime )]  + o(\alpha)
\end{equation*}
which implies 
$
\lim_{t \to \infty} \E[M_\eta(Z_{t} - Z_{kt}^\prime)] \in o(1).
$
By triangle inequality, we have
\begin{align*}
W_2\big( &\mathcal{L}(Z^{(\alpha
)}), \mathcal{L}(Z^{(\alpha
/k)})\big)\leq  \lim_{t \to \infty} \Big\{{W_2\big(\mathcal{L}(Z^{(\alpha
)}), \mathcal{L}(Z_t)\big)} +W_2\big(\mathcal{L}(Z_t), \mathcal{L}(Z_{kt}^\prime)\big) +{W_2\big(\mathcal{L}(Z_{kt}^\prime), \mathcal{L}(Z^{(\alpha/k)})\big)} \Big\}\\
&\overset{\text{(i)}}{=}\lim_{t \to \infty} W_2\big(\mathcal{L}(Z_t), \mathcal{L}(Z_{kt}^\prime)\big) \overset{\text{(ii)}}{\leq} \lim_{t \to \infty} \sqrt{\mathbb{E}[\|Z_{t} - Z_{kt}^\prime\|_c^2]}\overset{\text{(iii)}}{\leq} \lim_{t \to \infty} \sqrt{2u_{cm}^2\mathbb{E}[M(Z_{t} - Z_{kt}^\prime)]}\in o(1),
\end{align*}
where (i) follows from Theorem \ref{thm: additive convergence}, (ii) holds by the definition, and (iii) holds by Proposition~\ref{prop: Moreau envelope}. Therefore, we have that for all $k \in \mathbb{N}^+$ and $\alpha > 0$,
$
W_2\big(\mathcal{L}(Z^{(\alpha
)}), \mathcal{L}(Z^{(\alpha
/k)})\big) \in o(1).
$

When $k \in \mathbb{Q}^+, k > 1$ and $\alpha>0$, let $k = \frac{p}{q}$. We have
\begin{align*}
W_2\big(\mathcal{L}(Z^{(\alpha
)}), \mathcal{L}(Z^{(\alpha
/k)})\big) &\leq W_2\big(\mathcal{L}(Z^{(\alpha
)}), \mathcal{L}(Z^{(\alpha
/p)})\big) + W_2\big(\mathcal{L}(Z^{(\alpha/p
)}), \mathcal{L}(Z^{(\alpha
/k)})\big)\overset{(i)}{\leq} o(1) + o(1) \in o(1),
\end{align*}
where (i) holds because $\frac{\alpha}{p} = \frac{\frac{\alpha}{k}}{q}$ and $\frac{\alpha}{k} \leq \alpha$.
Then, by following the discussion in Section \ref{sec: technique}, we completes Step 2 of the proof of Theorem~\ref{thm: additive limit}.

\subsubsection{Proof of Lemma~\ref{lem:additive-limit-rational-T1} on $T_1$}
\label{sec: additive limit rational-T1}
By part (4) in Proposition \ref{prop: Moreau envelope} and $\{w_{kt+k-1-j}^\prime\}_{j=0}^{k-1}$ being i.i.d.\ zero mean, $\E[T_1]$ is bounded by 
\begin{align}
& \E\bigg[\|Z_{t} - Z _{kt}^\prime \|_m\|(1-\alpha - (1-\frac{\alpha}{k})^k)Z_{kt}^\prime\|_m\bigg] \tag{$T_{11}$}\\
&+ \E[\|Z_{t} - Z_{kt}^\prime \|_m\|\sqrt{\alpha}\left(\mT(\sqrt{\alpha}Z_t + \theta^*) - \mT(\sqrt{\alpha}Z_{kt}^\prime + \theta^*)\right)\|_m] \tag{$T_{12}$}\\
&+ \E\bigg[\|Z_{t} - Z_{kt}^\prime \|_m\|\sqrt{\alpha}\Big( \left(\mT(\sqrt{\alpha} Z_{kt}^\prime + \theta^*) - \mT(\theta^*) \right) - \sqrt{k}\big(\mT(\sqrt{\frac{\alpha}{k}} Z_{kt}^\prime + \theta^*) - \mT(\theta^*) \big)\Big)\|_m\bigg] \tag{$T_{13}$}\\
&+ \E\bigg[\|Z_{t} - Z_{kt}^\prime \|_m\|\sqrt{\frac{\alpha}{k}}\sum_{j = 0}^{k-1}(1-(1-\frac{\alpha}{k})^j )\big(\mT(\sqrt{\frac{\alpha}{k}} Z_{kt+k-1-j}^\prime + \theta^*) - \mT(\theta^*)\big)\|_m \bigg] \tag{$T_{14}$} \\
&+ \E\bigg[\|Z_{t} - Z_{kt}^\prime \|_m\|\sqrt{\frac{\alpha}{k}}\sum_{j = 0}^{k-1}\big(\mT(\sqrt{\frac{\alpha}{k}} Z_{kt}^\prime + \theta^*) - \mT(\sqrt{\frac{\alpha}{k}} Z_{kt+k-1-j}^\prime + \theta^*) \big)\|_m \bigg]. \tag{$T_{15}$}
\end{align}

\textbf{The $T_{11}$ Term:}
We begin with
    $
    T_{11} = |1-\alpha - (1-\frac{\alpha}{k})^k|\E[\|Z_{t} - Z_{kt}^\prime \|_m\|Z_{kt}^\prime\|_m].
    $
Note that $f(x) = (1-\frac{\alpha}{x})^x$ increases monotonically when $x \geq \alpha$. Therefore, when $\alpha \leq 1$, we obtain
\begin{equation}\label{eq: alpha order}
\begin{aligned}
|1-\alpha - (1-\frac{\alpha}{k})^k| &\leq \lim_{k \to \infty}(1-\frac{\alpha}{k})^k - 1+\alpha= \exp(-\alpha) -  1+\alpha \in \mathcal{O}(\alpha^2).
\end{aligned}
\end{equation}
By Cauchy–Schwarz inequality, we obtain
\begin{equation}\label{eq: Z}
\begin{aligned}
\E[\|Z_{t} - Z_{kt}^\prime \|_m\|Z_{kt}^\prime\|_m] &\leq \E[\|Z_{t} \|_m\|Z_{kt}^\prime\|_m]+\E[\|Z_{kt}^\prime \|_m\|Z_{kt}^\prime\|_m]\\
&\leq \sqrt{\E[\|Z_{t} \|_m^2]\E[\|Z_{kt}^\prime\|_m^2]} + \E[\|Z_{kt}^\prime\|_m^2] \overset{(\text{i})}{\in} \mathcal{O}(1),
\end{aligned}  
\end{equation}
where (i) holds by Proposition \ref{thm: additive 2n moment}. Therefore, we conclude that $T_{11} \in \mathcal{O}(\alpha^2)$.

\textbf{The $T_{12}$ Term:}
Turning to $T_{12}$, we have
\begin{align*}
T_{12} 
&\leq \frac{\sqrt{\alpha}}{l_{cm}}\E[\|Z_{t} - Z_{kt}^\prime \|_m\|\mT(\sqrt{\alpha}Z_t + \theta^*) - \mT(\sqrt{\alpha}Z_{kt}^\prime + \theta^*)\|_c]
\leq \frac{2\alpha\gamma u_{cm}}{l_{cm}}\E[M_\eta(Z_t - Z_{kt}^\prime)].
\end{align*}

\textbf{The $T_{13}$ Term:}
For $T_{13}$, by Cauchy–Schwarz inequality, we obtain
\begin{align*}
T_{13} 
&\leq \sqrt{\alpha}\sqrt{\E[\|Z_{t} - Z_{kt}^\prime \|_m^2] \cdot \E\Big[\Big\| \mT(\sqrt{\alpha} Z_{kt}^\prime + \theta^*) - \mT(\theta^*)  - \sqrt{k}\Big(\mT\Big(\sqrt{\frac{\alpha}{k}} Z_{kt}^\prime + \theta^*\Big) - \mT(\theta^*) \Big)\Big\|_m^2\Big]}.
\end{align*}
Note that 
$\E[\|Z_{t} - Z_{kt}^\prime \|_m^2] \leq 2\E[\|Z_{t} \|_m^2]+2\E[\| Z_{kt}^\prime \|_m^2] \in \mathcal{O}(1).$
For the second expectation term on the above RHS, we have
\begin{align}
&\E\Big[\Big\| \mT(\sqrt{\alpha} Z_{kt}^\prime + \theta^*) - \mT(\theta^*)  - \sqrt{k}\big(\mT (\sqrt{\frac{\alpha}{k}} Z_{kt}^\prime + \theta^* ) - \mT(\theta^*) \big)\Big\|_m^2\Big] \label{eq:T13exp} \\
=&  \E\bigg[\Big\| \mT(\sqrt{\alpha} Z_{kt}^\prime + \theta^*) - \mT(\theta^*)  - \sqrt{k}\big(\mT(\sqrt{\frac{\alpha}{k}} Z_{kt}^\prime + \theta^*) - \mT(\theta^*) \big)\Big\|_m^2\mathbbm{1}(\alpha^{\frac{1}{4}}Z_{kt}^\prime \in B^d(0, 1), Z_{kt}^\prime \neq 0)\bigg] \label{eq:T13exp_in_ball}\\
&+\E\bigg[\Big\| \mT(\sqrt{\alpha} Z_{kt}^\prime + \theta^*) - \mT(\theta^*)  - \sqrt{k}\big(\mT(\sqrt{\frac{\alpha}{k}} Z_{kt}^\prime + \theta^*) - \mT(\theta^*) \big)\Big\|_m^2\mathbbm{1}(\alpha^{\frac{1}{4}}Z_{kt}^\prime \notin B^d(0, 1))\bigg]. \label{eq:T13exp_nin_ball}
\end{align}
By Assumption \ref{as: contraction}, the term \eqref{eq:T13exp_in_ball} can be bounded as follows:
\begin{align}
&\eqref{eq:T13exp_in_ball}\nonumber\\
\leq& \E\bigg[\Big\| \mT(\sqrt{\alpha} Z_{kt}^\prime + \theta^*) - \mT(\theta^*)  - \sqrt{\alpha}\|Z_{kt}^\prime\|G(Z_{kt}^\prime/\|Z_{kt}^\prime\|)\Big\|_m^2\mathbbm{1}(\alpha^{\frac{1}{4}}Z_{kt}^\prime \in B^d(0, 1), Z_{kt}^\prime \neq 0)\bigg]\nonumber\\
&+\E\bigg[\Big\| \sqrt{k}\big(\mT(\sqrt{\frac{\alpha}{k}} Z_{kt}^\prime + \theta^*) - \mT(\theta^*) \big)-\sqrt{\alpha}\|Z_{kt}^\prime\|G(Z_{kt}^\prime/\|Z_{kt}^\prime\|)\Big\|_m^2\mathbbm{1}(\alpha^{\frac{1}{4}}Z_{kt}^\prime \in B^d(0, 1), Z_{kt}^\prime \neq 0)\bigg]\nonumber\\
=& \alpha\E\bigg[\Big\|\frac{\mT(\sqrt{\alpha}\|Z_{kt}^\prime\| \cdot Z_{kt}^\prime/\|Z_{kt}^\prime\|+ \theta^*) - \mT(\theta^*)}{\sqrt{\alpha}\|Z_{kt}^\prime\|}   - G(Z_{kt}^\prime/\|Z_{kt}^\prime\|)\Big\|_m^2\|Z_{kt}^\prime\|^2\mathbbm{1}(\alpha^{\frac{1}{4}}Z_{kt}^\prime \in B^d(0, 1), Z_{kt}^\prime \neq 0)\bigg]\label{eq:apple}\\
&+\alpha\E\bigg[\Big\|\frac{\big(\mT(\sqrt{\frac{\alpha}{k}}\|Z_{kt}^\prime\| \cdot Z_{kt}^\prime/\|Z_{kt}^\prime\|  + \theta^*) - \mT(\theta^*) \big)}{\sqrt{\frac{\alpha}{k}}\|Z_{kt}^\prime\|}-G(Z_{kt}^\prime/\|Z_{kt}^\prime\|)\Big\|_m^2\|Z_{kt}^\prime\|^2\mathbbm{1}(\alpha^{\frac{1}{4}}Z_{kt}^\prime \in B^d(0, 1), Z_{kt}^\prime \neq 0)\bigg].\label{eq:banana}
\end{align}
By Proposition~\ref{thm: additive 2n moment}, there exist constants
$\bar\alpha_1>0$ and $M>0$ such that, for every $\alpha\le \bar\alpha_1$, there
exists $t_\alpha>0$ satisfying $\E[\|Z_{kt}'\|^2] \leq M$ for any $t \geq t_\alpha$. Consider a fixed $\epsilon>0$. By the uniform convergence in Lemma~\ref{usefullemma1},
there exists $\delta>0$ such that, for all $\alpha\le \min\{\bar\alpha_1,\delta\}$
and all $t\ge t_\alpha$, the following bound holds almost surely:
\begin{align*}
\Big\|\frac{\big(\mT(\sqrt{\alpha}\|Z_{kt}^\prime\| \cdot Z_{kt}^\prime/\|Z_{kt}^\prime\|  + \theta^*) - \mT(\theta^*) \big)}{\sqrt{\alpha}\|Z_{kt}^\prime\|}-G(Z_{kt}^\prime/\|Z_{kt}^\prime\|)\Big\|_m^2\mathbbm{1}(\alpha^{\frac{1}{4}}Z_{kt}^\prime \in B^d(0, 1), Z_{kt}^\prime \neq 0) \leq \epsilon/M.
\end{align*}
Multiplying both sides by $\|Z_{kt}^\prime\|^2$ and taking expectations, we
obtain that for all $\alpha\le \min\{\bar\alpha_1,\delta\}$ and $t\ge t_\alpha$,
\begin{align*}
&\E\bigg[\Big\|\frac{\big(\mT(\sqrt{\alpha}\|Z_{kt}^\prime\| \cdot Z_{kt}^\prime/\|Z_{kt}^\prime\|  + \theta^*) - \mT(\theta^*) \big)}{\sqrt{\alpha}\|Z_{kt}^\prime\|}-G(Z_{kt}^\prime/\|Z_{kt}^\prime\|)\Big\|_m^2\|Z_{kt}^\prime\|^2\mathbbm{1}(\alpha^{\frac{1}{4}}Z_{kt}^\prime \in B^d(0, 1), Z_{kt}^\prime \neq 0)\bigg]\leq \epsilon,
\end{align*}Consequently, $\eqref{eq:apple}\in o(\alpha)$. By the same argument,
$\eqref{eq:banana}\in o(\alpha)$. Combining the two bounds yields
$\eqref{eq:T13exp_in_ball}\in o(\alpha)$.

For the term in \eqref{eq:T13exp_nin_ball}, by Cauchy–Schwarz inequality and Markov inequality, we obtain
\begin{align*}
\eqref{eq:T13exp_nin_ball}
&\leq \frac{2\gamma\alpha u_{cm}^2}{l_{cm}^2}\E[\|Z_{kt}^\prime\|_m^2\mathbbm{1}(\alpha^{\frac{1}{4}}Z_{kt}^\prime \notin B^d(0, 1))]\\
&\leq \frac{2\gamma\alpha u_{cm}^2}{l_{cm}^2}\sqrt{\E[\|Z_{kt}^\prime\|_m^4]}\sqrt{\P(\|Z_{kt}^\prime\|_2^4 \geq \frac{1}{\alpha
})}\leq \frac{2\gamma\alpha u_{cm}^2}{l_{cm}^2}\sqrt{\E[\|Z_{kt}^\prime\|_m^4]} \sqrt{\alpha \E[\|Z_{kt}^\prime\|_2^4]}  \in \mathcal{O}(\alpha^\frac{3}{2}),
\end{align*}
where the last step follows from $\E[\|Z_{kt}^\prime\|_m^4]=O(1)$.
Combining all the analysis together, we obtain that
$\eqref{eq:T13exp}\in o(\alpha),$
which in turn implies $T_{13} \in o(\alpha)$.

\textbf{The $T_{14}$ Term:}
For $T_{14}$, we have
\begin{align*}
T_{14} 
&\leq \sum_{j = 0}^{k-1}\E\bigg[\|Z_{t} - Z_{kt}^\prime \|_m \bigg\|\sqrt{\frac{\alpha}{k}}\big(1-(1-\frac{\alpha}{k})^j \big)\Big(\mT\Big(\sqrt{\frac{\alpha}{k}} Z_{kt+k-1-j}^\prime + \theta^*\Big) - \mT(\theta^*)\big) \bigg\|_m\bigg]\\
&\leq \frac{\gamma }{l_{cm}}\sum_{j = 0}^{k-1}\frac{\alpha}{k}(1-(1-\frac{\alpha}{k})^j )\E[\|Z_{t} - Z_{kt}^\prime \|_m\|Z^\prime_{kt+k-1-j}\|_c ]\\
&\leq \frac{\gamma }{l_{cm}}\sum_{j = 0}^{k-1}\frac{\alpha}{k}(1-(1-\frac{\alpha}{k})^j ) \big(\sqrt{\E[\|Z_t\|^2_m]\E[\|Z_{kt+k-1-j}^\prime\|^2_c]}+\sqrt{\E[\|Z_{kt}^\prime\|^2_m]\E[\|Z_{kt+k-1-j}^\prime\|^2_c]}\big)\\
&\leq \frac{\gamma }{l_{cm}} \sum_{j = 0}^{k-1}\frac{\alpha}{k}\big(1-(1-\frac{\alpha}{k}\big)^j ) \cdot \mathcal{O}(1) =\mathcal{O} \big(\alpha - 1+\big(1-\frac{\alpha}{k}\big)^k\big) \overset{(\text{i})}{\in } \mathcal{O}(\alpha^2),
\end{align*}
where (i) holds by equation \eqref{eq: alpha order}.

\textbf{The $T_{15}$ Term:}
Finally, we turn to $T_{15}$:
\begin{align*}
T_{15} &= \E[\|Z_{t} - Z_{kt}^\prime \|_m\|\sqrt{\frac{\alpha}{k}}\sum_{j = 0}^{k-1}\big(\mT(\sqrt{\frac{\alpha}{k}} Z_{kt}^\prime + \theta^*) - \mT(\sqrt{\frac{\alpha}{k}} Z_{kt+k-1-j}^\prime + \theta^*) \big)\|_m]\\
&\leq \frac{\alpha\gamma}{k l_{cm}}\sum_{j = 0}^{k-1}\E[\|Z_{t} - Z_{kt}^\prime \|_m\|Z_{kt+k-1-j}^\prime - Z_{kt}^\prime \|_c].
\end{align*}
By equation \eqref{eq:step2couple}, we obtain
\begin{align*}
&\|Z_{kt+k-1-j}^\prime - Z_{kt}^\prime \|_c\\
=& \|((1-\frac{\alpha}{k})^{k-1-j} - 1)Z_{kt}^\prime + \sqrt{\frac{\alpha}{k}}\sum_{l=1}^{k-1-j}(1-\frac{\alpha}{k})^{l-1}\big(\mT(\sqrt{\frac{\alpha}{k}}Z_{kt+k-1-j-l}^\prime + \theta^*) - \mT(\theta^*) + w_{kt+k-1-j-l}^\prime \big)\|_c.
\end{align*}
Therefore, we obtain that $T_{15}$ is upper bounded by
\begin{align}
& {\frac{\alpha\gamma}{k l_{cm}}\sum_{j = 0}^{k-1}\E[\|Z_{t} - Z_{kt}^\prime \|_m\|((1-\frac{\alpha}{k})^{k-1-j} - 1)Z_{kt}^\prime \|_c]} \tag{$T_{151}$}\\
&+ {\frac{\alpha\gamma}{k l_{cm}}\sum_{j = 0}^{k-1}\E[\|Z_{t} - Z_{kt}^\prime \|_m\|\sqrt{\frac{\alpha}{k}}\sum_{l=1}^{k-1-j}(1-\frac{\alpha}{k})^{l-1}\big(\mT(\sqrt{\frac{\alpha}{k}}Z_{kt+k-1-j-l}^\prime + \theta^*) - \mT(\theta^*) \big)\|_c]} \tag{$T_{152}$}\\
&+ {\frac{\alpha\gamma}{k l_{cm}}\sum_{j = 0}^{k-1}\E[\|Z_{t} - Z_{kt}^\prime \|_m\|\sqrt{\frac{\alpha}{k}}\sum_{l=1}^{k-1-j}(1-\frac{\alpha}{k})^{l-1} w_{kt+k-1-j-l}^\prime\|_c]} \tag{$T_{153}$}.
\end{align}

We analyze three terms $T_{151},T_{152},T_{153}$ separately. Note that
\begin{align*}
T_{151}&\leq  \frac{\alpha\gamma u_{cm}}{k l_{cm}} \E[\|Z_{t} - Z_{kt}^\prime \|_m\|Z_{kt}^\prime \|_m]\sum_{j = 0}^{k-1}(1-(1-\frac{\alpha}{k})^{k-1-j} ) 
 \leq \mathcal{O}(1) \cdot \frac{\alpha}{k}\sum_{j = 0}^{k-1}((1-\frac{\alpha}{k})^{k-1-j} - 1) \overset{(i)}{\in} \mathcal{O}(\alpha^2),
\end{align*}
where (i) holds by equation \eqref{eq: alpha order}. For $T_{152}$, we have
\begin{align*}
T_{152} &\leq \frac{\alpha\gamma}{k l_{cm}}\sum_{j = 0}^{k-1}\sum_{l=1}^{k-1-j}\E[\|Z_{t} - Z_{kt}^\prime \|_m\|\sqrt{\frac{\alpha}{k}}(1-\frac{\alpha}{k})^{l-1}\big(\mT(\sqrt{\frac{\alpha}{k}}Z_{kt+k-1-j-l}^\prime + \theta^*) - \mT(\theta^*) \big)\|_c]\\
&\leq \frac{\alpha\gamma}{k l_{cm}}\sum_{j = 0}^{k-1}\sum_{l=1}^{k-1-j}\frac{\alpha}{k}(1-\frac{\alpha}{k})^{l-1}\E[\|Z_{t} - Z_{kt}^\prime \|_m\|Z_{kt+k-1-j-l}^\prime\|_c]\\
&\leq \frac{\alpha\gamma}{k l_{cm}}\sum_{j = 0}^{k-1}\sum_{l=1}^{k-1-j}\frac{\alpha}{k}(1-\frac{\alpha}{k})^{l-1}\Big(\sqrt{\E[\|Z_t\|_m^2]\E[\|Z_{kt+k-1-j-l}^\prime\|_c^2]}+\sqrt{\E[\|Z_{kt}^\prime\|_m^2]\E[\|Z_{kt+k-1-j-l}^\prime\|_c^2]}\Big)\\
&\leq \mathcal{O}(1) \cdot \frac{\alpha^2}{k^2}\sum_{j = 0}^{k-1}\sum_{l=1}^{k-1-j}(1-\frac{\alpha}{k})^{l-1}\leq \mathcal{O}(1) \cdot \frac{\alpha^2}{k^2} \cdot k^2 \in \mathcal{O}(\alpha^2).
\end{align*}
Lastly, we have
\begin{align*}
T_{153} &\overset{(\text{i})}{=} \frac{\alpha\gamma}{k l_{cm}}\sum_{j = 0}^{k-1}{\E[\|Z_{t} - Z_{kt}^\prime \|_m]} \cdot \E\Big[\Big\|\sqrt{\frac{\alpha}{k}}\sum_{l=1}^{k-1-j}\Big(1-\frac{\alpha}{k}\Big)^{l-1} w_{kt+k-1-j-l}^\prime\Big\|_c\Big]\\
&\overset{(\text{ii})}{\leq}  \mathcal{O}(1) \cdot \frac{\alpha^{\frac{3}{2}}}{k^{\frac{3}{2}}} \sum_{j = 0}^{k-1}\E[\|\sum_{l=1}^{k-1-j}(1-\frac{\alpha}{k})^{l-1} w_{kt+k-1-j-l}^\prime\|_2]\\
&\leq \mathcal{O}(1) \cdot \frac{\alpha^{\frac{3}{2}}}{k^{\frac{3}{2}}} \sum_{j = 0}^{k-1}\sqrt{\E[\|\sum_{l=1}^{k-1-j}(1-\frac{\alpha}{k})^{l-1} w_{kt+k-1-j-l}^\prime\|_2^2]}\\
&\leq  \mathcal{O}(1) \cdot \frac{\alpha^{\frac{3}{2}}}{k^{\frac{3}{2}}} \sum_{j = 0}^{k-1}\sqrt{\sum_{l=1}^{k-1-j}(1-\frac{\alpha}{k})^{2l-2}}\overset{\text{(iii)}}{\le} \mathcal{O}(1) \cdot \frac{\alpha^{\frac{3}{2}}}{k^{\frac{3}{2}}} {\sum_{j = 0}^{k-1}\sqrt{k-1-j}} 
\in \mathcal{O}(\alpha^{\frac{3}{2}}),
\end{align*}
where (i) holds because $Z_t$, $Z_{kt}^\prime$ are independent with $w_{kt+k-1-j-l}^\prime$ for $0\le j \le k-1$ and $1 \le l \le  k-1-j$, (ii) follows from $\E[\|Z_{t} - Z_{kt}^\prime \|_m]\in \mathcal{O}(1)$, and (iii) holds because ${\sum_{j = 0}^{k-1}\sqrt{k-1-j}} \in \mathcal{O}(k^{\frac{3}{2}})$. Putting the bounds of $T_{151}, T_{152}$ and $T_{153}$ together, we obtain $T_{15} \in \mathcal{O}(\alpha^\frac{3}{2})$. 

Finally, combining our bounds for $T_{11}\sim T_{15}$, we have 
$
\E[T_1] \leq \frac{2\alpha\gamma u_{cm}}{l_{cm}}\E[M_\eta(Z_t - Z_{kt}^\prime)] + o(\alpha),
$
thereby completing the proof of Lemma~\ref{lem:additive-limit-rational-T1}.

\subsubsection{Proof of Lemma~\ref{lem:additive-limit-rational-T2} on $T_2$}
\label{sec: additive limit rational-T2}
By Cauchy–Schwarz inequality, we obtain
\begin{align}
\E[T_2] \leq& {\frac{5}{2\eta} \E[\|(1-\alpha - (1-\frac{\alpha}{k})^k)Z_{kt}^\prime\|_2^2]} \tag{$T_{21}$}\\
&+  {\frac{5}{2\eta} \E[\|\sqrt{\alpha}\big(\mT(\sqrt{\alpha}Z_t + \theta^*) - \mT(\sqrt{\alpha}Z_{kt}^\prime + \theta^*)\big) \|_2^2]} \tag{$T_{22}$}\\
&+  {\frac{5}{2\eta}\E[\|\sqrt{\alpha}\Big( \mT(\sqrt{\alpha} Z_{kt}^\prime + \theta^*) - \mT(\theta^*)  - \sqrt{k}\big(\mT(\sqrt{\frac{\alpha}{k}} Z_{kt}^\prime + \theta^*) - \mT(\theta^*) \big)\Big)\|_2^2]} \tag{$T_{23}$}\\
&+ {\frac{5}{2\eta}\E[\|\sqrt{\frac{\alpha}{k}}\sum_{j = 0}^{k-1}(1-(1-\frac{\alpha}{k})^j )\big(\mT(\sqrt{\frac{\alpha}{k}} Z_{kt+k-1-j}^\prime + \theta^*) - \mT(\theta^*) + w_{kt+k-1-j}^\prime\big)\|_2^2]} \tag{$T_{24}$}\\
&+ {\frac{5}{2\eta} \E[\|\sqrt{\frac{\alpha}{k}}\sum_{j = 0}^{k-1}\big(\mT(\sqrt{\frac{\alpha}{k}} Z_{kt}^\prime + \theta^*) - \mT(\sqrt{\frac{\alpha}{k}} Z_{kt+k-1-j}^\prime + \theta^*) \big)\|_2^2]} \tag{$T_{25}$}.
\end{align}

\textbf{The $T_{21}$ Term:}
    We have the following bound, with (i) following from equation \eqref{eq: alpha order}:
    \begin{align*}
    T_{21} &\leq \frac{5}{2\eta} \big(1-\alpha - (1-\frac{\alpha}{k})^k\big)^2 \cdot \E[\|Z_{kt}^\prime\|_2^2]
    \leq \frac{5}{2\eta} \big(1-\alpha - (1-\frac{\alpha}{k})^k\big)^2 \cdot \mathcal{O}(1)
    \overset{(\text{i})}{\in} \mathcal{O}(\alpha^4).
    \end{align*}

\textbf{The $T_{22}$ Term:}
    For $T_{22}$, we have
    \begin{align*}
    T_{22} &\leq \frac{5\alpha}{2\eta l_{cs}^2} \E \left[\|\mT(\sqrt{\alpha}Z_t + \theta^*) - \mT(\sqrt{\alpha}Z_{kt}^\prime + \theta^*) \|_c^2 \right]
    \leq \frac{5\alpha^2u_{cm}^2\gamma^2}{\eta l_{cs}^2} \E[M_\eta(Z_t - Z_{kt}^\prime)].
    \end{align*}

\textbf{The $T_{23}$ Term:}   
    Using $\eqref{eq:T13exp}\in o(\alpha)$ and equivalence of norms in $\R^d$, we obtain  $T_{23} \in o(\alpha^2).$

\textbf{The $T_{24}$ Term:}
    By Cauchy–Schwarz inequality, we obtain
    \begin{align*}
    T_{24} \leq& \frac{5}{\eta}\E[\|\sqrt{\frac{\alpha}{k}}\sum_{j = 0}^{k-1}(1-(1-\frac{\alpha}{k})^j )\big(\mT(\sqrt{\frac{\alpha}{k}} Z_{kt+k-1-j}^\prime + \theta^*) - \mT(\theta^*) \big)\|_2^2]\\
    &+ \frac{5}{\eta}\E[\|\sqrt{\frac{\alpha}{k}}\sum_{j = 0}^{k-1}(1-(1-\frac{\alpha}{k})^j ) w_{kt+k-1-j}^\prime\|_2^2]\\
    \leq& \frac{5\gamma^2}{\eta l_{cs}^2} \frac{\alpha^2}{k^2} \E[\|\sum_{j = 0}^{k-1}(1-(1-\frac{\alpha}{k})^j )Z_{kt+k-1-j}^\prime \|_c^2]
    + \frac{5}{\eta}\frac{\alpha}{k}\sum_{j = 0}^{k-1}\E[\|(1-(1-\frac{\alpha}{k})^j ) w_{kt+k-1-j}^\prime\|_2^2]\\
    \leq& \frac{5\gamma^2}{\eta l_{cs}^2} \frac{\alpha^2}{k} {\sum_{j = 0}^{k-1}\E[\|(1-(1-\frac{\alpha}{k})^j )Z_{kt+k-1-j}^\prime \|_c^2]} 
    + \frac{5}{\eta}\frac{\alpha}{k}\sum_{j = 0}^{k-1}\E[\|(1-(1-\frac{\alpha}{k})^j ) w_{kt+k-1-j}^\prime\|_2^2]\\
    \overset{\text{(i)}}{\leq} & \mathcal{O}(\alpha^2) + \mathcal{O}(1) \cdot \frac{\alpha}{k}\sum_{j = 0}^{k-1}(1-(1-\frac{\alpha}{k})^j )^2\leq \mathcal{O}(\alpha^2) + \mathcal{O}(1) \cdot \frac{\alpha}{k}\sum_{j = 0}^{k-1}(1-(1-\frac{\alpha}{k})^j ) \overset{(\text{ii})}{\in}  \mathcal{O}(\alpha^2),
    \end{align*}
    where (i) holds because ${\sum_{j = 0}^{k-1}\E[\|(1-(1-\frac{\alpha}{k})^j )Z_{kt+k-1-j}^\prime \|_c^2]} {\in \mathcal{O}(k)},$ and (ii) holds by equation \eqref{eq: alpha order}. 

\textbf{The $T_{25}$ Term:}
    For $T_{25}$, we have
    \begin{align*}
    T_{25} &\leq \frac{5}{2\eta} \E[\|\sqrt{\frac{\alpha}{k}}\sum_{j = 0}^{k-1}\big(\mT(\sqrt{\frac{\alpha}{k}} Z_{kt}^\prime + \theta^*) - \mT(\sqrt{\frac{\alpha}{k}} Z_{kt+k-1-j}^\prime + \theta^*) \big)\|_2^2]\\
    &\leq \mathcal{O}(1) \cdot \frac{\alpha^2}{k} {\sum_{j = 0}^{k-1}\E[\|Z_{kt}^\prime - Z_{kt+k-1-j}^\prime\|_c^2]} \in \mathcal{O}(\alpha^2),
    \end{align*}
where the last step holds because ${\sum_{j = 0}^{k-1}\E[\|Z_{kt}^\prime - Z_{kt+k-1-j}^\prime\|_c^2]} {\in \mathcal{O}(k)}.$

Combining pieces, we obtain 
$
T_2 \leq \frac{5\alpha^2u_{cm}^2\gamma^2}{\eta l_{cs}^2}\E[M_\eta(Z_t - Z_{kt}^\prime)] + \mathcal{O}(\alpha^2),
$
proving Lemma~\ref{lem:additive-limit-rational-T2}.

\subsection{Step 3: General Stepsize}\label{sec: additive limit continuity}

In this subsection, we aim to prove that there exists an $\alpha_0$ such that $\mathcal{L}(Z^{(\alpha)})$ is continuous in $\alpha$ when $\alpha \in (0,\alpha_0)$  with respect to $W_2$. Let us consider two stepsizes $\alpha>0$ and $\alpha'>0$. For simplicity, we will let 
$\{Z_t\}_{t\geq0}$ and $\{Z^\prime_t\}_{t\geq0}$ denote the sequence associated with stepsize $\alpha$ and $\alpha'$, respectively. We consider the coupling given by equation \eqref{eq:step3couple} and 
it follows that
\begin{align*}
Z_{t+1} - Z_{t+1}^\prime &= (1-\alpha )(Z_t-Z_t^\prime) + \sqrt{\alpha}(\mT(\sqrt{\alpha^\prime} Z_t + \theta^*)- \mT(\sqrt{\alpha^\prime} Z_t^\prime + \theta^*)) \\
&+ \sqrt{\alpha}(\mT(\sqrt{\alpha} Z_t + \theta^*)-\mT(\sqrt{\alpha^\prime} Z_t + \theta^*))\\
&+ (\sqrt{\alpha}-\sqrt{\alpha^\prime})(\mT(\sqrt{\alpha^\prime} Z_t^\prime + \theta^*)-\mT(\theta^*)) + (\sqrt{\alpha} - \sqrt{\alpha^\prime})w_t^\prime\\
& := (1-\alpha )(Z_t-Z_t^\prime) + A,
\end{align*}
where $A$ collects all but the first RHS terms.
Applying the  Moreau envelope  defined in \eqref{eq: Moreau envelope} to both side of above equation and by property (1) in Proposition \ref{prop: Moreau envelope}, we obtain
\begin{align*}
\mathbb{E}[M_\eta(Z_{t+1} - Z_{t+1}^\prime)] \leq (1-\alpha )^2\mathbb{E}[M_\eta(Z_t-Z_t^\prime)] + (1-\alpha)\underbrace{\mathbb{E}\langle \nabla M(Z_t-Z_t^\prime), A\rangle}_{T_1} + \underbrace{(1/(2\eta))\mathbb{E}\|A\|_2^2}_{T_2}.
\end{align*}

\textbf{Bounding the $T_{1}$ Term:}
By property (4) in Proposition \ref{prop: Moreau envelope} and $w_{t}^\prime$ being i.i.d zero mean noise and independent with $Z_t$ and $Z_{t}^\prime$, we obtain
\begin{align}
T_1 \leq& {\sqrt{\alpha}\mathbb{E}[ \|Z_t-Z_t^\prime\|_m \|\mT(\sqrt{\alpha^\prime} Z_t + \theta^*)- \mT(\sqrt{\alpha^\prime} Z_t^\prime + \theta^*)\|_m]} \tag{$T_{11}$}\\
&+ {\sqrt{\alpha}\mathbb{E}[ \|Z_t-Z_t^\prime\|_m \|\mT(\sqrt{\alpha} Z_t + \theta^*)- \mT(\sqrt{\alpha^\prime} Z_t + \theta^*)\|_m]} \tag{$T_{12}$}\\
&+ {(\sqrt{\alpha}-\sqrt{\alpha^\prime})\mathbb{E}[ \|Z_t-Z_t^\prime\|_m \|\mT(\sqrt{\alpha^\prime} Z_t^\prime + \theta^*) - \mT(\theta^*)\|_m]} \tag{$T_{13}$}.
\end{align}
Let $\delta = |\alpha - \alpha^\prime| \leq \min((\frac{1}{\sqrt{\gamma}}-1)\alpha, \frac{1}{2}\alpha)$. Below, we bound $T_{11} \sim T_{13}$ separately, beginning with $T_{11}$:
\begin{align*}
T_{11} &\leq  \frac{\sqrt{\alpha}}{l_{cm}}\mathbb{E}[ \|Z_t-Z_t^\prime\|_m \|\mT(\sqrt{\alpha^\prime} Z_t + \theta^*)- \mT(\sqrt{\alpha^\prime} Z_t^\prime + \theta^*)\|_c]\\
&\overset{(i)}{\leq} \frac{2\sqrt{\alpha\alpha^\prime} u_{cm}\gamma}{l_{cm}}\mathbb{E}[ M_\eta(Z_t-Z_t^\prime)]\leq 2\sqrt{\alpha\alpha^\prime} \sqrt{\gamma}\mathbb{E}[ M_\eta(Z_t-Z_t^\prime)]\leq 2\alpha \gamma^{\frac{1}{4}}\mathbb{E}[ M_\eta(Z_t-Z_t^\prime)],
\end{align*}
where (i) holds because we can always choose a proper $\eta$ such that $\frac{u_{cm}}{l_{cm}} \leq \frac{1}{\sqrt{\gamma}}$. We next have
\begin{align*}
T_{12} &\leq \frac{\sqrt{\alpha}|\sqrt{\alpha}-\sqrt{\alpha^\prime}|u_{cm}\gamma}{l_{cm}}{\mathbb{E}[ \|Z_t-Z_t^\prime\|_m\|Z_t\|_m]} \\
&\in \mathcal{O}\Big(\sqrt{\alpha}|\sqrt{\alpha}-\sqrt{\alpha^\prime}|\Big)
\in \mathcal{O}\bigg(  \frac{\sqrt{\alpha}\delta}{\sqrt{\alpha}+\sqrt{\alpha^\prime}}\bigg)
\in \mathcal{O}\bigg(\frac{\sqrt{\alpha}\delta}{\min(\alpha, \alpha^\prime)^\frac{1}{2}}\bigg),
\end{align*}
where we use ${\mathbb{E}[ \|Z_t-Z_t^\prime\|_m\|Z_t\|_m]} {\in \mathcal{O}(1)}$. Similarly, we have
\begin{align*}
T_{13} &\leq \frac{\sqrt{\alpha^\prime}|\sqrt{\alpha}-\sqrt{\alpha^\prime}|u_{cm}\gamma}{l_{cm}} {\mathbb{E}[ \|Z_t-Z_t^\prime\|_m\|Z_t^\prime\|_m]} \\
&\in \mathcal{O}\big(\sqrt{\alpha^\prime}|\sqrt{\alpha}-\sqrt{\alpha^\prime}|\big)\in \mathcal{O}\Big(\frac{\sqrt{\alpha^\prime}\delta}{\min(\alpha, \alpha^\prime)^\frac{1}{2}}\Big)\in \mathcal{O}\Big(\frac{\sqrt{\alpha}\delta}{\min(\alpha, \alpha^\prime)^\frac{1}{2}}\Big),
\end{align*}
where the last inequality holds because $\delta \leq \min\big(\big(\frac{1}{\sqrt{\gamma}}-1\big)\alpha, \frac{1}{2}\alpha\big)$.

\textbf{Bounding the $T_{2}$ Term:}
We  have
\begin{align}
T_2 \leq&  {\frac{2\alpha}{ \eta}\mathbb{E}\|(\mT(\sqrt{\alpha^\prime} Z_t + \theta^*)- \mT(\sqrt{\alpha^\prime} Z_t^\prime + \theta^*))\|_2^2} \tag{$T_{21}$}\\
&+ {\frac{2\alpha}{ \eta}\mathbb{E}\|(\mT(\sqrt{\alpha} Z_t + \theta^*)-\mT(\sqrt{\alpha^\prime} Z_t + \theta^*))\|_2^2} \tag{$T_{22}$}\\
&+ {\frac{2(\sqrt{\alpha}-\sqrt{\alpha^\prime})^2}{ \eta}\mathbb{E}\|(\mT(\sqrt{\alpha^\prime} Z_t^\prime + \theta^*)-\mT(\theta^*))\|_2^2} \tag{$T_{23}$} \\
&+ {\frac{2(\sqrt{\alpha}-\sqrt{\alpha^\prime})^2}{ \eta}\mathbb{E}[\|w_t^\prime\|_2^2]}. \tag{$T_{24}$}
\end{align}
Below, we bound $T_{21} \sim T_{24}$ separately. We begin with
\begin{align*}
T_{21} &\leq \frac{2\alpha}{l_{cs}^2}\mathbb{E}\|(\mT(\sqrt{\alpha^\prime} Z_t + \theta^*)- \mT(\sqrt{\alpha^\prime} Z_t^\prime + \theta^*))\|_c^2
\leq \frac{4\alpha\alpha^\prime\gamma^2u_{cm}^2}{l_{cs}^2}\E[M_\eta(Z_t-Z_t^\prime)]
\leq \frac{6\alpha^2\gamma^2u_{cm}^2}{l_{cs}^2}\E[M_\eta(Z_t-Z_t^\prime)],
\end{align*}
where the last inequality holds because $\delta \leq \min((\frac{1}{\sqrt{\gamma}}-1)\alpha, \frac{1}{2}\alpha)$. The next three terms satisfy
\begin{align*}
T_{22} &\in \mathcal{O}\big(\alpha (\sqrt{\alpha}-\sqrt{\alpha^\prime})^2\big)\in \mathcal{O}\big(\frac{\alpha \delta^2}{(\sqrt{\alpha}+\sqrt{\alpha^\prime})^2}\big)\in \mathcal{O}\big(\frac{\alpha \delta^2}{\min(\alpha, \alpha^\prime)}\big).
\end{align*}
\begin{align*}
T_{23} &\in \mathcal{O}\big(\alpha^\prime (\sqrt{\alpha}-\sqrt{\alpha^\prime})^2\big)\in \mathcal{O}\big(\frac{\alpha^\prime \delta^2}{\min(\alpha, \alpha^\prime)}\big) \in \mathcal{O}\big(\frac{\alpha \delta^2}{\min(\alpha, \alpha^\prime)}\big). \quad T_{24} \in \mathcal{O}\big(\frac{\delta^2}{\min(\alpha, \alpha^\prime)}\big).
\end{align*}
Combining the bounds for $T_1$ and $T_2$ and noting that there exist an $\alpha_0$ such that $0<\big(1-2(1 - \gamma^{\frac{1}{4}})\alpha_0 + \mathcal{O}(\alpha_0^2)\big)<1$, we see that for any $\alpha \leq \alpha_0$, there exist $t_\alpha$ such that for any  $t \geq t_\alpha$, we obtain
\begin{align*}
\mathbb{E}[M_\eta(Z_{t+1} - Z_{t+1}^\prime)] \leq  \big(1-2(1 - \gamma^{\frac{1}{4}})\alpha + \mathcal{O}(\alpha^2)\big)\mathbb{E}[M_\eta(Z_{t } - Z_{t }^\prime)] + \mathcal{O}\big({\sqrt{\alpha}\delta}/{\min(\alpha, \alpha^\prime)^\frac{1}{2}}\big).
\end{align*}
Then, we obtain
$
\lim_{t \to \infty}\mathbb{E}[M_\eta(Z_{t } - Z_{t }^\prime)] \in \mathcal{O} \big( \frac{\delta}{ \min(\alpha,\alpha^\prime)} \big).
$ Hence, we have
\begin{align*}
W_2\big(\mathcal{L}(Z^{(\alpha
)}), \mathcal{L}(Z^{(\alpha^\prime)})\big) & \leq \lim_{t \to \infty} W_2\big(\mathcal{L}(Z^{(\alpha
)}), \mathcal{L}(Z_t)\big)+W_2\big(\mathcal{L}(Z_t), \mathcal{L}(Z_t^\prime)\big)+W_2\big(\mathcal{L}(Z_t), \mathcal{L}(Z^{(\alpha^\prime)})\big)\\
&\leq \lim_{t \to \infty} \sqrt{\mathbb{E}[\|Z_{t } - Z_{ t }^\prime\|_c^2]}
\leq \lim_{t \to \infty} \sqrt{2u_{cm}^2\mathbb{E}[M_\eta(Z_{t } - Z_{ t }^\prime)]}
\leq  \frac{c\sqrt{\delta}}{ \min(\alpha,\alpha^\prime)^\frac{1}{2}},
\end{align*}
where $c$ is a universal constant that is independent with $\alpha, \alpha^\prime$. Then, given $\alpha>0$, for $\forall \epsilon > 0$,  we can choose a sufficient small $\delta_\epsilon$ such that $\frac{c\sqrt{\delta_\epsilon}}{(\alpha-\delta_\epsilon)^\frac{1}{2}} \leq \epsilon$ and $0<\delta_\epsilon < \min\big(\big(\frac{1}{\sqrt{\gamma}}-1\big)\alpha, \frac{1}{2}\alpha\big)$.
When $\alpha^\prime$ is chosen to satisfy $|\alpha-\alpha^\prime| \leq \delta_\epsilon$, we obtain
$
W_2\big(\mathcal{L}(Z^{(\alpha
)}), \mathcal{L}(Z^{(\alpha^\prime)})\big) \leq \epsilon.
$
Therefore, we complete the proof of continuity of $\mathcal{L}(Z^{(\alpha)})$ with respect to $W_2$.
Recall that 
$
\lim_{\alpha \to 0, \alpha \in \mathbb{Q}^+} W_2\big(\mathcal{L}(Z^{(\alpha
)}), \mathcal{L}(Y)\big) = 0.
$
Thus, for $\forall \epsilon>0$, there exist $\delta>0$, such that for all rational $\alpha \leq \delta$, $W_2\big(\mathcal{L}(Z^{(\alpha
)}), \mathcal{L}(Y)\big) \leq \frac{\epsilon}{2}$.
Given arbitrary real number r such that $0< r\leq\frac{\delta}{2} $, there exist 
 $q(r) \in \mathbb{Q}$ such that $|r - q(r)| \leq \frac{\delta}{2}$ and $W_2\big(\mathcal{L}(Z^{(r
)}), \mathcal{L}(Z^{(q(r))})\big) \leq \frac{\epsilon}{2}$ by Section \ref{sec: additive limit continuity}. Then, 
$$W_2\big(\mathcal{L}(Z^{(r
)}), \mathcal{L}(Y)\big) \leq W_2\big(\mathcal{L}(Z^{(r
)}), \mathcal{L}(Z^{(q(r))})\big)+W_2\big(\mathcal{L}(Z^{(q(r)
)}), \mathcal{L}(Y)\big)
\overset{\text{(i)}}{\leq} \epsilon,$$
where step (i) holds by $q(r) \leq \frac{\delta}{2}+\frac{\delta}{2} = \delta$.
Hence, there exists a unique limit $\mathcal{L}(Y)$, depending only on $\mT$ and $\operatorname{Var}(\tmT(x_0, \theta^*))$, such that 
$
\lim_{\alpha \to 0}W_2\big(\mathcal{L}(Z^{(\alpha
)}), \mathcal{L}(Y)\big) = 0.
$
This completes Step 3.

Then, combining Proposition \ref{prop:gaussian} and the discussion in Section \ref{sec: technique}, there exists  a unique limit $\mathcal{L}(Y)$, depending only on $\mT$ and $\operatorname{Var}(\tmT(x_0, \theta^*))$, such that for the $Y^{(\alpha)}$ defined by general noise, we have
$
\lim_{\alpha \to 0}W_2\big(\mathcal{L}(Y^{(\alpha
)}), \mathcal{L}(Y)\big) = 0,
$
thereby completing the proof of Theorem \ref{thm: additive limit}.

\section{Proof of Theorem \ref{thm: additive smooth}}\label{sec: prf_additive smooth}
Based on the discussion in Section \ref{sec:prf_outline_thm_additive_smooth}, we provide detailed analyses for terms $T_1$ and $T_2$ in this section.

For term $T_1$, by Cauchy–Schwarz inequality, we obtain
\begin{align*}
\|T_1\|_c &\leq  \frac{1}{\sqrt{\alpha}}\E[\|\mT(\sqrt{\alpha} Z^{(\alpha)} + \theta^*) - \mT(\theta^*)\|_c\mathbbm{1}(\alpha^{\frac{1}{4}}Z^{(\alpha)} \notin B^d(0, 1))]\\
&\leq \frac{1}{\sqrt{\alpha}}\sqrt{\E[\|\mT(\sqrt{\alpha} Z^{(\alpha)} + \theta^*) - \mT(\theta^*)\|_c^2]}\sqrt{\mathbb{P}(\alpha^{\frac{1}{4}}Z^{(\alpha)} \notin B^d(0, 1))}\\
&\leq \gamma\sqrt{\E[\|Z^{(\alpha)}\|_c^2]}\sqrt{\mathbb{P}(\alpha^{\frac{1}{4}}Z^{(\alpha)} \notin B^d(0, 1))}\\
&\leq \gamma\sqrt{\E[\|Z^{(\alpha)}\|_c^2]}\sqrt{\sqrt{\alpha}\E\|Z^{(\alpha)}\|_2^2} 
\overset{(\text{i})}{\in} \mathcal{O}(\alpha^\frac{1}{4}),
\end{align*}
where (i) holds because the equivalence of all norms in $\R^d$, Fatou's lemma \cite[Exercise 3.2.4]{durrett2019probability} and Proposition \ref{thm: additive 2n moment}.
Therefore, we obtain
$\lim_{\alpha \to 0}T_1 = 0.$

For term $T_2$, we have
\begin{align*}
    &\lim_{\alpha \to 0}T_2\\ =& \lim_{\alpha \to 0}\frac{1}{\sqrt{\alpha}}\E[(\mT(\sqrt{\alpha} Z^{(\alpha)} + \theta^*) - \mT(\theta^*))\mathbbm{1}(\alpha^{\frac{1}{4}}Z^{(\alpha)} \in B^d(0, 1))]\\
    =& \lim_{\alpha \to 0} \E[(\frac{\mT(\sqrt{\alpha}\|Z^{(\alpha)}\|\cdot \frac{Z^{(\alpha)}}{\|Z^{(\alpha)}\|}  + \theta^*) - \mT(\theta^*)}{\sqrt{\alpha}\|Z^{(\alpha)}\|}-G(\frac{Z^{(\alpha)}}{\|Z^{(\alpha)}\|}))\|Z^{(\alpha)}\| \mathbbm{1}(\alpha^{\frac{1}{4}}Z^{(\alpha)} \in B^d(0, 1),Z^{(\alpha)} \neq 0)]\\
    &+\lim_{\alpha \to 0}\E[\|Z^{(\alpha)}\|g(Z^{(\alpha)}/\|Z^{(\alpha)}\|) \mathbbm{1}(\alpha^{\frac{1}{4}}Z^{(\alpha)} \in B^d(0, 1),Z^{(\alpha)} \neq 0)].
\end{align*}
For the first term on the right–hand side of the equality above, because $\E[\|Z^{(\alpha)}\|] \in \mathcal{O}(1)$, by dominated convergence theorem and Assumption \ref{as: contraction0}, we have
\begin{align*}
    \lim_{\alpha \to 0} \E[(\frac{\mT(\sqrt{\alpha}\|Z^{(\alpha)}\|\cdot \frac{Z^{(\alpha)}}{\|Z^{(\alpha)}\|}  + \theta^*) - \mT(\theta^*)}{\sqrt{\alpha}\|Z^{(\alpha)}\|}-G(\frac{Z^{(\alpha)}}{\|Z^{(\alpha)}\|}))\|Z^{(\alpha)}\| \mathbbm{1}(\alpha^{\frac{1}{4}}Z^{(\alpha)} \in B^d(0, 1),Z^{(\alpha)} \neq 0)]=0.
\end{align*}
By Lemma \ref{usefullemma1} and the fact that $\E[\|Z^{(\alpha)}\|^2] \in \mathcal{O}(1)$, we have
\begin{align*}
&\lim_{\alpha \to 0}\E[\|Z^{(\alpha)}\|G(Z^{(\alpha)}/\|Z^{(\alpha)}\|) \mathbbm{1}(\alpha^{\frac{1}{4}}Z^{(\alpha)} \notin B^d(0, 1))] \\
\leq & \lim_{\alpha \to 0}\sqrt{\E[\|Z^{(\alpha)}\|^2G^2(Z^{(\alpha)}/\|Z^{(\alpha)}\|)]\mathbb{P}(\alpha^{\frac{1}{4}}Z^{(\alpha)} \notin B^d(0, 1))}\\
\leq & \lim_{\alpha \to 0}\sqrt{\E[\|Z^{(\alpha)}\|^2\gamma^2]\alpha^{1/4}\E[\|Z^{(\alpha)}\|]}=0.
\end{align*}
Therefore, by definition \eqref{eq:h}, we have
\begin{align*}
\lim_{\alpha \to 0}T_2 = \lim_{\alpha \to 0}\E[H(Z^{(\alpha)})].
\end{align*}
Then, by Lemma \ref{usefullemma1}, we argue that $\{H(Z^{(\alpha)})\}$ is uniformly integrable because $\|H(\theta)\|_c \in \mathcal{O}(\|\theta\|_c)$ and $\E[\|Z^{(\alpha)}\|_c] \in \mathcal{O}(1).$ Besides that, because $G$ is bounded by Lemma \ref{usefullemma1}, $H$ is a continuous function. Therefore, by Vitali convergence theorem, we have
\begin{align*}
\E[Y] = \lim_{\alpha \to 0}T_2 = \lim_{\alpha \to 0}\E[H(Z^{(\alpha)})] = \E[H(Y)].
\end{align*}
\textbf{Case 1:} If $\mT$ is differentiable at $\theta^{*}$, the argument already appears in the proof outline for Theorem \ref{thm: additive smooth} (see Section \ref{sec:prf_outline_thm_additive_smooth}).  We therefore omit the details here.

\textbf{Case 2:}   By following the proof outline in Section \ref{sec:prf_outline_thm_additive_smooth} and subtracting $(1-\alpha)^2\E[(\zeta^{\top}Z^{(\alpha)})^2]$ from both sides of the equation \eqref{eq:prfoutline} and then dividing by $\alpha$, we obtain
\begin{align*}
\underbrace{(2-\alpha)\E[(\zeta^{\top}Z^{(\alpha)})^2]}_{T_1} =& \frac{2(1-\alpha)}{\sqrt{\alpha}}\E[\zeta^{\top}Z^{(\alpha)}\cdot \zeta^{\top}\big(\mT(\sqrt{\alpha} Z^{(\alpha)} + \theta^*) - \mT(\theta^*)\big)]\\
&+ \E[(\zeta^{\top}\big(\mT(\sqrt{\alpha} Z^{(\alpha)} + \theta^*) - \mT(\theta^*) + w\big))^2]\\
=& \underbrace{\frac{2(1-\alpha)}{\sqrt{\alpha}}\E[\zeta^{\top}Z^{(\alpha)}\cdot \zeta^{\top}\big(\mT(\sqrt{\alpha} Z^{(\alpha)} + \theta^*) - \mT(\theta^*)\big)]}_{T_2}
\\
&+ \underbrace{\E[(\zeta^{\top}\big(\mT(\sqrt{\alpha} Z^{(\alpha)} + \theta^*) - \mT(\theta^*)\big))^2]}_{T_3} + \underbrace{\E[(\zeta^{\top}w)^2]}_{T_4}.
\end{align*}
$\E[(\zeta^{\top}Y)^2]=0$ and Theorem \ref{thm: additive limit} imply
$\lim_{\alpha \to 0}T_1 = 0.$
By Cauchy–Schwarz, we bound $T_2$ and $T_3$ as 
\begin{align*}
\lim_{\alpha \to 0}|T_2| &\leq \lim_{\alpha \to 0}\frac{2|1-\alpha|}{\sqrt{\alpha}}\sqrt{\E[(\zeta^{\top}Z^{(\alpha)})^2]}\sqrt{\E[\big( \zeta^{\top}\big(\mT(\sqrt{\alpha} Z^{(\alpha)} + \theta^*) - \mT(\theta^*)\big)\big)^2]}\\
&\leq \lim_{\alpha \to 0}\underbrace{\sqrt{\E[(\zeta^{\top}Z^{(\alpha)})^2]}}_{\in o(1)}\underbrace{\sqrt{\E[\|\zeta\|_c^2 \| Z^{(\alpha)}\|_c^2]}}_{\in \mathcal{O}(1)}
=0
\end{align*}
and 
$
 \lim_{\alpha \to 0}T_3 \leq \lim_{\alpha \to 0} \alpha \E[\|\zeta\|_c^2 \| Z^{(\alpha)}\|_c^2]
=0.
$
Because $\var(w)$ is positive definite, we obtain
$
T_4 = \zeta^{\top} \var(w) \zeta >0.
$
However, we have $T_4 = 0$ by letting $\alpha \to 0$, which contradicts with the fact that $T_4 > 0$. Therefore, we have
$\E[{Y}] \neq 0.$

\section{Proof of Theorem \ref{thm: q limit}}\label{sec: q limit}
Recall that the term $T_{13}$ in the proof of Lemma \ref{lem:additive-limit-rational-T1} is of order $o(1)$ for general nonsmooth SA. In this section, we reanalyze the term $T_{13}$ for Q-learning algorithms to have the order of $\mathcal{O}(\alpha^{\frac{3}{2}})$, which not only establishes steady-state convergence but also provides an explicit convergence rate. 

By the proof of Lemma \ref{lem:additive-limit-rational-T1}, we have
\begin{align*}
T_{13} 
&\leq \sqrt{\alpha}\sqrt{\E[\|Z_{t} - Z_{kt}^\prime \|_m^2]\E[\|( \mT(\sqrt{\alpha} Z_{kt}^\prime + \theta^*) - \mT(\theta^*)  - \sqrt{k}(\mT(\sqrt{\alpha/k} Z_{kt}^\prime + \theta^*) - \mT(\theta^*) ))\|_m^2]},
\end{align*}
and
\begin{align}
&\E\Big[\Big\|\big( \mT(\sqrt{\alpha} Z_{kt}^\prime + q^*) - \mT(\theta^*)  - \sqrt{k}\Big(\mT\Big(\sqrt{\frac{\alpha}{k}} Z_{kt}^\prime + \theta^*\big) - \mT(\theta^*) \Big)\Big)\Big\|_m^2\Big]  \nonumber\\
=&  \E\bigg[\Big\|\big( \mT(\sqrt{\alpha} Z_{kt}^\prime + \theta^*) - \mT(\theta^*)  - \sqrt{k}\Big(\mT\Big(\sqrt{\frac{\alpha}{k}} Z_{kt}^\prime + \theta^*\big) - \mT(\theta^*) \Big)\Big)\Big\|_m^2\mathbbm{1}(\alpha^{\frac{1}{2}}Z_{kt}^\prime \in B^d(0, \frac{\Delta}{2}))\bigg] \label{eq:q-limit-t1}\\
&+\E\bigg[\Big\| \mT(\sqrt{\alpha} Z_{kt}^\prime + \theta^*) \!-\! \mT(\theta^*)  - \sqrt{k}\Big(\mT\Big(\sqrt{\frac{\alpha}{k}} Z_{kt}^\prime + \theta^*\Big) - \mT(\theta^*) \Big)\Big\|_m^2\mathbbm{1}(\alpha^{\frac{1}{2}}Z_{kt}^\prime \notin B^d(0, \frac{\Delta}{2}))\bigg]\label{eq:q-limit-t2}. 
\end{align}

For term \eqref{eq:q-limit-t1}, when $\alpha \leq 1$, we have $\|\sqrt{\frac{\alpha}{k}} Z_{kt}^\prime\|_c\leq \frac{\Delta}{2}$ and $\|\sqrt{\alpha} Z_{kt}^\prime\|_c\leq \frac{\Delta}{2}$ because $k \geq 1.$ By equation \eqref{eq:q-ph1}, we have $\mT$ to be positive homogenoues of degree 1 at $q^*$. Therefore, we have $\eqref{eq:q-limit-t1} = 0$. 
\begin{align*}
\eqref{eq:q-limit-t2} &\leq \frac{2\gamma\alpha u_{cm}^2}{l_{cm}^2}\E[\|Z_{kt}^\prime\|_m^2\mathbbm{1}(\alpha^{\frac{1}{2}}Z_{kt}^\prime \notin B^d(0, \frac{\Delta}{2}))]\\
&\leq \frac{2\gamma\alpha u_{cm}^2}{l_{cm}^2}\sqrt{\E[\|Z_{kt}^\prime\|_m^4]}\sqrt{\P(\|Z_{kt}^\prime\|_2^4 \geq \frac{\Delta^4}{16\alpha^2
})}\leq \frac{2\gamma\alpha u_{cm}^2}{l_{cm}^2}\sqrt{\E[\|Z_{kt}^\prime\|_m^4]} \sqrt{\frac{16\alpha^2\E[\|Z_{kt}^\prime\|_2^4]}{\Delta^4}}  \in \mathcal{O}(\alpha^2).
\end{align*}
Combining terms \eqref{eq:q-limit-t1} and \eqref{eq:q-limit-t2} together, we obtain $T_{13}\in \mathcal{O}(\alpha^{\frac{3}{2}})$.

Therefore, by keeping other analyses in the proof of Theorem \ref{thm: additive limit}, we prove that there exists a unique random variable $Y$ such that 
$
W_2\big(\mathcal{L}(Z^{(\alpha
)}), \mathcal{L}(Y)\big) \in \mathcal{O}(\alpha^\frac{1}{4}),
$
where $Y$ only depends on $\mT$ and $\operatorname{Var}(\tmT(\{D_0, P_0, r_0\}, q^*))$.
Furthermore, by Proposition \ref{prop:gaussian}, we have $$W_2\big(\mathcal{L}(Y^{(\alpha
)}), \mathcal{L}(Y)\big) \leq W_2\big(\mathcal{L}(Y^{(\alpha
)}), \mathcal{L}(Z^{(\alpha
)})\big)+W_2\big(\mathcal{L}(Z^{(\alpha
)}), \mathcal{L}(Y)\big)\in \mathcal{O}(\alpha^\frac{1}{4}),
$$
thereby completing the proof of Theorem \ref{thm: q limit}.

\section{Proof of Theorem \ref{thm: q EY}}\label{sec:q-EY}

By Corollary \ref{thm: q convergence} and equation \eqref{eq:q_learning_SA}, we have the following equalities in distribution:
\begin{equation}\label{eq: q equation in distribution}
Y^{(\alpha)} \overset{\textup{d}}{=} Y^{(\alpha)} + \sqrt{\alpha}\big(\tmT(\{D_0,P_0,r_0\}, \sqrt{\alpha}Y^{(\alpha)}+q^*) - \sqrt{\alpha}Y^{(\alpha)}-q^*\big).
\end{equation}
Taking expectation to both sides of the above equation, we obtain
\begin{align*}
\E[Y^{(\alpha)}] &= \frac{1}{\sqrt{\alpha}}\E[\mT(\sqrt{\alpha}Y^{(\alpha)}+q^*)-q^*]
= \gamma DP\E[f(Y^{(\alpha)}+\frac{q^*}{\sqrt{\alpha}})-f(\frac{q^*}{\sqrt{\alpha}})] + (I-D)\E[Y^{(\alpha)}],
\end{align*}
where the last equality holds because $f$ is positive homogeneous of degree 1. Rearranging terms, we obtain the equality
$
\E[Y^{(\alpha)}] = \gamma P\E[f(Y^{(\alpha)}+\frac{q^*}{\sqrt{\alpha}})-f(\frac{q^*}{\sqrt{\alpha}})].
$

Recall the $g$ defined in the proof of Proposition \ref{prop:q} and we define  
$
h(Y^{(\alpha)}, \frac{q^*}{\sqrt{\alpha}}):= f(Y^{(\alpha)}+\frac{q^*}{\sqrt{\alpha}})-f(\frac{q^*}{\sqrt{\alpha}}) - g(Y^{(\alpha)}).
$
Therefore, we have $\E[Y^{(\alpha)}] = \E[\gamma P g(Y^{(\alpha)}) + \gamma P h(Y^{(\alpha)}, \frac{q^*}{\sqrt{\alpha}})].$ By Fatou's lemma \cite[Exercise 3.2.4]{durrett2019probability} and the following Lemma \ref{lemma: h}, we obtain
\[\|E[h(Y^{(\alpha)}, \frac{q^*}{\sqrt{\alpha}})]\|_c \leq E[\|h(Y^{(\alpha)}, \frac{q^*}{\sqrt{\alpha}})\|_c] \leq \sqrt{E[\|h(Y^{(\alpha)}, \frac{q^*}{\sqrt{\alpha}})\|_c^2]} \in \mathcal{O}(\sqrt{\alpha}).\]
\begin{lemma}\label{lemma: h}
Consider iterates $\{Y_t\}_{t \geq 0}$ generated by equation \eqref{eq:q_learning_SA} with stepsize $\alpha$, under the same setting as Corollary~\ref{thm: q 2n moment} with $n = 2$, we obtain $\mathbb{E}\left[\|h(Y_t, \frac{q^*}{\sqrt{\alpha}})\|_c^2\right] \in \mathcal{O}(\alpha) .$

\end{lemma}
We leave the proof of Lemma \ref{lemma: h} to the end of this section. It is well known that weak convergence in $\mathcal{P}_2(\mathbb{R}^{|\mS||\mA|})$ is equivalent to convergence in distribution and the convergence of the first two moments. By \cite[Exercise 3.2.5]{durrett2019probability} and the Lipschitz continuity of $g$, we obtain $\lim_{\alpha \to 0}\E[Y^{(\alpha)}] = \E[Y]$ and $\lim_{\alpha \to 0}\E[g(Y^{(\alpha)})] = \E[g(Y)].$ Therefore, we have
$
    \E[Y] = \gamma P\E[g(Y)].
$
Below, we discuss the $\E[Y]$ in three cases.
\paragraph{Case 1:}
    Suppose there exists a state $s^\prime$ that is both tied and not rooted. It is straight forward to verify that the second condition of Theorem \ref{thm: additive smooth} is satisfied given that $\operatorname{Var}(\tmT(q^*, \{D_0, P_0, r_0\}))$ is positive definite. Therefore, by Theorem \ref{thm: additive smooth}, we conclude that $\E[Y] \neq 0$ for Case 1.

\paragraph{Case 2:} If there is no tied state, by definition, $g(\cdot)$ will be a linear function. Recall that
\[\E[Y^{(\alpha)}] = \gamma P \E [g(Y^{(\alpha)}) + h(Y^{(\alpha)}, \frac{q^*}{\sqrt{\alpha}})] .\]
For $n \geq 2$, by \eqref{eq: h} in the proof of Lemma \ref{lemma: h}, Assumption \ref{as: q noise}(\textbf{n}) and the above equation, we obtain
\begin{align*}
\|\E[Y^{(\alpha)}]\|_c &\leq \|\gamma P \E[g(Y^{(\alpha)})]\|_c + \|\gamma P \E[h(Y^{(\alpha)}, \frac{q^*}{\sqrt{\alpha}})]\|_c \\
&\leq \gamma\|  g(\E[Y^{(\alpha)}])\|_c+ \gamma\|  \E[h(Y^{(\alpha)}, \frac{q^*}{\sqrt{\alpha}})]\|_c\\
&\leq \gamma\|\E[Y^{(\alpha)}]\|_c+ \gamma\|  \E[h(Y^{(\alpha)}, \frac{q^*}{\sqrt{\alpha}})]\|_c\\
&\leq \gamma\|\E[Y^{(\alpha)}]\|_c + \gamma \left(\E\|f(Y^{(\alpha)})-g(Y^{(\alpha)})\|_c^{2n}\right)^{\frac{1}{2n}} \cdot \left(\mathbb{P}(\|Y^{(\alpha)}\|_c \geq \frac{\Delta}{2\sqrt{\alpha}})\right)^{\frac{2n-1}{2n}}
\end{align*}
We continue by bounding the second right hand term and obtain
\begin{align*}
\|\E[Y^{(\alpha)}]\|_c 
&\overset{(i)}{\leq  }\gamma\|\E[Y^{(\alpha)}]\|_c + \mathcal{O}\Big(\big(\mathbb{P}(\|Y^{(\alpha)}\|_c \geq \frac{\Delta}{2\sqrt{\alpha}})\big)^{\frac{2n-1}{2n}}\Big)\\
&\leq \gamma\|\E[Y^{(\alpha)}]\|_c + \mathcal{O}\Big(\big(\frac{\E[\|Y^{(\alpha)}\|_c^{2n}]4^n\alpha^n}{\Delta^{2n}}\big)^{\frac{2n-1}{2n}}\Big)\overset{(ii)}{\leq} \gamma\|\E[Y^{(\alpha)}]\|_c +\mathcal{O} (\alpha^{\frac{2n-1}{2}} ),
\end{align*}
where (i) and (ii) hold by Corollary \ref{thm: q 2n moment} and Fatou's lemma. Then, we have 
$
\E[Y^{(\alpha)}] \in \mathcal{O} (\alpha^{\frac{2n-1}{2}} ),
$
which implies $\E[Y] = 0$. Furthermore, recall that $\E[Y^{(\alpha)}] = \frac{\E[q^{(\alpha)}] - q^*}{\sqrt{\alpha}}$ and we obtain
$
\E[q^{(\alpha)}] = q^* + \mathcal{O}(\alpha^n).
$

\paragraph{Case 3:}
If tied states are always rooted states, the MDP other than all these tied and rooted states will form a new MDP with no tied state. Then, for any state $s$ and action $a$ in the new MDP, we have proved that $\E[Y^{(\alpha)}(s,a)]\in \mathcal{O} (\alpha^{\frac{2n-1}{2}} )$. We notice that, for any state $s$ that is tied and rooted, the next state $s^\prime$ is always in the new MDP by definition of rooted state. Therefore, for any state $s$ that is tied and rooted and action $a$, we obtain
\begin{align*}
\E[Y^{(\alpha)}(s,a)] &= \gamma \sum_{s^\prime}P(s^\prime|s,a)\big(E[g_{s^\prime}(Y^{(\alpha)})] + \E[h_{s^\prime}(Y^{(\alpha)}, \frac{q^*}{\sqrt{\alpha}})]\big)\\
&= \gamma \sum_{s^\prime}P(s^\prime|s,a)\big(g_{s^\prime}(E[Y^{(\alpha)}]) + \E[h_{s^\prime}(Y^{(\alpha)}, \frac{q^*}{\sqrt{\alpha}})]\big) \in \mathcal{O}\big(\alpha^{\frac{2n-1}{2}}\big).
\end{align*}
Then, we conclude that $\E[Y] = 0$ and $\E[q^{(\alpha)}] = q^* + \mathcal{O}(\alpha^n)$.
\subsection{Proof of Lemma \ref{lemma: h}}
By definition, for $\forall s \in \mS$, we obtain
\begin{align*}
h_s(Y_t, \frac{q^*}{\sqrt{\alpha}}) &= f_s(Y_t+\frac{q^*}{\sqrt{\alpha}}) - f_s(\frac{q^*}{\sqrt{\alpha}})- g_s(Y_t)\\
&= \max_{a \in \mA}\big(Y_t(s,a) +\frac{q^*(s,a)}{\sqrt{\alpha}}\big) - \max_{a \in \mA}\frac{q^*(s,a)}{\sqrt{\alpha}} - \max_{a \in \mA^*(s)}Y_t(s,a) \\
&= \max_{a \in \mA}\big(Y_t(s,a) +\frac{q^*(s,a)}{\sqrt{\alpha}}\big) - \max_{a \in \mA^*(s)}\big(Y_t(s,a) +\frac{q^*(s,a)}{\sqrt{\alpha}}\big),
\end{align*}
where the last inequality holds because $\mA^*(s) = \argmax_{a \in \mA}q^*(s,a)$.

We can obtain that $h_s(Y_t, \frac{q^*}{\sqrt{\alpha}}) \geq 0$. Recall the function $\Delta:\mS \to \mathbb{R}$ defined as $\Delta(s) := \infty$ if $\mA = \mA^*(s)$ and $\Delta(s):=
\max_{a \in \mA} q^*(s,a) -\max_{a \in \mA \setminus \mA^*(s)}q^*(s,a)$ if $\mA \neq \mA^*(s)$,
We observe that when $\|Y_t\|_c \leq \frac{\Delta(s)}{2\sqrt{\alpha}}$, it holds that $h_s(Y_t, \frac{q^*}{\sqrt{\alpha}}) = 0$.
Therefore, with $\Delta_{\min} = \min_{s \in \mS}\Delta(s)$, we conclude that
\begin{equation}\label{eq: h}
\begin{aligned}
0\leq h_s(Y_t, \frac{q^*}{\sqrt{\alpha}}) &\leq \big(f_s(Y_t+\frac{q^*}{\sqrt{\alpha}}) - f_s(\frac{q^*}{\sqrt{\alpha}})- g_s(Y_t)\big)\mathbbm{1}_{\{\|Y_t\|_c \geq \frac{\Delta(s)}{2\sqrt{\alpha}}\}}\\
&\leq \big(f_s(Y_t)-g_s(Y_t)\big)\mathbbm{1}_{\{\|Y_t\|_c \geq \frac{\Delta_{\min}}{2\sqrt{\alpha}}\}},
\end{aligned}
\end{equation}

By Cauchy–Schwarz inequality, we obtain
\begin{align*}
\mathbb{E}\big[\|h(Y_t, \frac{q^*}{\sqrt{\alpha}})\|_c^2\big] &\leq \sqrt{\mathbb{E}\|f(Y_t)-g(Y_t)\|_c^4} \cdot \sqrt{\mathbb{P}(\|Y_t\|_c \geq \frac{\Delta_{\min}}{2\sqrt{\alpha}})}\\
&\leq \sqrt{8\mathbb{E}\|f(Y_t)\|_c^4+8\mathbb{E}\|g(Y_t)\|_c^4}\cdot \sqrt{\frac{\mathbb{E}(\|Y_t\|_c^4)16\alpha^2}{\Delta_{\min}^4}}\\
&\overset{(\text{i})}{\leq} \sqrt{16\mathbb{E}\|Y_t\|_c^4}\cdot \sqrt{\frac{\mathbb{E}(\|Y_t\|_c^4)16\alpha^2}{\Delta_{\min}^4}} \overset{(\text{ii})}{\in}\mathcal{O}(\alpha),
\end{align*}
where (i) holds because the non-expansion of $f(\cdot)$ and $g(\cdot)$ with respect to $\|\cdot\|_c$ and (ii)  holds because of the  Corollary \ref{thm: q 2n moment} with $n = 2$.

\section{Proof of Proposition \ref{co: average}} \label{sec:proof_average}

We prove the first and second moment bounds in Proposition \ref{co: average}. We start with the following lemma.

\begin{lemma}\label{lemma: general convergence rate}
For iterates $\theta_t^{(\alpha)}$ that are generated by equation \eqref{eq: markovian general SA} and satisfy the Condition \ref{condition: geo convergence} and \ref{condition: bias}, there exist two universal constants $C_2$ and $C_3$ such that
\begin{enumerate}
\item $\E[\|\theta^{(\alpha)}\|_2] \in \mathcal{O}(1) \quad \text{and} \quad \var(\theta^{(\alpha)}) \in \mathcal{O}(\alpha\tau_\alpha).$
    \item$
\|\E[\theta_t^{(\alpha)}] - \E[\theta^{(\alpha)}]\|_2 \leq C_2 \cdot (1-\alpha C_1 )^\frac{t}{2}, \quad \forall \alpha \leq \bar{\alpha} \text{ and } t \geq \tau_\alpha.$ 
\item $
\|\E[\theta_t^{(\alpha)}{\theta_t^{(\alpha)}}^T] - \E[\theta^{(\alpha)}{\theta^{(\alpha)}}^T]\|_2 \leq C_3 \cdot (1-\alpha C_1 )^\frac{t}{2}, \quad \forall \alpha \leq \bar{\alpha} \text{ and } t \geq \tau_\alpha.$ 
\end{enumerate}

\end{lemma}

\begin{proof}{Proof of Lemma \ref{lemma: general convergence rate}}

By Condition \ref{condition: geo convergence}, we obtain
\[
\mathbb{E}[\Vert \theta^{(\alpha)}\Vert^2_2] 
\leq \underbrace{2\mathbb{E}(\Vert \theta^{(\alpha)} - \theta^*\Vert^2_2)}_{\in \mathcal{O}(\alpha)} + \underbrace{2\Vert \theta^*\Vert^2_2}_{\in \mathcal{O}(1)} \in \mathcal{O}(1)
  \;\;\text{ and }\;\;
\|\var(\theta^{(\alpha)})\|_2 
\leq \E[\|\theta^{(\alpha)} - \theta^*\|_2^2] \in \mathcal{O}(\alpha \tau_\alpha).
\]
By \cite[Theorem 4.1]{villani2009optimal}, there exists a coupling between $\theta_t$ and $\theta^{(\alpha)}$ such that 
$
W^2_2(\mathcal{L}(\theta_t), \mathcal{L}(\theta^{(\alpha)})) = \mathbb{E}[\Vert \theta_t - \theta^{(\alpha)}\Vert_c^2].
$
By the above bounds and applying Jensen's inequality twice, we obtain that 
\begin{align*}
\Vert\mathbb{E}[\theta_t - \theta^{(\alpha)}]\Vert_2^2 &\leq \big(\mathbb{E}[\Vert \theta_t - \theta^{(\alpha)}\Vert_2]\big)^2
\leq \mathbb{E}[\Vert \theta_t - \theta^{(\alpha)}\Vert_2^2]
\leq \frac{1}{l_{cs}^2}\mathbb{E}[\Vert \theta_t - \theta^{(\alpha)}\Vert_c^2] \leq \frac{C_0}{l_{cs}^2} (1-\alpha C_1 )^t.
\end{align*}
We thus have 
$
\Vert\mathbb{E}[\theta_t] - \mathbb{E}[\theta^{(\alpha)}]\Vert_2 \leq \mathbb{E}[\Vert \theta_t - \theta^{(\alpha)}\Vert_2] \leq C_2 \cdot (1-\alpha C_1 )^\frac{t}{2}.
$

For the second moment, by \cite[Equation A.28]{huo2023bias}, we obtain
\begin{equation}\label{eq: secondmoment}
\begin{aligned}
  \|\mathbb{E} [\theta_t \theta_t^{\top} ]-\mathbb{E} [\theta^{(\alpha)} {\theta^{(\alpha)}}^{\top} ] \|_2 & \leq \mathbb{E} [ \|\theta_t-{\theta^{(\alpha)}} \|_2^2 ]+2 (\mathbb{E} [ \|\theta_t-{\theta^{(\alpha)}} \|_2^2 ] \mathbb{E} [ \|{\theta^{(\alpha)}} \|_2^2 ] )^{1 / 2}\\
&\leq \mathbb{E} [ \|\theta_t-{\theta^{(\alpha)}} \|_2^2 ]+2 (\mathbb{E} [ \|\theta_t-{\theta^{(\alpha)}} \|_2^2 ] \mathbb{E} [2 \|{\theta^{(\alpha)}} - \theta^* \|_2^2+2 \|{\theta^*} \|_2^2 ] )^{1 / 2}\\
\end{aligned}
\end{equation}
Meanwhile, we have
$
\mathbb{E} [ \|\theta_t-{\theta^{(\alpha)}} \|_2^2 ] \leq \frac{C_0}{l_{cs}^2} (1-\alpha C_1 )^t 
\text{ and }
\mathbb{E} [ \|{\theta^{(\alpha)}} - \theta^* \|_2^2 ] \in \mathcal{O}(\alpha\tau_\alpha ).
$
Substituting the above bounds into the right-hand side of inequality \eqref{eq: secondmoment} yields
$
 \|\mathbb{E} [\theta_t \theta_t^{\top} ]-\mathbb{E}\big[{\theta^{(\alpha)}} {\theta^{(\alpha)}}^{\top}\big] \|_2 \leq C_3 \cdot (1-\alpha C_1 )^\frac{t}{2}.
$
\end{proof}

\subsection{First Moment}
First, we have
$
\mathbb{E} [\bar{\theta}_{k_0, k} ]-\theta^*= (\mathbb{E} [\theta^{(\alpha)} ]-\theta^* )+\frac{1}{k-k_0} \sum_{t=k_0}^{k-1} \mathbb{E} [\theta_t-\theta^{(\alpha)} ].
$
To bound the sum on the right hand side, we use Lemma \ref{lemma: general convergence rate} to obtain
$
\Vert\mathbb{E}[\theta_k] - \mathbb{E}[{\theta^{(\alpha)}}]\Vert_2  \leq C_2 \cdot (1-\alpha C_1 )^\frac{k}{2}.
$
It follows that
\begin{equation*}
\begin{aligned}
  \|\sum_{t=k_0}^{k-1} \mathbb{E} [\theta_t-{\theta^{(\alpha)}} ] \|_2 
 &\leq \sum_{t=k_0}^{k-1} \|\mathbb{E} [\theta_t ]-\mathbb{E}[\theta^{(\alpha)}] \|_2 \\
& \leq C_2 \cdot (1-\alpha C_1 )^\frac{k_0}{2}\frac{1}{1 - \sqrt{1-\alpha C_1}}
 \leq C_2 \cdot (1-\alpha C_1 )^\frac{k_0}{2}\frac{2}{\alpha C_1} \\
& \stackrel{(i)}{\leq} C_2 \cdot \exp\Big(-\frac{\alpha C_1k_0}{2}\Big)\frac{2}{(1-\sqrt{\gamma}) \alpha}\leq C \cdot \frac{1}{\alpha} \cdot \exp\Big(-\frac{\alpha C_1k_0}{2}\Big),
\end{aligned}
\end{equation*}
where $(i)$ follows from the inequality that $(1-u)^m\leq \exp(-um)$ for $0<u<1.$

Together with Condition \ref{condition: bias} we have
    \[\mathbb{E} [\bar{\theta}_{k_0, k} ]-\theta^*= \alpha^\beta B  + o(\alpha^{\beta+\delta})+\mathcal{O} \Big(\frac{1}{\alpha(k - k_0)} \exp \Big(-\frac{\alpha C_1 k_0}{2} \Big) \Big),\]
thereby finishing the proof of Proposition \ref{co: average} for the first moment.

\subsection{Second Moment}

In this subsection, we follow the proof technique in \cite[Section A.6.2]{huo2023bias} 
to bound the second moment of the tail-averaged iterate. Here we consider the same decomposition:
\begin{align*}
 \mathbb{E} [ (\bar{\theta}_{k_0, k}-\theta^* ) (\bar{\theta}_{k_0, k}-\theta^* )^{\top} ] = & \underbrace{\mathbb{E} [ (\bar{\theta}_{k_0, k}-\mathbb{E} [\theta^{(\alpha)} ] ) (\bar{\theta}_{k_0, k}-\mathbb{E} [\theta^{(\alpha)} ] )^{\top} ]}_{T_1}+\underbrace{\mathbb{E} [ (\bar{\theta}_{k_0, k}-\mathbb{E} [\theta^{(\alpha)} ] ) (\mathbb{E} [\theta^{(\alpha)} ]-\theta^* )^{\top} ]}_{T_2} \\
&+  \underbrace{\mathbb{E} [ (\mathbb{E} [\theta^{(\alpha)} ]-\theta^* ) (\bar{\theta}_{k_0, k}-\mathbb{E} [\theta^{(\alpha)} ] )^{\top} ]}_{T_3}+\underbrace{\mathbb{E} [ (\mathbb{E} [\theta^{(\alpha)} ]-\theta^* ) (\mathbb{E} [\theta^{(\alpha)} ]-\theta^* )^{\top} ]}_{T_4}.
\end{align*}

For $T_2$, we have
\begin{align*}
T_2 &= \frac{1}{k-k_0} (\sum_{t=k_0}^{k-1} \mathbb{E} [\theta_t-{\theta^{(\alpha)}} ] ) (\mathbb{E}[\theta^{(\alpha)}]-\theta^* )^{\top}\\
&=\mathcal{O} (\frac{1}{\alpha(k - k_0)} \exp (-\frac{\alpha C_1 k_0}{2} ) ) \cdot (\alpha^d B + o(\alpha^{\beta+\delta}))
\in  \mathcal{O} (\frac{\alpha^{\beta-1}}{(k - k_0)} \exp (-\frac{\alpha C_1 k_0}{2} ) ).
\end{align*}
The term $T_3$ is similar to $T_2$ and obeys the same bound.
For $T_4$, we have
$
T_4 = (\alpha^\beta B + o(\alpha^{\beta+\delta}))(\alpha^\beta B + o(\alpha^{\beta+\delta}))^T
= \alpha^{2\beta}BB^T + o(\alpha^{2\beta+\delta}).
$

For $T_1$, we have
\begin{align}
 T_1
 =&\frac{1}{ (k-k_0 )^2} \sum_{t=k_0}^{k-1} \mathbb{E} [\big(\theta_t-\mathbb{E}[\theta^{(\alpha)}]\big)\big(\theta_t-\mathbb{E}[\theta^{(\alpha)}]\big)^{\top} ] \label{eq:T1_T_{11}}\\
&+\frac{1}{ (k-k_0 )^2} \sum_{t=k_0}^{k-1} \sum_{l=t+1}^{k-1}\mathbb{E} [\big(\theta_t-\mathbb{E}[\theta^{(\alpha)}]\big)\big(\theta_l-\mathbb{E}[\theta^{(\alpha)}]\big)^{\top} ] \label{eq:T1_T_{12}} \\
& +\frac{1}{ (k-k_0 )^2} \sum_{t=k_0}^{k-1} \sum_{l=t+1}^{k-1}\mathbb{E} [\big(\theta_l-\mathbb{E}[\theta^{(\alpha)}]\big)\big(\theta_t-\mathbb{E}[\theta^{(\alpha)}]\big)^{\top} ]. \label{eq:T1_T_{13}} 
\end{align}

By the same decomposition \cite{huo2023bias} and Lemma \ref{lemma: general convergence rate}, we have $ \mathbb{E} [ (\theta_t-\mathbb{E}[\theta^{(\alpha)}] ) (\theta_t-\mathbb{E}[\theta^{(\alpha)}] )^{\top} ] \in  \mathcal{O} ((1-\alpha C_1)^{\frac{t}{2}} + \alpha\tau_\alpha ).$ Therefore, for the term in \eqref{eq:T1_T_{11}}, we have
\begin{align*}
\eqref{eq:T1_T_{11}} & \in\frac{1}{ (k-k_0 )^2} \sum_{t=k_0}^{k-1} \mathcal{O} ((1-\alpha C_1)^{\frac{t}{2}} + \alpha\tau_\alpha ) \\
& \in\mathcal{O} \Big(\frac{1}{ (k-k_0 )^2} \sum_{t=k_0}^{\infty} (1-\alpha C_1 )^{\frac{t}{2}} \Big)+\mathcal{O} \Big(\frac{\alpha \tau_\alpha}{k-k_0} \Big) \in \mathcal{O} \Big(\frac{1}{\alpha (k-k_0 )^2} \exp \Big(-\frac{\alpha C_1k_0}{2}\Big)+\frac{\alpha \tau_\alpha}{k-k_0} \Big) .
\end{align*}
We make the following claim, whose proof is nearly identical as the proof of Claim 4 in \cite{huo2023bias}.
\begin{claim}\label{claim: TA}
For $t \geq \frac{2}{\alpha C_1}\log (\frac{1}{\alpha\tau_\alpha} )$ and $l \geq t+\tau_\alpha$, we have $ \|\mathbb{E}[(\theta_t-\mathbb{E}[\theta^{(\alpha)}])(\theta_l-\mathbb{E}[\theta^{(\alpha)}])^{\top}] \|\in \mathcal{O} ((\alpha \tau_\alpha) \cdot (1-\alpha C_1 )^{\frac{(l-t)}{2}} ).$
\end{claim}
By this claim, we have 
\eqref{eq:T1_T_{12}} $\in \mathcal{O}\big(\frac{\tau_\alpha}{k-k_0}\big).$
Similarly, we have
\eqref{eq:T1_T_{13}}$\in \mathcal{O}\big(\frac{\tau_\alpha}{k-k_0}\big) .
$
Therefore, we have
\begin{equation}\label{eq: TTTT}
T_1  \in \mathcal{O}\bigg(\frac{1}{\alpha (k-k_0 )^2} \exp \Big(-\frac{\alpha C_1k_0}{2}\Big)+\frac{\tau_\alpha}{k-k_0}\bigg) .
\end{equation}

By summing up the bounds for $T_1 \sim T_4$, we obtain
\begin{align*}
\mathbb{E} [ (\bar{\theta}_{k_0, k}-\theta^* ) (\bar{\theta}_{k_0, k}-\theta^* )^{\top} ] =& \alpha^{2\beta}BB^T + o(\alpha^{2\beta+\delta}) + \mathcal{O}\bigg(\frac{\alpha^{\beta-1}}{(k - k_0)} \exp\Big(-\frac{\alpha C_1 k_0}{2}\Big)\bigg)\\
&+\mathcal{O}\bigg(\frac{1}{\alpha (k-k_0 )^2} \exp \Big(-\frac{\alpha C_1k_0}{2}\Big)+\frac{\tau_\alpha}{k-k_0}\bigg)\\
=& \alpha^{2\beta}BB^T + o(\alpha^{2\beta+\delta}) + \mathcal{O}\bigg(\frac{1}{\alpha (k-k_0 )^2} \exp \Big(-\frac{\alpha C_1k_0}{2}\Big)+\frac{\tau_\alpha}{k-k_0}\bigg).
\end{align*}

\end{document}